%% file: main.tex
\documentclass[mnsc,non-blindrev]{cls/informs3}

\OneAndAHalfSpacedXI

\TheoremsNumberedThrough     
\ECRepeatTheorems

\EquationsNumberedThrough

\input{macros}

\allowdisplaybreaks
\begin{document}

\RUNAUTHOR{}

\RUNTITLE{}

\TITLE{Weak-to-Strong Learning in Decision Making }

\ARTICLEAUTHORS{
\AUTHOR{Jingwei Ji}
\AFF{Management Science and Engineering, Stanford University, \EMAIL{jingwei.ji@stanford.edu} }
\AUTHOR{Renyuan Xu}
\AFF{Management Science and Engineering, Stanford University, \EMAIL{renyuanxu@stanford.edu} }
} 

\ABSTRACT{
Many operational decisions rely on predictive models that estimate uncertain outcomes conditional on observable contexts. Training such models, however, often faces a fundamental data asymmetry: labeled outcomes are scarce or costly to obtain, while contextual covariates are abundant.
Motivated by this data asymmetry, we develop a decision-aware weak-to-strong (W2S) framework that leverages both labeled and unlabeled data to improve contextual stochastic optimization. Specifically, we first train a weak model using limited labeled data and then use it to generate predicted outcome distributions on unlabeled contexts. These distributions provide soft supervision for training a strong model. We establish a non-asymptotic upper bound on the excess decision risk of W2S and a complementary lower bound for a strong-only benchmark. Their comparison yields explicit sufficient conditions under which W2S improves downstream decision performance.
The key quantity is the correlation dimension between the weak and strong feature representations: when it is small, abundant unlabeled data reduce the effect of teacher errors along non-overlapping directions. A synthetic newsvendor experiment and a comment moderation experiment based on real-world data provide empirical evidence consistent with the theory.
}


\maketitle

\section{Introduction} \label{sec:intro}

Many operational decisions rely on predictive models that estimate uncertain outcomes conditional on observable contexts. Examples include inventory planning and newsvendor problems with contextual demand forecasts \citep{ban2019big, chang2025feature, chen2021nonparametric}; dynamic pricing, revenue management, online retail pricing, and joint pricing-inventory learning with demand prediction \citep{cohen2020feature, javanmard2024multi, ferreira2016analytics, chen2026utility, chen2022dynamic}; routing with travel-time estimation \citep{guo2023data}; and portfolio allocation with return-distribution modeling \citep{gu2020empirical}.
Related formulations also appear in data-driven robust optimization and offline policy learning \citep{wang2016likelihood, zhou2023offline}; see \citet{bastani2022applied} for an overview of machine learning in operations management. These settings can be unified as instances of \textit{contextual stochastic optimization}.

When predictions are used as input to operational decisions, a growing literature recognizes that model training should account for the optimization problems in which those predictions will be used.
The key insight is simple: predictive accuracy alone does not guarantee decision quality. Errors that are small under standard statistical losses can induce large downstream cost if they distort decision-critical directions.
This observation has motivated decision-aware \citep{elmachtoub2022smart} and integrated learning frameworks \citep{qi2025integrated} that explicitly align predictive training with operational objectives.
In this work, we adopt the integrated conditional estimation-optimization (ICEO) framework proposed by \cite{qi2025integrated}.

Yet an increasingly common practical challenge lies outside the standard formulation, which assumes sufficiently many labeled outcomes.
In many applications, labeled outcomes are scarce and/or costly while contextual covariates are abundant.
At the same time, practitioners often have access to multiple models or feature representations of different strengths.
For example, they may have access to a smaller model that can be reliably trained on limited labeled data and a larger model that is more expressive but requires more compute to train.

This raises a natural question which we aim to answer rigorously:
\begin{center}
\textit{
    Can we leverage a weak model trained on limited labeled data to guide the training of a stronger model using abundant unlabeled data, thereby improving \textbf{downstream decision} performance in contextual stochastic optimization?
    }
\end{center}

We answer this question through a weak-to-strong learning framework tailored to downstream decision making problems. The weak model is first adapted using the labeled data and then used to produce predicted outcome distributions on unlabeled contexts. These predicted distributions serve as soft supervision for training the strong model. The resulting strong model induces a plug-in decision policy, and its performance is evaluated by downstream decision risk rather than prediction loss.

We focus on a label-scarce regime in which the weak model contains useful task-relevant signal after adaptation, while the stronger model is expressive but difficult to train reliably from the limited labeled data alone. In this regime, abundant unlabeled contexts create an opportunity for knowledge transfer: the weak model can provide task-specific supervision across many contexts, and the strong model can use this supervision to learn a better decision policy.
Such regimes arise in modern decision-making applications where decision-relevant outcomes are costly, delayed, censored, or observed only under historical policies, including personalized pricing and recommendation, medical treatment decisions, vehicle routing, and inventory control \citep{shi2016nonparametric,zhan2023policy,cao2025collaborative,keyvanshokooh2025contextual,serrano2026contextual}.
Similar weak-to-strong procedures have been shown to let a strong model outperform its weak supervisor in prediction settings, including classification and language modeling \citep{burns2024weak,dong2025discrepancies}. Whether this phenomenon can be translated into downstream decision making, where performance is measured by decision risk rather than prediction loss, remains open.

\subsection{Our contribution}

We summarize our contributions as follows:
\begin{enumerate}
    \item \textbf{W2S formulation in decision making.} Existing W2S theory has largely focused on predictive objectives such as classification and regression \citep{burns2024weak,dong2025discrepancies,lang2024theoretical,charikar2024quantifying}, leaving open whether and when W2S can improve \textit{decision making} problems.
   We extend the W2S paradigm to downstream decision making by formulating weak-to-strong transfer within contextual stochastic optimization.
In our formulation, the weak model does not merely provide pseudo-labels for prediction; instead, it produces pseudo-distributions over uncertain outcomes on unlabeled contexts, which are then used to train a stronger model whose value is measured by the decision risk of its induced plug-in policy.
This provides a decision-theoretic modeling framework for studying when weak supervision can improve operational decisions, and lays the groundwork for extending W2S analysis to richer decision settings such as sequential decision making and reinforcement learning.
    \item \textbf{Algorithm and its theoretical understanding.}
    Building on this formulation, we develop and analyze a decision-aware W2S training algorithm.
We prove a non-asymptotic upper bound on the excess decision risk of the induced W2S policy in Theorem~\ref{thm:upper_bound_w2s}.
By comparing this bound with a strong-only benchmark trained directly on labeled data, we obtain an explicit certificate in Corollary~\ref{cor:w2s_outperformance_certificate} for when W2S improves downstream decision performance.
The comparison reveals a key structural mechanism: W2S is most beneficial when the weak and strong representations have limited overlap, so that errors made by the weak teacher are less likely to align systematically with the directions used by the strong model.
In this case, abundant unlabeled contexts can dilute teacher errors rather than simply transfer them to the strong model.
    \item \textbf{Empirical evidence.}  We complement the theory with two empirical studies. We first use a controlled synthetic experiment to test whether the regimes predicted by the theory are observed in simulation. We then evaluate W2S on a real-world comment moderation task, where predictions guide automatic and human-review moderation decisions. In both studies, the qualitative patterns are consistent with the theory: W2S delivers its largest gains when labeled data are scarce and unlabeled contexts are plentiful, and these benefits shrink as labels accumulate.
\end{enumerate}

This data asymmetry is pervasive in Operations Research: contextual information is often available at scale, while reliable outcome labels are scarce, delayed, or shaped by historical decisions, as in inventory and pricing with realized demand or routing with travel-time uncertainty \citep{ban2019big,cohen2020feature,guo2023data}.
The weak-to-strong framework developed in this paper provides a first step toward exploiting this asymmetry for downstream decision making.
It uses scarce labeled outcomes to extract task-specific signal from a weak decision-aware model and transfers this signal across abundant unlabeled contexts to adapt a stronger model, suggesting a promising route for broader label-limited decision problems.

\subsection{Relevant literature}

Our work advances the frontier of several lines of research.

\paragraph{Contextual optimization. }
There has been a surge of interest in contextual stochastic optimization in the operations research community in recent years. This line of work studies how predictive models and optimization methods can be combined to improve decision-making under uncertainty.
The survey paper \citep{sadana2025survey} provides a comprehensive review of the literature on this topic, and classifies the existing works into three broad categories: decision rule optimization \citep{donti2017task,zhang2017assessing,ban2019big,bertsimas2022data,huber2019data}, sequential learning and optimization \citep{deng2022predictive,wang2026data} and integrated learning and optimization \citep{elmachtoub2022smart,qi2025integrated,loke2022decision}.
Our work falls within the third category. \cite{elmachtoub2022smart} introduce the SPO/SPO+ framework and formalize the idea that predictive accuracy alone may be poorly aligned with downstream decision quality.
\cite{el2019generalization} complement this line by establishing out-of-sample guarantees for decision-aware learning, while \cite{elmachtoub2020decision} develop interpretable decision trees trained directly for optimization performance. More recently, \cite{elmachtoub2023estimate} compare estimate-then-optimize, integrated-estimation-optimization, and sample average approximation, helping clarify the relative strengths of these paradigms.
Recent work also seeks to improve the computational scalability of decision-focused training, e.g., \cite{wan2026solverfree} construct a solver-free surrogate for contextual linear programming.
Among these works, \cite{qi2025integrated} is closest to ours. They propose the integrated conditional estimation-optimization framework, which learns the conditional distribution of the uncertain outcome and evaluates it through the downstream decision risk of the induced plug-in policy.
They show that this framework enjoys strong statistical guarantees. Our paper builds on this perspective and studies how a weak-to-strong training pipeline can be incorporated when labeled data are scarce but unlabeled contexts are abundant.
Interested readers can also refer to the survey papers \cite{qi2022integrating} and \cite{sadana2025survey}.

\paragraph{Weak-to-strong generalization. }
\cite{burns2024weak} first formalize weak-to-strong generalization as the empirical phenomenon that a strong pre-trained model, finetuned on labels produced by a much weaker model, can outperform its weak supervisor across tasks (NLP benchmarks, chess puzzles, and reward modeling).
\cite{lang2025debate} show that a debate stage, where strong models generate competing answers and a weak supervisor selects between them, improves pseudo-label quality.
Interestingly, \cite{goel2025great} find that as models become more capable they make increasingly correlated mistakes, which can reduce the benefits of W2S training and pose risks for AI oversight because similar models have less complementary knowledge to transfer.

There is also a line of works focusing on theoretical understanding of W2S generalization.
Our work adopts the perspective of \cite{dong2025discrepancies} and \cite{liu2025does}.
They propose a discrepancy-based framework to analyze the W2S generalization phenomenon.
They find that W2S gains arise from variance reduction in low-dimensional finetuning.
When teacher and student rely on different feature directions, the teacher's errors appear as noise to the student and can be averaged out with many pseudo-labels. Hence, moderate discrepancy can improve W2S.
A relevant intuition is also provided in \cite{charikar2024quantifying}, where they prove that the gain of W2S comes from the misfit between the weak and the strong model, i.e., the erroneous knowledge of the strong model is not inherited from the weak model.
Other theoretical explanations include \cite{lang2024theoretical}, who focuses more on a geometry perspective.
They show that under an expansion condition \citep{cai2021theory} on the data graph, any student classifier that agrees with a large set of teacher mistakes must also disagree on many neighboring points (collateral mistakes), making it impossible to maintain low error while preserving those mistakes. Existing theoretical studies thus characterize W2S primarily through prediction error, whereas we study W2S through the downstream decision risk of the induced policy.

We note that the W2S generalization is also closely related to broader paradigms of learning from imperfect supervision, such as semi-supervised learning  \citep{zhu2005semi,yang2022survey}, where unlabeled samples are used to improve learning when labels are scarce.
Within this broad regime, existing methods differ in how they extract information from the unlabeled dataset.
Pseudo-labeling and self-training augment the labeled sample with model-generated labels for unlabeled observations, with self-training iteratively updating the model using its own confident predictions \citep{lee2013pseudo,weitheoretical2021}.
Co-training instead uses predictors based on distinct views to provide labels for one another \citep{blum1998combining,ma2020selfpaced,lang2022cotrain}.
Other methods exploit the geometry of the unlabeled inputs: graph-based methods propagate labels across a similarity graph \citep{zhu2003semi,iscen2019label}, whereas entropy minimization encourages confident predictions and thereby favors decision boundaries in low-density regions \citep{grandvalet2004semi,berthelot2019mixmatch}.
Emerging more recently, semi-supervised methods with side information use helper covariates to improve the pseudo-responses used to train a deployable predictor based only on the primary covariates \citep{xia2024prediction,shi2026coupled}.
Beyond these algorithmic developments, theoretical work studies when unlabeled data can reduce sample complexity \citep{ben2008does,wegel2025sample} or improve robustness \citep{carmon2019unlabeled}. Related connections include other forms of weak supervision \citep{zhang2025weakly} and teacher--student transfer through knowledge distillation \citep{yang2025survey}.
Unlike these prediction-oriented methods, our W2S framework leverages weak-to-strong supervision directly for downstream optimization problem, and proves that this supervision achieves lower decision risk when the representations of the two models are different enough.
For broader background on teacher-student transfer mechanisms that overlap with W2S, we refer readers to a survey on knowledge distillation for LLMs \citep{yang2025survey}.

The semi-supervised methods above extract training signals from a fixed labeled–unlabeled sample, while active learning uses unlabeled observations to determine where additional ground-truth labels should be acquired.
This information-acquisition perspective has also been studied in contextual linear optimization recently.
For example, \citet{liu2023active} query the cost vector of an arriving context when the current prediction lies near degeneracy of the downstream linear program.
\citet{wan2026directional} instead select among candidate experimental designs using directional disagreement among plausible cost predictors and employ inverse-propensity weighting to account for the resulting adaptive sampling.
Our W2S framework likewise operates on a fixed labeled--unlabeled sample but does not query or acquire additional ground-truth labels.

\paragraph{Organization.}

In Section~\ref{sec:setting}, we introduce the problem setting and the W2S framework.
Section~\ref{sec:upper_bound_w2s} presents the main W2S decision-risk upper bound and its proof.
Section~\ref{sec:comparison_bounds} develops the benchmark and teacher-side estimates needed to interpret the main theorem, and uses them in a W2S performance case study.
Section~\ref{sec:numerical_experiments} presents numerical experiments on synthetic and text-based operational data to validate our theoretical findings.

\paragraph{Notation.}
We use $\lesssim$ to denote inequality up to a universal constant factor, independent of any problem parameters.
For a positive integer $m$, let $[m]\defeq\braces{1,\ldots,m}$.
For a vector $a$, we use either $a_k$ or $[a]_k$ to denote its $k$-th entry.
For a symmetric matrix $A$, we write $A\succeq0$ if $A$ is positive semidefinite.
Expectation subscripts indicate the source of randomness; for example, $\EE{x}{\cdot}$ denotes expectation with respect to $x$.
For any vector $\eta \in \mathbb{R}^K$, we write $\operatorname{softmax}(\eta) \in \Delta(\Xi)$ with $k$th coordinate
\(
\operatorname{softmax}_k(\eta)
=
\frac{\exp(\eta_k)}{\sum_{j=1}^K \exp(\eta_j)}, ~ k \in [K].
\)

\section{The Setting} \label{sec:setting}

In Section~\ref{subsec:overview}, we first overview our decision making problem at a high level.
Then in Section~\ref{subsec:w2s_framework}, we explain in detail our W2S training framework.

\subsection{Problem Setup}
\label{subsec:overview}

    \paragraph{Contextual Stochastic Optimization.} We consider a convex contextual stochastic optimization framework, which arises in many operations research applications.
    The feasible region for the decision variable $w$ is a convex set $\calA \subseteq \mathbb{R}^d$.
    The form of the cost function $c(\cdot, \xi): \mathcal{A} \rightarrow \mathbb{R}$ is fixed and known to the decision maker; while $c$ also depends on a random parameter $\xi \in \Xi$.
    In this work, we consider the case where the random parameter $\xi$ has finite discrete support, namely, $\Xi = \braces{z_1, \cdots, z_K}$.
    The realized outcome $\xi$ is not known at the time of decision-making.
    However, we assume that the decision-maker has access to a context vector $x \in \mathcal{X} \subseteq \mathbb{R}^{d_x}$ at the time of decision-making, and the goal is to make a decision $w$ based on this context.
    We denote by $\calD$ the joint distribution of $(x,\xi)$, and by $\mathcal{D}_x$ the marginal distribution of $x$.
    To emphasize when $x$ denotes a random variable, rather than its realization, we write $x$ in the subscript of the expectation operator.

    Formally, the decision maker's goal is, when given a context vector $x$, to solve the contextual stochastic optimization problem:
    \begin{equation} \label{eq:problem}
        \min_{w \in \calA} ~ \EE{\xi}{ c(w, \xi) \mid x } ,
    \end{equation}
    where the expectation is taken with respect to the conditional distribution of $\xi$ given $x$.
    Hence, a policy $\pi: \mathcal{X} \rightarrow \calA$ is a function that maps each context to a decision.

    \paragraph{The labeled and unlabeled datasets.} Since the conditional distribution of $\xi$ given $x$ is unknown, it must be learned from data.
   We consider a setting in which two datasets are available: one labeled data set $\tilde{\mathcal{B}} = \braces{ \tilde{x}_i, \tilde{\xi}_i }_{i=1}^{n}$ of size $n$, and one unlabeled dataset  $\mathcal{B} = \left\{x_j\right\}_{j=1}^N$ of size $N$.
    We assume $N$ is much larger than $n$, as is common in practice.
    The labeled dataset $\tilde{\calB}$ is generated i.i.d. according to the ground-truth joint distribution $\calD$, while the unlabeled dataset $\calB$ is generated i.i.d. according to the ground-truth marginal distribution $\calD_x$ of contexts, and hence $\calB$ is independent of $\tilde{\calB}$.

    \paragraph{The two feature mappings. }    A central question, then, is how to leverage the large unlabeled dataset to improve decision quality.
    Inspired by \cite{dong2025discrepancies}, we consider two (pre-trained) features that process the same context in different ways.
    Concretely, let us fix a common feature dimension $d_\phi$. We consider two feature mappings (a.k.a. models) from the context space to this common feature space: a \textit{strong} one $\phi_{\rms}: \mathcal{X} \rightarrow \mathbb{R}^{d_{\phi}}$ and a \textit{weak} one $\phi_{\rmw}: \mathcal{X} \rightarrow \mathbb{R}^{d_{\phi}}$.
    Given a parameter matrix $\Theta \in \mathbb{R}^{d_\phi \times K}$, these feature mappings can induce two parametric conditional models $P_{\Theta}^{\rms}, P_{\Theta}^{\rmw}: \mathcal{X} \rightarrow \Delta(\Xi)$, obtained by applying a softmax to the logits $\Theta^\top \phi_{\rms}(x)$ and $\Theta^\top \phi_{\rmw}(x)$, respectively, where $\Delta(\Xi)$ is the set of all probability distributions over $\Xi$.
    In our framework, the strong feature mapping $\phi_{\rms}(\cdot)$ is required to be more \textit{expressive} in the sense that it can reproduce any representation induced by the weak feature mapping $\phi_{\rmw}(\cdot)$.
    We will make it precise in Section~\ref{subsec:setting_upper_bound}.

    In practice, the two feature mappings can arise from different pre-trained models available to the decision maker.
    By \textit{pre-training}, we mean training a feature mapping $\phi$ on a generic objective before adapting it to a downstream task (e.g., language models trained on large text corpora using next-token prediction).
    For example, a smaller model trained with limited resources and a larger model trained with more data, compute, or capacity may process the same context but encode different information.
    In our framework, we abstract these pre-trained representations as fixed feature mappings $\phi_{\rmw}$ and $\phi_{\rms}$, and study how the downstream W2S procedure uses limited labeled data and abundant unlabeled contexts to adapt them for decision-making.

    We now introduce the notion of \textit{correlation dimension}, which turns out to be a key quantity that characterizes the W2S phenomenon in our setting.
    We denote $\Sigma_{\rmw} \defeq \EE{x \sim \mathcal{D}_x}{ \phi_{\rmw}(x) \phi_{\rmw}(x)^\top }$ and $\Sigma_{\rms} \defeq \EE{x \sim \mathcal{D}_x}{ \phi_{\rms}(x) \phi_{\rms}(x)^\top }$.

    \begin{definition}[intrinsic and correlation dimensions]
        We let
    $
        d_{\rms} = \operatorname{rank}(\Sigma_{\rms}) \,\,\text{ and } \,\,d_{\rmw} = \operatorname{rank}(\Sigma_{\rmw})
    $
        be the \textit{intrinsic dimensions} of the strong and weak feature mappings, respectively.
        Consider spectral decompositions $\Sigma_{\rms} = V_{\rms} \Lambda_{\rms} V_{\rms}^{\top} \in \mathbb{R}^{d_\phi \times d_\phi}$ and $\Sigma_{\rmw} = V_{\rmw} \Lambda_{\rmw} V_{\rmw}^{\top} \in \mathbb{R}^{d_\phi \times d_\phi}$, where $\Lambda_{\rms} \in \mathbb{R}^{d_{\rms} \times d_{\rms}}$ and $\Lambda_{\rmw} \in \mathbb{R}^{d_{\rmw} \times d_{\rmw}}$ are diagonal matrices with positive eigenvalues in non-increasing order; while $V_{\rms} \in \mathbb{R}^{d_{\phi} \times d_{\rms}}$ and $V_{\rmw} \in \mathbb{R}^{d_{\phi} \times d_{\rmw}}$ consist of the corresponding orthonormal eigenvectors.
        We denote $\Sigma_{\rms}^{-\frac{1}{2}} = V_{\rms} \Lambda_{\rms}^{-\frac{1}{2}} V_{\rms}^{\top}$ and $\Sigma_{\rmw}^{-\frac{1}{2}} = V_{\rmw} \Lambda_{\rmw}^{-\frac{1}{2}} V_{\rmw}^{\top}$.
        Let
        \begin{eqnarray}
            d_{\rms \wedge \rmw} \defeq \left\| \Sigma_{\rms}^{-\frac{1}{2}} \EE{ x \sim \mathcal{D}_x }{ \phi_{\rms}(x) \phi_{\rmw}(x)^{\top} } \Sigma_{\rmw}^{-\frac{1}{2}}  \right\|_F^2
        \end{eqnarray}
        be the \textit{correlation dimension} between $\phi_{\rms}$ and $\phi_{\rmw}$ such that $0 \leqslant d_{\rms \wedge \rmw} \leqslant \min \left\{d_{\rms}, d_{\rmw}\right\}$.
    \end{definition}
The correlation dimension measures the alignment between the strong and weak features. Intuitively, a larger correlation dimension indicates that the two features share more common information.
The fact that $0 \leqslant d_{\rms \wedge \rmw} \leqslant \min \left\{d_{\rms}, d_{\rmw}\right\}$ is proved in Lemma~\ref{lemma:overlap_dimension_range} in the Appendix.

    \paragraph{Decision-making pipeline.}
    To study the decision-making pipeline induced by these feature-based models, we next specify how a conditional model over outcomes generates a decision.
    For any conditional model $P: \mathcal{X} \rightarrow \Delta(\Xi)$, we denote by $\tilde{w}: \Delta(\Xi) \rightarrow \calA$ the oracle policy (i.e., plug-in policy) that selects the optimal action for each given context $x$
    \begin{equation} \label{eq:policy_oracle}
        \tilde{w}( P(x))
        \in \argmin _{w \in \mathcal{A}} ~ \mathbb{E}_{\xi \sim P(x) }[c(w, \xi)\,|\,x] .
    \end{equation}
For notational convenience, when $P$ is a conditional model, we also write $\tilde{w}(P\mid x)$ for $\tilde{w}(P(x))$, where $P(x)=P(\cdot\mid x)\in\Delta(\Xi)$.
Such a decision pipeline is studied, for example, in \cite{qi2025integrated}, where the induced empirical risk minimization (ERM) training is shown to enjoy favorable statistical guarantees in terms of decision risk.

\paragraph{Performance measure. }

    Given a policy $\pi: \mathcal{X} \rightarrow \calA$, we define its expected decision risk to be
    \begin{equation}
        R(\pi) = \mathbb{E}_{x,\xi \sim \mathcal{D} } \brackets{ c ( \pi(x) , \xi) } .
    \end{equation}
    Here, $\EE{x, \xi \sim \calD}{\cdot}$ denotes the expectation with respect to $(x, \xi) \sim \mathcal{D}$.
    The benchmark decision risk is the optimal decision risk achieved by the oracle policy induced by the ground-truth conditional distribution $P^*$:
    \begin{equation} \label{eq:benchmark_decision_risk}
        R^* = \mathbb{E}_{x,\xi \sim \mathcal{D} } \brackets{ c ( \tilde{w} ( P^* ( x ) ) , \xi) } ,
    \end{equation}
    where $P^*(x) \in \Delta(\Xi)$ denotes the ground-truth conditional distribution of $\xi$ given $x$.
    It is shown in \cite{qi2025integrated} that the ground-truth conditional distribution $P^*$ is the distribution that yields the lowest value for expression~\eqref{eq:benchmark_decision_risk}.

\subsection{The W2S Framework}
\label{subsec:w2s_framework}

In the previous section, we have introduced the two feature mappings $\phi_{\rms}(\cdot)$ and $\phi_{\rmw}(\cdot)$, acquired from pre-training.
However, they are not yet specialized to the downstream decision problem.
Next, we will explain how W2S pipeline can be used to adapt these two representations for the downstream decision problem.

    At a high level, the W2S framework first trains a weak model $P_{\widehat{\Theta}_{\rmw}}^{\rmw}$ on the labeled dataset $\tilde{\calB}$, and then uses its predicted outcome distributions as {\it supervision} to train a strong model $P_{\widehat{\Theta}_{\mathrm{w2s}}}^{\rms}$ on the unlabeled dataset $\calB$.
    The resulting strong model then induces a plug-in decision policy $ x \mapsto \tilde{w}\paren{ P_{\widehat{\Theta}_{\mathrm{w2s}}}^{\rms} (x) }  $.

\paragraph{Weak model.}
    Within our W2S framework, a weak model $P_{ \Theta}^{\mathrm{w}}: \mathcal{X} \to \Delta(\Xi) $ is used to generate pseudo-distributions for the unlabeled data. We do not impose a specific (post-)training procedure for this weak model, except that it must be constructed solely from the labeled dataset $\tilde{\calB}$ and based on the weak feature map $\phi_{\rmw}(\cdot)$. We assume that its estimation error for the ground-truth model parameter is controlled in mean squared error, in a sense to be specified precisely below. One natural way to obtain such a weak model is via maximum likelihood estimation (MLE).

\paragraph{W2S model.}

Given a weak model $P_{ \widehat{\Theta}_{\mathrm{w}}}^{\mathrm{w}} $ trained on the labeled dataset $\tilde{\calB}$, the W2S framework trains the strong model by solving the following optimization problem:
\begin{equation} \label{eq:w2s_optimization_procedure}
    \widehat{\Theta}_{\mathrm{w2s}} \in \argmin_{ \Theta \in \mathbb{R}^{d_{\phi} \times K}, \norm{\Theta}_F \leq B } \frac{1}{N}
    \sum_{j=1}^{N}
    \mathbb{E}_{ \xi \sim P_{ \widehat{\Theta}_{\mathrm{w}}}^{\mathrm{w}} (\cdot \mid x_j ) }
    \left[ c \left(\tilde{w} \left( P^{\mathrm{s}}_\Theta  ( x_j ) \right), \xi \right) \right] ,
\end{equation}
where $B > 0$ is some constant.
Here, $P_{ \widehat{\Theta}_{\mathrm{w}}}^{\mathrm{w}} (\cdot \mid x_j )$ denotes the conditional distribution over $\xi$ given $x_j$, induced by the weak model.
Thus, rather than providing only a scalar prediction, the weak model quantifies the likelihood of each possible outcome in a given context. This conditional distribution serves as a teacher signal for the strong model. The strong model then leverages this ``soft'' information to learn a decision rule that performs well across the outcomes deemed plausible by the teacher, thereby extracting supervision from unlabeled contexts.

We let
\begin{equation} \label{eq:w2s_policy}
    \widehat{ \pi }_{\mathrm{w2s}}(x) \defeq \tilde{w}\left(P_{\widehat{\Theta}_{\mathrm{w2s}}}^{\mathrm{s}} (x) \right)
\end{equation}
be the policy induced by the W2S training framework.
We measure its generalization error via the expected decision risk over both $\tilde{\calB}$ and $\calB$.
Namely,
\begin{equation}
    \EE{ \tilde{\calB}, \calB }{ R( \widehat{ \pi }_{\mathrm{w2s}} ) - R^* } .
\end{equation}
Here, the expectation is taken with respect to the randomness from both the labeled data set $\tilde{\calB}$ and the unlabeled data set $\calB$.

\paragraph{Strong model.}
    A natural benchmark is to train the strong model directly on the labeled sample $\tilde{\calB}$, as empirical scaling laws motivate using stronger pre-trained models when sufficient data and compute are available \citep{kaplan2020scaling,hoffmann2022training,hernandez2021scaling}.
For the downstream decision-making problem, we can train the strong model by directly minimizing the empirical decision risk on the labeled data:
\begin{equation} \label{eq:strong_training_procedure}
    \widehat{\Theta}_{\mathrm{s}} \in \argmin _{\Theta \in \mathbb{R}^{d_{\phi} \times K} }
        \frac{1}{n} \sum_{i=1}^{n} c \left( \tilde{w}\left(P_{\Theta}^{\mathrm{s}} \paren{ \tilde{x}_i }\right), \tilde{\xi}_i \right) .
\end{equation}
Let $\widehat{ \pi }_{\mathrm{s}}(x):=\tilde{w}\left(P_{\widehat{\Theta}_{\mathrm{s}}}^{\mathrm{s}} (x) \right)$ be the policy induced by the strong model trained on labeled data only. We measure its generalization error via the expected decision risk over $\tilde{\calB}$ only. Namely,
\begin{equation}
    \EE{ \tilde{\calB} }{ R( \widehat{ \pi }_{\mathrm{s}} ) - R^* } .
\end{equation}

\paragraph{Metric for W2S performance.}

We use the \underline{o}ut\underline{p}erforming \underline{r}atio (OPR) to compare the expected excess decision risk of the strong model and the W2S model:
\begin{equation} \label{eq:opr_def}
    \operatorname{OPR} = \frac{ \EE{ \tilde{\calB} }{ R( \widehat{ \pi }_{\mathrm{s}} ) - R^* }   }{ \EE{ \tilde{\calB}, \calB }{ R( \widehat{ \pi }_{\mathrm{w2s}} ) - R^* } } .
\end{equation}
A higher OPR implies better performance of the W2S model compared to the strong model: $\widehat{\pi}_{\mathrm{w2s}}$ outperforms $\widehat{\pi}_{\mathrm{s}}$ when $\operatorname{OPR} > 1$.

\section{Analysis of the W2S Framework}
\label{sec:upper_bound_w2s}
This section presents the main theoretical result of the paper: a non-asymptotic upper bound on the expected decision risk of the W2S framework. We first state the structural assumptions and then prove the bound by decomposing the W2S risk into imitation, generalization, teacher-estimation, and approximation terms.

\subsection{Blanket assumptions}
\label{subsec:setting_upper_bound}

\paragraph{ Underlying problem structure }
We use $P^*(\cdot \mid x)$ to denote the ground-truth conditional distribution of $\xi$ given $x$.
Throughout this work, we make the following assumption on realizability.

\begin{assumption}[realizability] \label{assumption:realizability}
    There exists a  measurable logit mapping $\eta^{*}: \mathcal{X} \rightarrow \mathbb{R}^K$ such that
    \begin{equation}
        P^* ( \xi = z_k \mid x ) = \frac{\exp \left(\eta^*_{k}(x)\right)}{\sum_{j=1}^K \exp \left(\eta^*_{j}(x)\right)}, \quad k=1, \ldots, K .
    \end{equation}
    In particular, the ground-truth conditional probabilities are represented by finite softmax logits, and we assume $\E{\norm{\eta^*(x)}_2^4}<\infty$.
    Moreover, without loss of generality, we assume that $\eta^{*}(x)$ is centered, i.e., $\eta^{*}(x)^\top \bfone =0$ for all $x \in \mathcal{X}$.
\end{assumption}
Based on the realizability assumption, we hence consider a particular parametric model for the conditional distributions.
Given a feature mapping $\phi: \mathcal{X} \rightarrow \mathbb{R}^{d_\phi}$, we let $\Theta \in \mathbb{R}^{d_\phi \times K}$ be the model parameter, and define the conditional model $P_{\Theta} $ to be
\begin{eqnarray}
P_{\Theta} (\xi= z_k \mid x) &=& \frac{\exp \left([\eta_{\Theta}(x)]_k\right)}{\sum_{k'=1}^K \exp \left( [\eta_{\Theta}(x)]_{k'} \right)}, ~ \text{for} ~ k=1, \ldots, K  \label{eq:softmax_feature} .
\end{eqnarray}
with linear feature mapping $\eta_{\Theta}(x) = \Theta^{\top} \phi(x) \in \mathbb{R}^K$.
Note that linear feature mapping is commonly used in literature and it represents a broad class of models \citep{jacot2018neural, woodworth2020kernel, aouad2025sign}.

\medskip

Moreover, we make the following assumption on the cost function.
\begin{assumption} [cost function] \label{assumption:cost_function}
    \begin{enumerate}
        \item For any action $a \in \mathcal{A}$, and outcome $z_k \in \Xi$, the cost function $c(a, z_k)$ is $\mu$-strongly convex and $L$-smooth in the first argument, i.e.,
    $$\mu I \preceq \nabla_a^2 c\left(a, z_k\right) \preceq L I.$$
        \item  Let $\mathcal{W}=\operatorname{conv}\{\tilde{w}(p):p\in\Delta(\Xi)\}$ be the convex hull of all plug-in decisions. The gradients of the scenario costs are uniformly bounded on $\mathcal{W}$: for all $a\in\mathcal{W}$ and $k\in[K]$,
        $$\normtwo{\nabla_a c(a,z_k)}\leq G.$$
        In particular, since $\mathcal{W}$ is convex, for all $w,w'\in\mathcal{W}$ and $k\in[K]$,
        $$\abs{c(w,z_k)-c(w',z_k)}\leq G\normtwo{w-w'}.$$
    \end{enumerate}
\end{assumption}
While the strong convexity assumption makes the analysis cleaner and more tractable, it is not essential regarding the outperforming phenomenon of W2S.
It can be relaxed to convexity with some additional technical efforts, at the cost of slightly worse rates. Moreover, in Section~\ref{sec:numerical_experiments}, we show numerically that W2S outperforms the strong model even when the cost function is not strongly convex.

We consider the feasible region to be the whole space, i.e., $\calA = \mathbb{R}^d$.
This assumption is not essential regarding the outperforming phenomenon of W2S.
It can be relaxed to a bounded convex set with some additional technical efforts.

\medskip

\paragraph{ The two models. }
Recall that we are given two feature mappings, a strong one $\phi_{\rms}: \mathcal{X} \rightarrow \mathbb{R}^{d_{\phi}}$ and a weak one $\phi_{\rmw}: \mathcal{X} \rightarrow \mathbb{R}^{d_{\phi}}$ with different model capacities.
We can define the \textit{approximation error} for the strong and weak feature model, respectively. Namely, we define
\begin{equation} \label{eq:approximation_error_s}
    \rho_{\rms} = \min_{ \Theta \in \mathbb{R}^{ d_{\phi} \times K}, ~ \Theta \bfone = 0 }
    \EE{x \sim \calD_x}{ \norm{ \Theta^{\top} \phi_{\rms}(x) - \eta^{*}(x) }_{2}^{2} } ,
\end{equation}
with $\Theta^{*}_{\rms} $ being a minimizer.
Similarly, we define
\begin{equation} \label{eq:approximation_error_w}
    \rho_{\rmw} = \min_{ \Theta \in \mathbb{R}^{ d_{\phi} \times K}, ~ \Theta \bfone = 0 }
    \EE{x \sim \calD_x}{ \norm{ \Theta^{\top} \phi_{\rmw}(x) - \eta^{*}(x) }_{2}^{2} },
\end{equation}
with $\Theta^{*}_{\rmw} $ being a minimizer.

We introduce some notation for the two models.
For a parameter $\Theta$, we denote the logits of the two feature mappings by $\eta_{\Theta}^{\mathrm{s}}(x)=\Theta^{\top} \phi_{\mathrm{s}}(x) \in \mathbb{R}^K$ and $\eta_{\Theta}^{\mathrm{w}}(x)=\Theta^{\top} \phi_{\mathrm{w}}(x) \in \mathbb{R}^K$.  In addition, we denote $P^{\rms}_{\Theta}(\cdot \mid x)$ $\defeq \operatorname{softmax}\paren{\Theta^{\top}\phi_{\rms}(x)}$ (and $P^{\rmw}_{\Theta}(\cdot \mid x)$) to be the conditional model obtained by composing softmax with $\eta_{\Theta}^{\mathrm{s}}(x)$ (and $\eta_{\Theta}^{\mathrm{w}}(x)$), in the same way as in \eqref{eq:softmax_feature}.
We denote $\calH_B \defeq \braces{ \Theta \in \mathbb{R}^{d_{\phi} \times K}, ~ \norm{\Theta}_F \leq B  }.$

We then quantify how well the strong feature class can reproduce the logits induced by the learned weak model. Specifically, given a weak model estimator $\widehat{\Theta}_{\mathrm{w}}$, we define the weak-to-strong logit approximation error to be
\begin{equation} \label{eq:student_teacher_logit_approximation_error}
    \rho_{\rms \rightarrow \rmw} \paren{\widehat{ \Theta}_{\mathrm{w}} }
    \defeq \inf _{\Theta \in \mathcal{H}_B} \mathbb{E}_x\left\|\Theta^{\top} \phi_{\mathrm{s}}(x)-\eta_{\widehat{\Theta}_{\mathrm{w}}}^{\mathrm{w}}(x)\right\|_2^2 .
\end{equation}
Namely, it measures the extent to which the strong feature representation can emulate the weak model at the logit level.
As we will see in Theorem~\ref{thm:upper_bound_w2s}, we need the strong model to be more expressive than the weak model in the sense that $\rho_{\rms \rightarrow \rmw} \paren{\widehat{ \Theta}_{\mathrm{w}} }$ is small.

To facilitate the analysis, we consider the following least-squares formulation to train the strong model based on the pseudo-labels (more precisely, pseudo-distributions over outcomes) generated by the weak model:
\begin{equation} \label{eq:least_squares}
    \widehat{\Theta}_{\mathrm{ls}} \in \argmin_{ \Theta \in \mathbb{R}^{d_{\phi} \times K} }
    \frac{1}{N} \sum_{j=1}^{N} \normtwo{  {\Theta}_{}^{\top} \phi_{\mathrm{s}} (x_j)  - \widehat{\Theta}_{\mathrm{w}}^{\top} \phi_{\mathrm{w}} (x_j)   }^2 .
\end{equation}

\begin{assumption} \label{assumption:bounded_ls_theta}
    The least-squares estimate $\widehat{\Theta}_{\mathrm{ls}}$ is bounded in Frobenius norm, i.e., $\norm{ \widehat{\Theta}_{\mathrm{ls}} }_F \leq B$ almost surely.
\end{assumption}
Assumption~\ref{assumption:bounded_ls_theta} is made merely for technical convenience.
It allows us to place the least-squares estimator $\widehat{\Theta}_{\mathrm{ls}}$ in the bounded class $\calH_B$, which simplifies the empirical process arguments used in the proof.
The assumption is not central to the weak-to-strong effect itself, and could be relaxed with additional technical efforts.
For example, we can show that $\widehat{\Theta}_{\mathrm{ls}}$ is bounded with high probability when the features are subgaussian.

\medskip

Moreover, we need the following assumption on the feature mappings, which controls the tail behavior of the features.
\begin{assumption}[subgaussian features] \label{assumption:feature_joint_subgaussian}
    Let $\phi(x) = [\phi_{\rms}(x) ; \phi_{\rmw}(x)] \in \mathbb{R}^{2 d_{\phi}}$ be the concatenation of $\phi_{\rms}(x)$ and $\phi_{\rmw}(x)$, with covariance matrix $\Sigma_{\phi}$.
    There exists a constant $c > 0$ such that for any $u \in \mathbb{R}^{2 d_{\phi}}$, we have
    \begin{equation} \label{eq:phi_joint_subgaussian}
        \EE{x}{ \exp \paren{  u^{\top} \phi(x)  } } \leq \exp \paren{ \frac{1}{2} c^2 u^{\top} \Sigma_{\phi} u }   .
    \end{equation}
\end{assumption}
This notion of subgaussianity is commonly used when it comes to covariance estimation. See, for example, Theorem 4.7.1 and Theorem 9.2.4 in \citep{vershynin2018high}, as well as \cite{lobo2024fair}.
The zero-mean assumption (implied by \eqref{eq:phi_joint_subgaussian}) is merely for technical convenience, as we can always center the features without loss of generality.

\subsection{W2S decision-risk upper bound}

Now we are ready to present the main result of our work, which is a non-asymptotic upper bound on the expected decision risk of the W2S framework.

\begin{theorem} \label{thm:upper_bound_w2s}
    Under Assumptions~\ref{assumption:realizability}, \ref{assumption:cost_function}, \ref{assumption:bounded_ls_theta},  \ref{assumption:feature_joint_subgaussian},
    whenever $ N \gtrsim d_{\rms} $, we have
    \begin{eqnarray}
        && \EE{ \tilde{\calB}, \calB }{ R( \widehat{ \pi }_{\mathrm{w2s}} ) - R^* } \nonumber \\
        &\lesssim& \paren{ \frac{ L^2 K G^2  }{\mu^3} + \frac{L K G^2}{\mu^2} }
        \cdot
        \Bigg (
            \mathbb{E}_{\tilde{\mathcal{B}}}\left[ \rho_{\rms \rightarrow \rmw} \paren{\widehat{ \Theta}_{\mathrm{w}} } \right]  \nonumber \\
        &&
            + \frac{\paren{B_{\Theta}^{\rmw}}^2}{\sqrt{N}}\paren{\mathbb{E}_{x}\left\|\phi_{\rmw}(x)\right\|_2^4}^{1/2}
            +\frac{B^2}{\sqrt{N}} \paren{\mathbb{E}_x\left\|\phi_{\mathrm{s}}(x)\right\|_2^4}^{\frac{1}{2}}
            + \frac{B B_{\Theta}^{\rmw}}{\sqrt{N}}
            \paren{\mathbb{E}_{x}\left[\left\|\phi_{\rms}(x)\right\|_2^2\left\|\phi_{\rmw}(x)\right\|_2^2\right]}^{1/2}
        \Bigg ) \nonumber \\
        && + \frac{L K G^2 }{\mu^2} \cdot \Bigg (
         \frac{1}{\sqrt{N}}\paren{\EE{x}{ \norm{\eta^{*}(x)}_2^4 }}^{\frac{1}{2}} +
        B \sqrt{\tr{\Sigma_{\rms}}}   \frac{1}{ \sqrt{N} }
        + \frac{  B}{\sqrt{N}}   \paren{ \EE{x}{ \norm{\eta^{*}(x)}_2^4 } }^{\frac{1}{4}} \paren{ \EE{x}{ \norm{\phi_{\rms}(x)}_2^4 }  }^{\frac{1}{4}}
        \nonumber  \\
        && +
        \lambda_{\max} \paren{ \Lambda_{\rmw} }
        \mathbb{E}_{ \tilde{\mathcal{B}}  } \brackets{  \norm{ \widehat{\Theta}_{\mathrm{w}} - \Theta_{\mathrm{w}}^{*}  }_F^2 } \paren{ d_{\rms \wedge \rmw}+ \frac{1}{N} d_{\rms}\left(d_{\rmw}-d_{\rms \wedge \rmw}\right)  }
        + \rho_{\rmw}
        + \rho_{\rms}  \Bigg )
        + \frac{L C_{\mathrm{unif}}}{\mu} \frac{1}{\sqrt{N}} .
        \label{eq:main_w2s_decision_risk_bound}
	    \end{eqnarray}
\end{theorem}
We recall that $N$ is the number of unlabeled samples, $K$ is the support size of the outcome variable $\xi$, and $\mu, L, G$ are the curvature and gradient parameters from Assumption~\ref{assumption:cost_function}. The quantity $B$ is the F-norm bound on the least squares fit, and $d_{\rms \wedge \rmw}$ is the correlation dimension between the strong and weak features.
Here $C_{\mathrm{unif}} \defeq \max_{k \in [K]} \abs{ c\paren{\tilde{w}\paren{\operatorname{softmax}(0)},z_k} }$.

To interpret the upper bound, it is useful to decompose the bound into four parts:
\begin{enumerate}
    \item Imitation approximation error:  $\mathcal{E}_{\mathrm{imit}}
        \defeq
        \paren{ \frac{ L^2 K G^2  }{\mu^3} + \frac{L K G^2}{\mu^2} }
        \cdot
        \mathbb{E}_{\tilde{\mathcal{B}}}\brackets{\rho_{\rms \rightarrow \rmw}\paren{\widehat{\Theta}_{\mathrm{w}}}}$.
    It measures how well the strong feature class can reproduce the logits induced by the weak model.
    \item Unlabeled-sample statistical error:
    Let $\mathcal{E}_{N}$ denote the following unlabeled-sample statistical-error  term:
    {\small
    \[
    \begin{aligned}
        \mathcal{E}_{N}
        \defeq&
        \paren{ \frac{ L^2 K G^2  }{\mu^3} + \frac{L K G^2}{\mu^2} }
        \Bigg[
            \frac{\paren{B_{\Theta}^{\rmw}}^2}{\sqrt{N}}\paren{\mathbb{E}_{x}\left\|\phi_{\rmw}(x)\right\|_2^4}^{1/2}
            +\frac{B^2}{\sqrt{N}} \paren{\mathbb{E}_x\left\|\phi_{\mathrm{s}}(x)\right\|_2^4}^{\frac{1}{2}} \\
        &\qquad\qquad\qquad
            + \frac{B B_{\Theta}^{\rmw}}{\sqrt{N}}
            \paren{\mathbb{E}_{x}\left[\left\|\phi_{\rms}(x)\right\|_2^2\left\|\phi_{\rmw}(x)\right\|_2^2\right]}^{1/2}
        \Bigg] \\
        &+
        \frac{L K G^2 }{\mu^2}
        \Bigg[
            \frac{1}{\sqrt{N}}\paren{\EE{x}{ \norm{\eta^{*}(x)}_2^4 }}^{\frac{1}{2}}
            + B \sqrt{\tr{\Sigma_{\rms}}} \frac{1}{\sqrt{N}} \\
        &\qquad\qquad\qquad
            + \frac{B}{\sqrt{N}}
            \paren{\EE{x}{ \norm{\eta^{*}(x)}_2^4 }}^{\frac{1}{4}}
            \paren{\EE{x}{ \norm{\phi_{\rms}(x)}_2^4 }}^{\frac{1}{4}}
        \Bigg]
        + \frac{L C_{\mathrm{unif}}}{\mu\sqrt{N}} .
    \end{aligned}
    \]
    }
    This term captures the finite-sample fluctuation from replacing population expectations over $x$ with empirical averages over the unlabeled sample $\calB$; for fixed problem parameters, it decays at the rate $N^{-1/2}$.
    It comes from generalization error with respect to the unlabeled data.
    We denote $C_N  =  \mathcal{E}_{N} \sqrt{N} = \order{1} $.
    \item Weak-to-strong term: $\mathcal{E}_{\mathrm{teacher}}
        \defeq
        \frac{L K G^2}{\mu^2}
        \lambda_{\max} \paren{ \Lambda_{\rmw} }
        \mathbb{E}_{ \tilde{\mathcal{B}}  } \brackets{  \norm{ \widehat{\Theta}_{\mathrm{w}} - \Theta_{\mathrm{w}}^{*}  }_F^2 }
        \paren{ d_{\rms \wedge \rmw}+ \frac{1}{N} d_{\rms}\left(d_{\rmw}-d_{\rms \wedge \rmw}\right)  }$.
        It captures how the weak teacher's error propagates to the final W2S decision risk.
        The persistent component is governed by the overlap dimension $d_{\rms \wedge \rmw}$, whose contribution is not reduced by W2S training.
        A smaller overlap dimension can reduce the error inherited from the weak teacher and thereby create room for the student to outperform the teacher. Intuitively, when ways of thinking of the teacher and the student are more different, from the student's perspective the teacher's mistakes behave more like random noise than like a systematic bias.
        The remaining component $\frac{1}{N} d_{\rms}\left(d_{\rmw}-d_{\rms \wedge \rmw}\right)$ quantifies the residual contribution from the non-overlapping directions, and vanishes when the unlabeled sample size $N$ is sufficiently large.

    \item Model misspecification error: $\mathcal{E}_{\mathrm{approx}}
        \defeq
        \frac{L K G^2 }{\mu^2} \cdot \paren{ \rho_{\rmw} + \rho_{\rms} }$.
    It measures the approximation error of the strong and weak feature class, respectively.
    Unlike the unlabeled-sample statistical error, this term does not vanish with larger $N$ unless the feature classes themselves are enriched.
\end{enumerate}

\begin{remark}[No free lunch]
    At first glance, when $\mathcal{E}_{\mathrm{imit}}=\mathcal{E}_{\mathrm{approx}}=0$, the bound may seem to suggest that, if $d_{\rms \wedge \rmw}=0$, increasing $N$ can wash out the weak-teacher estimation error $\mathbb{E}_{ \tilde{\mathcal{B}}  } \brackets{  \norm{ \widehat{\Theta}_{\mathrm{w}} - \Theta_{\mathrm{w}}^{*}  }_F^2 }$ and drive the W2S error arbitrarily low.
    This is not the case; the formal statement and proof are given in Appendix~\ref{appendix:proof_lemma_case_study_overlap_lower_bound}.
\end{remark}

\subsection{Proof of Theorem~\ref{thm:upper_bound_w2s}}
This section is devoted to the proof of Theorem~\ref{thm:upper_bound_w2s}. Proofs of the supporting technical lemmas are deferred to the appendix.

To start off the analysis, we first present a useful result, which is a consequence of the strong convexity and smoothness of the cost function.

\begin{lemma} \label{lemma:convex_smooth_expected_cost}
    Under Part (1) of Assumption~\ref{assumption:cost_function},
    consider two conditional distributions $P, P': \mathcal{X} \rightarrow \Delta(\Xi)$.
    For any $x$, we have
    \begin{equation}
        \frac{\mu}{2} \norm{ \tilde{w} ( P' \mid x ) - \tilde{w} ( P \mid x ) }_2^2
    \leq
    \mathbb{E}_{ \xi \sim P  (\cdot \mid x ) } \brackets{
        c \paren{ \tilde{w}\left( P' \mid x \right), \xi }
        - c \paren{ \tilde{w} \paren{ P \mid x } , \xi }
    }
    \leq \frac{L}{2} \norm{ \tilde{w} ( P' \mid x ) - \tilde{w} ( P \mid x ) }_2^2  .
    \end{equation}
\end{lemma}
The proof appears in Appendix~\ref{appendix:proof_lemma_convex_smooth_expected_cost}.
Moreover, the following observation is also useful for the analysis.
\begin{lemma} \label{lemma:decision_to_logit_continuity}
    Under Assumption~\ref{assumption:cost_function},
    given two logits $\eta, \eta': \mathcal{X} \rightarrow \mathbb{R}^K$, let $P, P'$ be the corresponding softmax distributions, namely $P_{\eta}(\cdot \mid x)=\operatorname{softmax}(\eta(x))$ and $P_{\eta'}(\cdot \mid x)=\operatorname{softmax}(\eta'(x))$.
    For any given $x \in \mathcal{X}$, we have
    \begin{equation}
        \norm{ \tilde{w}(P_{\eta} \mid x ) - \tilde{w}(P_{\eta'} \mid x ) }_2
        \leq \frac{G \sqrt{K}}{2 \mu} \norm{ \eta(x) - \eta'(x) }_2 .
    \end{equation}
\end{lemma}
The proof appears in Appendix~\ref{appendix:proof_lemma_decision_to_logit_continuity}.

\medskip

Consider a policy $\pi = \tilde{w} \circ P$, where $P: \mathcal{X} \rightarrow \Delta(\Xi)$ is some conditional model.
In view of Lemma~\ref{lemma:convex_smooth_expected_cost}, we conclude that
\begin{eqnarray}
    R( \pi ) - R^* &=& \mathbb{E}_{x} \brackets{ \mathbb{E}_{\xi \mid x} \brackets{
        c ( \tilde{w} ( P  \mid x ) , \xi) - c ( \tilde{w} ( P^* \mid x ) , \xi)
        } } \\
    &=& \mathbb{E}_{x} \brackets{ \sum_{k=1}^K P^* ( \xi = z_k \mid x ) \brackets{
        c ( \tilde{w} ( P  \mid x ) , z_k ) - c ( \tilde{w} ( P^* \mid x ) , z_k )
        } }  \nonumber  \\
    &\leq&  \frac{L}{2} \mathbb{E}_{x} \brackets{ \norm{ \tilde{w} ( P  \mid x ) - \tilde{w} ( P^* \mid x ) }_2^2 } .
\end{eqnarray}
Therefore, to control $\EE{ \tilde{\calB}, \calB }{ R( \widehat{ \pi }_{\mathrm{w2s}} ) - R^* }$, it suffices to upper bound $\EE{ \tilde{\calB}, \calB }{ \mathbb{E}_{x} \brackets{ \normtwo{
        \tilde{w} \paren{ P^{\mathrm{s}}_{\widehat{\Theta}_{\mathrm{w2s}}} \mid x  }
        - \tilde{w} \paren{ P^{*} \mid x  }
        }^2 } } $.
To this end, we recall that $\widehat{\Theta}_{\mathrm{ls}}$, defined in \eqref{eq:least_squares}, is obtained by the multivariate least-squares procedure trained using the strong model based on the pseudo-labels generated by the weak model.
We proceed to use $\widehat{\Theta}_{\mathrm{ls}}$ as an intermediate quantity for the analysis.
Hence, we consider the following decomposition
\begin{eqnarray}
    & & \mathbb{E}_{x} \brackets{ \normtwo{
        \tilde{w} \paren{ P^{\mathrm{s}}_{\widehat{\Theta}_{\mathrm{w2s}}} \mid x  }
        - \tilde{w} \paren{ P^{*} \mid x  }
        }^2 }    \nonumber  \\
    &\leq&  2\mathbb{E}_{x} \brackets{ \normtwo{
        \tilde{w} \paren{ P^{\mathrm{s}}_{\widehat{\Theta}_{\mathrm{w2s}}} \mid x  }
        - \tilde{w} \paren{ P^{\mathrm{s}}_{ \widehat{\Theta}_{\mathrm{ls}}} \mid x  }
        }^2 }
    + 2\mathbb{E}_{x} \brackets{ \normtwo{
        \tilde{w} \paren{ P^{\mathrm{s}}_{\widehat{\Theta}_{\mathrm{ls}}} \mid x  }
        - \tilde{w} \paren{ P^{*} \mid x  }
        }^2 }  . \label{eq:w_w2s_to_star_decomposition}
\end{eqnarray}
In what follows, we bound the two terms on the right-hand side separately.

\subsubsection*{Step 1: the first term in \eqref{eq:w_w2s_to_star_decomposition} .}

Let us focus on the first term $\mathbb{E}_{x} \brackets{ \normtwo{
        \tilde{w} \paren{ P^{\mathrm{s}}_{\widehat{\Theta}_{\mathrm{w2s}}} \mid x  }
        - \tilde{w} \paren{ P^{\mathrm{s}}_{ \widehat{\Theta}_{\mathrm{ls}}} \mid x  }
        }^2 }$.
Inside the norm is the difference between the actions recommended by two different training procedures (cf.~\eqref{eq:w2s_optimization_procedure} and \eqref{eq:least_squares}), both of which are based upon a given empirical weak teacher $P^{\mathrm{w}}_{\widehat{\Theta}_{\mathrm{w}}}$.
It turns out that this quantity can be controlled as the number of unlabeled samples $N$ increases, at a rate enjoyed by standard generalization bounds for empirical risk minimization.

The remainder of this step is devoted to proving the following bound.
    \begin{eqnarray}
        && \EE{ \calB, \tilde{\calB} }{ \mathbb{E}_{x} \brackets{ \normtwo{
        \tilde{w} \paren{ P^{\mathrm{s}}_{\widehat{\Theta}_{\mathrm{w2s}}} \mid x  }
        - \tilde{w} \paren{ P^{\mathrm{s}}_{ \widehat{\Theta}_{\mathrm{ls}}} \mid x  }
        }^2 } } \nonumber \\
        &\leq&  \paren{ \frac{1}{2}\frac{L K G^2  }{\mu^3}+\frac{1}{2}\frac{K G^2}{\mu^2} }
        \cdot
        \Bigg (
            \mathbb{E}_{\tilde{\mathcal{B}}}\left[\inf _{\Theta \in \mathcal{H}_B} \mathbb{E}_x\left\|\Theta^{\top} \phi_{\mathrm{s}}(x)-\eta_{\widehat{\Theta}_{\mathrm{w}}}^{\mathrm{w}}(x)\right\|_2^2\right]  \nonumber \\
        &&
            + \frac{2\paren{B_{\Theta}^{\rmw}}^2}{\sqrt{N}}\sqrt{\mathbb{E}_{x}\left\|\phi_{\rmw}(x)\right\|_2^4}
            +\frac{8 B^2}{\sqrt{N}} \paren{\mathbb{E}_x\left\|\phi_{\mathrm{s}}(x)\right\|_2^4}^{\frac{1}{2}}
            + \frac{16 B B_{\Theta}^{\rmw}}{\sqrt{N}} \sqrt{\mathbb{E}_{x}\left[\left\|\phi_{\mathrm{s}}(x)\right\|_2^2\left\|\phi_{\rmw}(x)\right\|_2^2\right]}
        \Bigg ) \nonumber \\
	        && + 16 \sqrt{2} \frac{ K G^2 B \sqrt{\tr{\Sigma_{\rms}}} }{ \mu^2}  \frac{1}{ \sqrt{N} }
	        + \frac{8 C_{\mathrm{unif}}}{\mu} \frac{1}{\sqrt{N}} .
	    \end{eqnarray}

First, using the fact that $\normtwo{a+b}^2 \leq 2\normtwo{a}^2 + 2\normtwo{b}^2$, we have
\begin{eqnarray}
    &&  \EE{ \calB, \tilde{\calB} }{ \mathbb{E}_{x} \brackets{ \normtwo{
        \tilde{w} \paren{ P^{\mathrm{s}}_{\widehat{\Theta}_{\mathrm{w2s}}} \mid x  }
        - \tilde{w} \paren{ P^{\mathrm{s}}_{ \widehat{\Theta}_{\mathrm{ls}}} \mid x  }
        }^2 } } \nonumber \\
    &\leq& 2 \EE{ \calB, \tilde{\calB} }{ \mathbb{E}_{x} \brackets{ \normtwo{
        \tilde{w} \paren{ P^{\mathrm{s}}_{\widehat{\Theta}_{\mathrm{w2s}}} \mid x  }
        - \tilde{w} \paren{ P^{\mathrm{w}}_{ \widehat{\Theta}_{\mathrm{w}}} \mid x  }
        }^2 } }
    + 2  \EE{ \calB, \tilde{\calB} }{
        \mathbb{E}_{x} \brackets{ \normtwo{
        \tilde{w} \paren{ P^{\mathrm{w}}_{\widehat{\Theta}_{\mathrm{w}}} \mid x  }
        - \tilde{w} \paren{ P^{\mathrm{s}}_{ \widehat{\Theta}_{\mathrm{ls}}} \mid x  }
        }^2 }  }  \label{eq:w2s_excess_risk_decomposition} .
\end{eqnarray}
It turns out that first term in \eqref{eq:w2s_excess_risk_decomposition} can be controlled by logits difference between the least-squares strong model and the weak model, proved in Lemma~\ref{lemma:w2s_action_btw_w2s_and_weak} in Appendix~\ref{appendix:proof_lemma_w2s_action_btw_w2s_and_weak}.
Namely, we have
\begin{eqnarray}
        && \EE{ \calB, \tilde{\calB} }{ \mathbb{E}_{x} \brackets{ \normtwo{
        \tilde{w} \paren{ P^{\mathrm{s}}_{\widehat{\Theta}_{\mathrm{w2s}}} \mid x  }
        - \tilde{w} \paren{ P^{\mathrm{w}}_{ \widehat{\Theta}_{\mathrm{w}}} \mid x  }
        }^2 } } \nonumber \\
        &\leq& \frac{1}{4} \frac{ L K G^2 }{\mu^3 }
        \EE{ \tilde{\calB} , \calB }{
            \mathbb{E}_{x} \brackets{
                \norm{ \eta^{\mathrm{s}}_{ \widehat{\Theta}_{\mathrm{ls}}} (x) - \eta^{\mathrm{w}}_{\widehat{\Theta}_{\mathrm{w}}} (x) }_2^2
            }
	        } + 8 \sqrt{2} \frac{ K G^2 B \sqrt{\tr{\Sigma_{\rms}}} }{ \mu^2}  \frac{1}{ \sqrt{N} }
	        + \frac{4 C_{\mathrm{unif}}}{\mu} \frac{1}{\sqrt{N}} \label{eq:w_w2s_w_to_eta_ls_w} .
	    \end{eqnarray}

Building on \eqref{eq:w2s_excess_risk_decomposition} and \eqref{eq:w_w2s_w_to_eta_ls_w}, and invoking Lemma~\ref{lemma:decision_to_logit_continuity}, we have
\begin{eqnarray}
    &&  \EE{ \tilde{\calB} , \calB }{  \mathbb{E}_{x} \brackets{ \normtwo{
        \tilde{w} \paren{ P^{\mathrm{s}}_{\widehat{\Theta}_{\mathrm{w2s}}} \mid x  }
        - \tilde{w} \paren{ P^{\mathrm{s}}_{ \widehat{\Theta}_{\mathrm{ls}}} \mid x  }
        }^2 } } \nonumber \\
    &\leq&   \frac{1}{2}\frac{ L K G^2 }{\mu^3 }
        \EE{ \tilde{\calB} , \calB }{
            \mathbb{E}_{x} \brackets{
                \norm{ \eta^{\mathrm{s}}_{ \widehat{\Theta}_{\mathrm{ls}}} (x) - \eta^{\mathrm{w}}_{\widehat{\Theta}_{\mathrm{w}}} (x) }_2^2
            }
	        } + 16 \sqrt{2} \frac{ K G^2 B \sqrt{\tr{\Sigma_{\rms}}} }{ \mu^2}  \frac{1}{ \sqrt{N} }
	        + \frac{8 C_{\mathrm{unif}}}{\mu} \frac{1}{\sqrt{N}} \nonumber \\
    &&
    + 2  \EE{ \calB, \tilde{\calB} }{
        \mathbb{E}_{x} \brackets{ \normtwo{
        \tilde{w} \paren{ P^{\mathrm{w}}_{\widehat{\Theta}_{\mathrm{w}}} \mid x  }
        - \tilde{w} \paren{ P^{\mathrm{s}}_{ \widehat{\Theta}_{\mathrm{ls}}} \mid x  }
        }^2 }  }  \nonumber  \\
    &\leq&  \paren{ \frac{1}{2}\frac{L K G^2  }{\mu^3}+\frac{1}{2}\frac{K G^2}{\mu^2} }
        \EE{ \tilde{\calB} , \calB }{
            \mathbb{E}_{x} \brackets{
                \norm{ \eta^{\mathrm{s}}_{ \widehat{\Theta}_{\mathrm{ls}}} (x) - \eta^{\mathrm{w}}_{\widehat{\Theta}_{\mathrm{w}}} (x) }_2^2
            }
	        } + 16 \sqrt{2} \frac{ K G^2 B \sqrt{\tr{\Sigma_{\rms}}} }{ \mu^2}  \frac{1}{ \sqrt{N} }
	        + \frac{8 C_{\mathrm{unif}}}{\mu} \frac{1}{\sqrt{N}}  .
	\end{eqnarray}

\medskip

Now, it suffices to upper bound the quantity $\EE{ \tilde{\calB} , \calB }{
            \mathbb{E}_{x} \brackets{
                \norm{ \eta^{\mathrm{s}}_{ \widehat{\Theta}_{\mathrm{ls}}} (x) - \eta^{\mathrm{w}}_{\widehat{\Theta}_{\mathrm{w}}} (x) }_2^2
            }
        }$.

Given $\eta^{\mathrm{w}}_{\widehat{\Theta}_{\mathrm{w}}} (x)$, we can write
\begin{eqnarray}
    && \mathbb{E}_{x} \brackets{ \norm{ \eta^{\mathrm{s}}_{ \widehat{\Theta}_{\mathrm{ls}} } (x)
    - \eta^{\mathrm{w}}_{\widehat{\Theta}_{\mathrm{w}}} (x) }_2^2 } \nonumber \\
    &=& \inf_{ \Theta \in \mathcal{H}_B } \mathbb{E}_{x} \brackets{ \norm{ \Theta^{\top} \phi_{\rms} (x)
    - \eta^{\mathrm{w}}_{\widehat{\Theta}_{\mathrm{w}}} (x) }_2^2 }
    + \mathbb{E}_{x} \brackets{ \norm{ \eta^{\mathrm{s}}_{ \widehat{\Theta}_{\mathrm{ls}}} (x)
    - \eta^{\mathrm{w}}_{\widehat{\Theta}_{\mathrm{w}}} (x) }_2^2 }
    - \inf_{ \Theta \in \mathcal{H}_B } \mathbb{E}_{x} \brackets{ \norm{ \Theta^{\top} \phi_{\rms} (x)
    - \eta^{\mathrm{w}}_{\widehat{\Theta}_{\mathrm{w}}} (x) }_2^2 } .
\end{eqnarray}
To proceed, let us consider the population and the empirical multivariate least-squares objectives:
\begin{equation} \nonumber
    L_{\mathrm{ls}} (\Theta ) \defeq \EE{x}{ \normtwo{ \Theta^{\top} \phi_{\rms} (x) - \eta^{\mathrm{w}}_{\widehat{\Theta}_{\mathrm{w}}} (x) }^2 }
    \quad \text{and} \quad
    \widehat{L}_{\mathrm{ls}} (\Theta ) \defeq \frac{1}{N} \sum_{j=1}^N \normtwo{ \Theta^{\top} \phi_{\rms} (x_j) - \eta^{\mathrm{w}}_{\widehat{\Theta}_{\mathrm{w}}} (x_j) }^2  .
\end{equation}
Recalling $\eta^{\mathrm{s}}_{ \widehat{\Theta}_{\mathrm{ls}}} (x) = ( \widehat{\Theta}_{\mathrm{ls}} )^{\top} \phi_{\mathrm{s}}(x) $, we see that our goal is exactly to upper bound $\EE{ \tilde{\calB} , \calB }{
            \mathbb{E}_{x} \brackets{
                \norm{ \eta^{\mathrm{s}}_{ \widehat{\Theta}_{\mathrm{ls}}} (x) - \eta^{\mathrm{w}}_{\widehat{\Theta}_{\mathrm{w}}} (x) }_2^2
            }
        } = \EE{\calB, \tilde{\calB}}{ L_{\mathrm{ls}} ( \widehat{\Theta}_{\mathrm{ls}} ) }  $.

Let $\Theta_* \in \operatorname{argmin}_{\Theta \in \mathcal{H}_B} L_{\mathrm{ls}}(\Theta)$ be any population minimizer.
Because $\widehat{\Theta}_{\mathrm{ls}}$ minimizes $\widehat{L}_{\mathrm{ls}}$, we have
$
\widehat{L}_{\mathrm{ls}}\left(\widehat{\Theta}_{\mathrm{ls}}\right) \leq \widehat{L}_{\mathrm{ls}}\left(\Theta_*\right) .
$
In addition, since we assume that $\widehat{\Theta}_{\mathrm{ls}} \in \mathcal{H}_B$ almost surely, we can write
\begin{eqnarray*}
    L_{\mathrm{ls}}\left(\widehat{\Theta}_{\mathrm{ls}}\right)-L_{\mathrm{ls}}\left(\Theta_*\right)  &=& \left(L_{\mathrm{ls}}\left(\widehat{\Theta}_{\mathrm{ls}}\right)-\widehat{L}_{\mathrm{ls}}\left(\widehat{\Theta}_{\mathrm{ls}}\right)\right)+\left(\widehat{L}_{\mathrm{ls}}\left(\widehat{\Theta}_{\mathrm{ls}}\right)-\widehat{L}_{\mathrm{ls}}\left(\Theta_*\right)\right)+\left(\widehat{L}_{\mathrm{ls}}\left(\Theta_*\right)-L_{\mathrm{ls}}\left(\Theta_*\right)\right) \\
    &\leq& \sup _{\Theta \in \mathcal{H}_B}\left|L_{\mathrm{ls}}(\Theta)-\widehat{L}_{\mathrm{ls}}(\Theta)\right|+0+\sup _{\Theta \in \mathcal{H}_B}\left|L_{\mathrm{ls}}(\Theta)-\widehat{L}_{\mathrm{ls}}(\Theta)\right| \\
    &=& 2 \sup _{\Theta \in \mathcal{H}_B}\left|L_{\mathrm{ls}}(\Theta)-\widehat{L}_{\mathrm{ls}}(\Theta)\right| .
\end{eqnarray*}
Taking conditional expectations on both sides yields that
\begin{equation} \label{eq:conditional_expectation_L_ls}
\mathbb{E}_{\mathcal{B} \mid \tilde{\mathcal{B}}}\left[L_{\mathrm{ls}}\left(\widehat{\Theta}_{\mathrm{ls}}\right)\right] \leq \inf _{\Theta \in \mathcal{H}_B} L_{\mathrm{ls}}(\Theta)
+ 2 \mathbb{E}_{\mathcal{B} \mid \tilde{\mathcal{B}}}\left[\sup _{\Theta \in \mathcal{H}_B}\left|L_{\mathrm{ls}}(\Theta)-\widehat{L}_{\mathrm{ls}}(\Theta)\right|\right] .
\end{equation}
To proceed, we need the following lemma, which is a generalization bound for multivariate regression with squared loss.
\begin{lemma} \label{lemma:generalization_error_ell}
Conditioned on the labeled dataset \(\tilde{\calB}\), we have
\begin{equation}
    \begin{array}{l}
    \mathbb{E}_{\mathcal{B} \mid \tilde{\mathcal{B}}}\left[
    \sup _{\Theta \in \mathcal{H}_B}\left|L_{\mathrm{ls}}(\Theta)-\widehat{L}_{\mathrm{ls}}(\Theta)\right|
    \right] \\
    \leq
    \dfrac{\paren{B_{\Theta}^{\rmw}}^2}{\sqrt{N}}\left(\mathbb{E}_x\left\|\phi_{\mathrm{w}}(x)\right\|_2^4\right)^{1/2}
    +\dfrac{4 B^2}{\sqrt{N}}\left(\mathbb{E}_x\left\|\phi_{\mathrm{s}}(x)\right\|_2^4\right)^{1 / 2}
    +\dfrac{8 B B_{\Theta}^{\rmw}}{\sqrt{N}}\left(\mathbb{E}_x\left[\left\|\phi_{\mathrm{w}}(x)\right\|_2^2\left\|\phi_{\mathrm{s}}(x)\right\|_2^2\right]\right)^{1 / 2}.
    \end{array}
    \label{eq:generalization_error_ell_feature_moment_bound}
\end{equation}
\end{lemma}
The proof is deferred to Appendix~\ref{appendix:proof_lemma_generalization_error_ell}.

Invoking Lemma~\ref{lemma:generalization_error_ell} and taking expectation over $\tilde{\mathcal{B}}$ on both sides of \eqref{eq:conditional_expectation_L_ls}, we have
\begin{eqnarray}
    && \mathbb{E}_{\tilde{\mathcal{B}}, \mathcal{B}}\left[\mathbb{E}_x\left\|\eta_{\widehat{\Theta}_{\mathrm{ls}}}^{\mathrm{s}}(x)-\eta_{\widehat{\Theta}_{\mathrm{w}}}^{\mathrm{w}}(x)\right\|_2^2\right] \nonumber \\
    &\leq&  \mathbb{E}_{\tilde{\mathcal{B}}}\left[\inf _{\Theta \in \mathcal{H}_B} \mathbb{E}_x\left\|\Theta^{\top} \phi_{\mathrm{s}}(x)-\eta_{\widehat{\Theta}_{\mathrm{w}}}^{\mathrm{w}}(x)\right\|_2^2\right] \nonumber \\
    && + \frac{2\paren{B_{\Theta}^{\rmw}}^2}{\sqrt{N}}\sqrt{\mathbb{E}_{x}\left\|\phi_{\mathrm{w}}(x)\right\|_2^4}
    + \frac{8 B^2}{\sqrt{N}} \sqrt{\mathbb{E}_x\left\|\phi_{\mathrm{s}}(x)\right\|_2^4}
    + \frac{16 B B_{\Theta}^{\rmw}}{\sqrt{N}} \sqrt{\mathbb{E}_{x}\left[\left\|\phi_{\mathrm{s}}(x)\right\|_2^2\left\|\phi_{\mathrm{w}}(x)\right\|_2^2\right]} .
    \label{eq:ls_gap_with_feature_moments}
\end{eqnarray}

Step 1 now is complete.

\medskip

\subsubsection*{Step 2: the second term in \eqref{eq:w_w2s_to_star_decomposition}.  }

Now, let us turn attention to the term
$
    \EE{\tilde{\calB}, \calB}{ \mathbb{E}_{x} \brackets{ \normtwo{
        \tilde{w} \paren{ P^{\mathrm{s}}_{\widehat{\Theta}_{\mathrm{ls}}} \mid x  }
        - \tilde{w} \paren{ P^{*}_{  } \mid x  }
        }^2 } } .
$
We proceed to show that whenever $N \gtrsim d_{\rms}$, we have
\begin{eqnarray}
        && \EE{\tilde{\calB}, \calB}{ \mathbb{E}_{x} \brackets{ \normtwo{
        \tilde{w} \paren{ P^{\mathrm{s}}_{\widehat{\Theta}_{\mathrm{ls}}} \mid x  }
        - \tilde{w} \paren{ P^{*}_{  } \mid x  }
        }^2 } }  \nonumber \\
        &\lesssim&
         \frac{G^2 K }{\mu^2}
        \paren{  \lambda_{\max} \paren{ \Lambda_{\rmw} } \mathbb{E}_{ \tilde{\mathcal{B}}  } \brackets{ \norm{ \widehat{\Theta}_{\mathrm{w}} - \Theta_{\mathrm{w}}^{*}  }_F^2 } \paren{ d_{\rms \wedge \rmw}+ \frac{1}{N} d_{\rms}\left(d_{\rmw}-d_{\rms \wedge \rmw}\right)  }
        + \rho_{\rmw}
        + \rho_{\rms} } \nonumber \\
        &&
        + \frac{1}{\sqrt{N}} \frac{G^2 K }{\mu^2} \paren{  \paren{\E{ \norm{\eta^{*}(x)}_2^4 }}^{\frac{1}{2}} +  B^2  \paren{ \E{ \norm{ \phi_{\rms}(x) }_2^4 } }^{\frac{1}{2}}
    +   B \paren{ \E{ \norm{\eta^{*}(x)}_2^4 } }^{\frac{1}{4}} \paren{ \E{ \norm{\phi_{\rms}(x)}_2^4 }  }^{\frac{1}{4}}  }   \label{eq:main_objective_step_2_upper_bound_w2s}.
\end{eqnarray}

To proceed, invoking Lemma~\ref{lemma:decision_to_logit_continuity} yields that
\begin{equation}
    \mathbb{E}_{x} \brackets{ \normtwo{
        \tilde{w} \paren{ P^{\mathrm{s}}_{\widehat{\Theta}_{\mathrm{ls}}} \mid x  }
        - \tilde{w} \paren{ P^{*}_{  } \mid x  }
        }^2 }
    \leq \frac{G^2 K}{4 \mu^2} \mathbb{E}_{x} \brackets{ \norm{ \eta^{\mathrm{s}}_{ \widehat{\Theta}_{\mathrm{ls}}} (x)
    - \eta^{*} (x) }_2^2 }  .
\end{equation}
Let us consider the population and the empirical least-squares objectives:
\begin{equation}
    L  (\Theta ) \defeq \EE{x}{ \normtwo{ \Theta^{\top} \phi_{\rms} (x) - \eta^{*} (x) }^2 }
    \quad \text{and} \quad
    \widehat{L}  (\Theta ) \defeq \frac{1}{N} \sum_{j=1}^N \normtwo{ \Theta^{\top} \phi_{\rms} (x_j)
    - \eta^{*} (x_j) }^2  .
\end{equation}
Recall $\eta^{\mathrm{s}}_{ \widehat{\Theta}_{\mathrm{ls}}} (x) = \widehat{\Theta}_{\mathrm{ls}}^{\top} \phi_{\rms}(x) \in \mathbb{R}^{K}$.
Hence, we have
\begin{equation}
    \mathbb{E}_{x} \brackets{ \norm{ \eta^{\mathrm{s}}_{ \widehat{\Theta}_{\mathrm{ls}}} (x)
    - \eta^{*} (x) }_2^2 }
    \leq \frac{1}{N} \sum_{j=1}^{N} \norm{ \eta^{\mathrm{s}}_{ \widehat{\Theta}_{\mathrm{ls}}} (x_j) - \eta^{*} (x_j) }_2^2
    + \sup _{\Theta \in \mathcal{H}_B}\left|L_{\mathrm{}}(\Theta)-\widehat{L}_{\mathrm{}}(\Theta)\right| .
    \label{eq:eta_ls_to_eta_star}
\end{equation}
For the latter term, following the same reasoning as in the proof of Lemma~\ref{lemma:generalization_error_ell} with \(g(x)=\eta^*(x)\)  yields that
\begin{equation}
    \begin{array}{l}
    \mathbb{E}_{\tilde{\mathcal{B}}, \mathcal{B}}\left[\sup _{\Theta \in \mathcal{H}_B}\left|L(\Theta)-\widehat{L}(\Theta)\right|\right]
    \\
    \leq \frac{1}{\sqrt{N}} \paren{ \paren{ \E{ \norm{\eta^{*}(x)}_2^4 } }^{\frac{1}{2}} + 4 B^2  \paren{ \E{ \norm{ \phi_{\rms}(x) }_2^4 } }^{\frac{1}{2}}
    + 8 B \paren{ \E{ \norm{\eta^{*}(x)}_2^4 } }^{\frac{1}{4}} \paren{ \E{ \norm{\phi_{\rms}(x)}_2^4 }  }^{\frac{1}{4}}  } .
    \end{array}
    \label{eq:eta_ls_to_eta_star_1}
\end{equation}
Hence, it remains to control the first term in \eqref{eq:eta_ls_to_eta_star}.

To this end, we need to introduce some notation.
For the labeled data set $\tilde{\mathcal{B}} = \braces{ \tilde{x}_i, \tilde{\xi}_i }_{i=1}^{n}$, we define the feature matrices $\tilde{\Phi}_{\rmw},\tilde{\Phi}_{\rms}\in\mathbb{R}^{n\times d_{\phi}}$ with $[\tilde{\Phi}_{\rmw}]_{i,:}=\phi_{\rmw}(\tilde{x}_i)^\top$ and $[\tilde{\Phi}_{\rms}]_{i,:}=\phi_{\rms}(\tilde{x}_i)^\top$.
For the unlabeled data set $\mathcal{B} = \left\{x_j\right\}_{j=1}^N$, we define the feature matrices $\Phi_{\rmw},\Phi_{\rms}\in\mathbb{R}^{N\times d_{\phi}}$ with $[\Phi_{\rmw}]_{j,:}=\phi_{\rmw}(x_j)^\top$ and $[\Phi_{\rms}]_{j,:}=\phi_{\rms}(x_j)^\top$.

We denote $\eta^{*}(X) \in \mathbb{R}^{N \times K}$ as the matrix whose $j$-th row is $\eta^{*}(x_j)^{\top}$.
Denote $P_{\rms} = \Phi_{\rms} \Phi_{\rms}^{\dagger}$, where $\dagger$ represents the Moore-Penrose pseudo-inverse.
We denote $\eta^{\mathrm{s}}_{ \widehat{\Theta}_{\mathrm{ls}}} (X) \in \mathbb{R}^{N \times K}$ as the matrix whose $j$-th row is $\eta^{\mathrm{s}}_{ \widehat{\Theta}_{\mathrm{ls}}} (x_j)$.
Then we can write
$
 \sum_{j=1}^{N} \norm{ \eta^{\mathrm{s}}_{ \widehat{\Theta}_{\mathrm{ls}}} (x_j) - \eta^{*} (x_j) }_2^2
= \norm{ \eta^{\mathrm{s}}_{ \widehat{\Theta}_{\mathrm{ls}}} (X) - \eta^{*}(X) }_F^2
$.
Recalling the definition of $\widehat{\Theta}_{\mathrm{ls}}$ in \eqref{eq:least_squares}, we have $\eta^{\mathrm{s}}_{ \widehat{\Theta}_{\mathrm{ls}}} (X)
= \Phi_{\rms} \widehat{\Theta}_{\mathrm{ls}}
= \Phi_{\rms} \Phi_{\rms}^{\dagger}     \eta^{\mathrm{w}}_{\widehat{\Theta}_{\mathrm{w}}} (X)
= P_{\rms} \Phi_{\mathrm{w}} \widehat{\Theta}_{\mathrm{w}}$.
Denote the residual matrices by $R_{\rmw}(X) = \eta^{*}(X) - \Phi_{\rmw} \Theta_{\rmw}^* \in \mathbb{R}^{N \times K }$ and $R_{\rms}(X) = \eta^{*}(X) - \Phi_{\rms} \Theta_{\rms}^*  \in \mathbb{R}^{N \times K }$,
where we recall that $\Theta_{\rmw}^*$ and $\Theta_{\rms}^*$ are defined in \eqref{eq:approximation_error_s} and \eqref{eq:approximation_error_w}.
Then, clearly we have $\EE{\calB}{ \norm{R_{\rmw}(X)}_F^2 } = N \rho_{\rmw}$. Similarly, we have $\EE{\calB}{ \norm{R_{\rms}(X)}_F^2 } = N \rho_{\rms}$.
By adding and subtracting terms, we can write the decomposition
\begin{eqnarray}
    \eta^{\mathrm{s}}_{ \widehat{\Theta}_{\mathrm{ls}}} (X)  - \eta^{*}(X)
    &=& P_{\rms} \paren{  \Phi_{\mathrm{w}} \widehat{\Theta}_{\mathrm{w}} - \eta^{*}(X)} - (I-P_{\rms}) \eta^{*}(X)   \nonumber \\
    &=& P_{\rms} \paren{  \Phi_{\mathrm{w}} \widehat{\Theta}_{\mathrm{w}}
    - \Phi_{\mathrm{w}} {\Theta}_{\mathrm{w}}^{*}
    + \Phi_{\mathrm{w}} {\Theta}_{\mathrm{w}}^{*}
    - \eta^{*}(X)}
    - (I-P_{\rms}) \paren{ \Phi_{\rms} \Theta_{\rms}^{*} + R_{\rms}(X) }  \nonumber \\
    &=& P_{\rms}   \Phi_{\mathrm{w}} \paren{ \widehat{\Theta}_{\mathrm{w}} - {\Theta}_{\mathrm{w}}^{*} }
    - P_{\rms} R_{\rmw}(X)
    - (I-P_{\rms}) R_{\rms}(X) \label{eq:eta_ls_hat_eta_star_decomposition} .
\end{eqnarray}
Since $P_{\rms}$ is an orthogonal projection, we have
\begin{eqnarray}
    \EE{\calB}{ \norm{ P_{\rms} R_{\rmw}(X)  }_F^2 }
    &\leq& \EE{\calB}{ \norm{  R_{\rmw}(X)  }_F^2 }
    = N \rho_{\rmw}  , \nonumber \\
    \EE{\calB}{ \norm{ (I-P_{\rms}) R_{\rms}(X)  }_F^2 }
    &\leq& \EE{\calB}{ \norm{  R_{\rms}(X)  }_F^2 }
    = N \rho_{\rms}  .  \label{eq:projected_residual}
\end{eqnarray}

Combining \eqref{eq:eta_ls_hat_eta_star_decomposition} and \eqref{eq:projected_residual}, we have
\begin{eqnarray}
    && \EE{\calB, \tilde{\calB}}{ \norm{\eta^{\mathrm{s}}_{ \widehat{\Theta}_{\mathrm{ls}}} (X)  - \eta^{*}(X)}_{F}^{2} } \nonumber \\
    &\leq& 3 \EE{\calB, \tilde{\calB}}{ \norm{ P_{\rms}   \Phi_{\mathrm{w}} \paren{ \widehat{\Theta}_{\mathrm{w}} - {\Theta}_{\mathrm{w}}^{*} } }_{F}^{2} }
    + 3 \EE{\calB}{ \norm{ P_{\rms} R_{\rmw}(X) }_{F}^{2} }
    + 3 \EE{\calB}{ \norm{ (I-P_{\rms}) R_{\rms}(X) }_{F}^{2} } \nonumber \\
    &\leq& 3 \EE{ \calB, \tilde{\calB} }{ \norm{ P_{\rms}   \Phi_{\mathrm{w}} \paren{ \widehat{\Theta}_{\mathrm{w}} - {\Theta}_{\mathrm{w}}^{*} } }_{F}^{2} }
    + 3 N \rho_{\rmw}
    + 3 N \rho_{\rms} .  \label{eq:first_upper_bound_on_eta_ls_eta_star}
\end{eqnarray}

We proceed to upper bound the first term in \eqref{eq:first_upper_bound_on_eta_ls_eta_star}.
Let $\Delta = \widehat{\Theta}_{\mathrm{w}} - \Theta_{\mathrm{w}}^{*} \in \mathbb{R}^{ d_{\phi}  \times K }$.
Recall the spectral decomposition $\Sigma_{\rmw} = V_{\rmw} \Lambda_{\rmw} V_{\rmw}^{\top}$ and $\Sigma_{\rms} = V_{\rms} \Lambda_{\rms} V_{\rms}^{\top}$.
We define $\gamma_{\rms} (x) = \Lambda_{\rms}^{-\frac{1}{2}} V_{\rms}^{\top} \phi_{\rms}(x) \in \mathbb{R}^{d_{\rms}}$ and $\gamma_{\rmw} (x) = \Lambda_{\rmw}^{-\frac{1}{2}} V_{\rmw}^{\top} \phi_{\rmw}(x) \in \mathbb{R}^{d_{\rmw}}$.
One can verify that $\gamma_{\rms} (x)$ and $\gamma_{\rmw} (x)$ are random vectors with zero mean and identity covariance.
Indeed, for instance, $\E{\gamma_{\rms}(x) \gamma_{\rms}(x)^{\top}} = \E{ \Lambda_{\rms}^{-\frac{1}{2}} V_{\rms}^{\top} \phi_{\rms}(x)  \phi_{\rms}(x)^{\top} V_{\rms} \Lambda_{\rms}^{-\frac{1}{2}} } = \Lambda_{\rms}^{-\frac{1}{2}} V_{\rms}^{\top} V_{\rms} \Lambda_{\rms} V_{\rms}^{\top} V_{\rms} \Lambda_{\rms}^{-\frac{1}{2}} = I_{d_s}$.
We define the whitened design matrices to be
$$
\Gamma_s=\Phi_{\rms} V_{\rms} \Lambda_{\rms}^{-\frac{1}{2}} \in \mathbb{R}^{N \times d_s}, \quad \Gamma_{\rmw}=\Phi_{\rmw} V_w \Lambda_{\rmw}^{-\frac{1}{2}} \in \mathbb{R}^{N \times d_w} .
$$
Hence, row $i$ of $\Gamma_s$ is exactly $\gamma_{\rms}\left(x_i\right)^{\top}$, and row $i$ of $\Gamma_{\rmw}$ is $\gamma_{\rmw}\left(x_i\right)^{\top}$.
This way, we have the exact expression $\Phi_{\rmw}=\Gamma_{\rmw} \Lambda_{\rmw}^{\frac{1}{2}} V_w^{\top}$ and $ \Phi_{\rms}=\Gamma_s \Lambda_{\rms}^{\frac{1}{2}} V_{\rms}^{\top} $.
In addition, we denote $\Delta_{\rmw} = \Lambda_{\rmw}^{\frac{1}{2}} V_{\rmw}^{\top} \Delta$.
Therefore, we can write
\begin{equation}
    P_{\mathrm{s}} \Phi_{\mathrm{w}} \paren{ \widehat{\Theta}_{\mathrm{w}} - \Theta_{\mathrm{w}}^{*} }
    = P_{\rms} \Phi_{\rmw}  \Delta
    = P_{\rms} \Gamma_{\rmw} \Lambda_{\rmw}^{\frac{1}{2}} V_w^{\top} \Delta
    = P_{\rms} \Gamma_{\rmw} \Delta_{\rmw} .
\end{equation}
Using the fact $\norm{A B}_F^2 = \tr{B^{\top} A^{\top} A B}$ and $P_{\rms}$ is an orthogonal projection matrix, we further have
\begin{equation}
   \norm{ P_{\mathrm{s}} \Phi_{\mathrm{w}} \paren{ \widehat{\Theta}_{\mathrm{w}} - \Theta_{\mathrm{w}}^{*} } }_F^2
    = \norm{ P_{\rms} \Gamma_{\rmw} \Delta_{\rmw} }_F^2
    = \tr{ \Delta_{\rmw}^{\top} \Gamma_{\rmw}^{\top} P_{\rms}^{\top}  P_{\rms} \Gamma_{\rmw} \Delta_{\rmw} }
    = \tr{ \Delta_{\rmw}^{\top} \Gamma_{\rmw}^{\top} P_{\rms} \Gamma_{\rmw} \Delta_{\rmw} }
    = \tr{  \Gamma_{\rmw}^{\top} P_{\rms} \Gamma_{\rmw} \Delta_{\rmw} \Delta_{\rmw}^{\top} }  \label{eq:Ps_Phi_w_Delta_algebra_calculation}.
\end{equation}

Next, we make the following derivation:
\begin{eqnarray}
    & & \mathbb{E}_{ \tilde{\mathcal{B}}, \calB } \brackets{ \norm{ P_{\mathrm{s}} \Phi_{\mathrm{w}} \Delta  }_F^2  } \nonumber \\
    &\overset{(\text{a})}{=}& \mathbb{E}_{ \tilde{\mathcal{B}}, \calB }
    \brackets{ \tr{  \Gamma_{\rmw}^{\top} P_{\rms} \Gamma_{\rmw} \Delta_{\rmw} \Delta_{\rmw}^{\top} }  } \nonumber \\
    &=& \mathbb{E}_{ \calB } \brackets{
        \mathbb{E}_{ \tilde{\mathcal{B}} \mid \calB }   \brackets{ \tr{  \Gamma_{\rmw}^{\top} P_{\rms} \Gamma_{\rmw} \Delta_{\rmw} \Delta_{\rmw}^{\top} } \mid \calB }
      } \nonumber \\
    &\overset{(\text{b})}{=}& \mathbb{E}_{ \calB } \brackets{
        \tr{  \Gamma_{\rmw}^{\top} P_{\rms} \Gamma_{\rmw}
        \mathbb{E}_{ \tilde{\mathcal{B}} \mid \calB } \brackets{  \Delta_{\rmw} \Delta_{\rmw}^{\top}  \mid \calB }
        } }  \nonumber  \\
    &\overset{(\text{c})}{=}& \mathbb{E}_{ \calB } \brackets{
        \tr{  \Gamma_{\rmw}^{\top} P_{\rms} \Gamma_{\rmw}
        \mathbb{E}_{ \tilde{\mathcal{B}} \mid \calB } \brackets{  \Lambda_{\rmw}^{\frac{1}{2}} V_{\rmw}^{\top} \Delta
        \Delta^{\top} V_{\rmw}^{} \Lambda_{\rmw}^{\frac{1}{2}}       \mid \calB }
        } } \nonumber  \\
    &\overset{(\text{d})}{=}& \mathbb{E}_{ \calB } \brackets{
        \tr{  \Gamma_{\rmw}^{\top} P_{\rms} \Gamma_{\rmw}
        \Lambda_{\rmw}^{\frac{1}{2}} V_{\rmw}^{\top}
        \mathbb{E}_{ \tilde{\mathcal{B}} \mid \calB } \brackets{   \Delta
        \Delta^{\top}        \mid \calB  }
        V_{\rmw}^{} \Lambda_{\rmw}^{\frac{1}{2}}
        } } \nonumber  \\
    &\overset{(\text{e})}{=}& \mathbb{E}_{ \calB } \brackets{
        \tr{  \Gamma_{\rmw}^{\top} P_{\rms} \Gamma_{\rmw}
        \Lambda_{\rmw}^{\frac{1}{2}} V_{\rmw}^{\top}
        \mathbb{E}_{ \tilde{\mathcal{B}}  } \brackets{   \Delta \Delta^{\top} }
        V_{\rmw}^{} \Lambda_{\rmw}^{\frac{1}{2}}
        } } \nonumber  \\
    &\overset{(\text{f})}{=}& \mathbb{E}_{ \calB } \brackets{
        \tr{ V_{\rmw}^{} \Lambda_{\rmw}^{\frac{1}{2}}     \Gamma_{\rmw}^{\top} P_{\rms} \Gamma_{\rmw}
        \Lambda_{\rmw}^{\frac{1}{2}} V_{\rmw}^{\top}
        \mathbb{E}_{ \tilde{\mathcal{B}}  } \brackets{   \Delta \Delta^{\top} }
        } } \nonumber  \\
    &\overset{(\text{g})}{\leq}& \mathbb{E}_{ \calB } \brackets{
        \tr{ V_{\rmw}^{} \Lambda_{\rmw}^{\frac{1}{2}}     \Gamma_{\rmw}^{\top} P_{\rms} \Gamma_{\rmw}
        \Lambda_{\rmw}^{\frac{1}{2}} V_{\rmw}^{\top} }
        \lambda_{\max} \paren{ \mathbb{E}_{ \tilde{\mathcal{B}}  } \brackets{   \Delta \Delta^{\top} }  }
        } \nonumber  \\
    &\overset{(\text{h})}{=}& \mathbb{E}_{ \calB } \brackets{
        \tr{ \Gamma_{\rmw}^{\top} P_{\rms} \Gamma_{\rmw}
        \Lambda_{\rmw}^{}  }
        }
        \lambda_{\max} \paren{ \mathbb{E}_{ \tilde{\mathcal{B}}   } \brackets{   \Delta \Delta^{\top} }  } \nonumber  \\
    &\overset{(\text{i})}{\leq}& \mathbb{E}_{ \calB } \brackets{
        \tr{ \Gamma_{\rmw}^{\top} P_{\rms} \Gamma_{\rmw} }
        \lambda_{\max} \paren{ \Lambda_{\rmw} }
        }
        \lambda_{\max} \paren{ \mathbb{E}_{ \tilde{\mathcal{B}}   } \brackets{   \Delta \Delta^{\top} }  } \nonumber  \\
    &=& \mathbb{E}_{ \calB } \brackets{
        \tr{ \Gamma_{\rmw}^{\top} P_{\rms} \Gamma_{\rmw} }
        }
        \lambda_{\max} \paren{ \Lambda_{\rmw} }
        \lambda_{\max} \paren{ \mathbb{E}_{ \tilde{\mathcal{B}}  } \brackets{   \Delta \Delta^{\top} }  } \label{eq:final_expression_upper_bound_P_s_Phi_w_Delta} .
\end{eqnarray}
Equation~(a) simply follows from \eqref{eq:Ps_Phi_w_Delta_algebra_calculation}.
Equation~(b) holds due to the linearity property of trace.
Equation~(c) is by the definition of $\Delta_{\rmw}$.
Equation~(d) holds since both $\Lambda_{\rmw}^{\frac{1}{2}}$ and $V_{\rmw}$ are deterministic matrices.
Equation~(e) is due to the fact that $\calB$ and $\tilde{\calB}$ are independent.
Equation~(f) follows from the cyclic property of trace.
The inequality (g) and (i) follow from the fact that for psd matrices $A, B$, we have $\operatorname{tr}(A B) \leq \lambda_{\max }(A) \operatorname{tr}(B) $.
Equation~(h) is due to $\tr{ V_{\rmw}^{} \Lambda_{\rmw}^{\frac{1}{2}}     \Gamma_{\rmw}^{\top} P_{\rms} \Gamma_{\rmw}
        \Lambda_{\rmw}^{\frac{1}{2}} V_{\rmw}^{\top} }
        = \tr{ \Gamma_{\rmw}^{\top} P_{\rms} \Gamma_{\rmw}
         \Lambda_{\rmw}^{\frac{1}{2}} V_{\rmw}^{\top} V_{\rmw}^{} \Lambda_{\rmw}^{\frac{1}{2}} }
        = \tr{ \Gamma_{\rmw}^{\top} P_{\rms} \Gamma_{\rmw}
        \Lambda_{\rmw}^{}  }$, which follows from the cyclic property of trace and the fact that $V_{\rmw}$ is an orthogonal matrix.

In what follows, we proceed to upper bound \eqref{eq:final_expression_upper_bound_P_s_Phi_w_Delta}. We proceed in two steps.
\begin{itemize}
    \item First, we deal with the quantity $\mathbb{E}_{ \calB } \brackets{
        \tr{ \Gamma_{\rmw}^{\top} P_{\rms} \Gamma_{\rmw} }
}$.
To this end, we need some more notation.
    We denote
    $M \defeq \E{ \gamma_{\rms}(x) \gamma_{\rmw}(x)^{\top} } = \Sigma_s^{-\frac{1}{2}} \E{ \phi_{\rms}(x) \phi_{\rmw}(x)^{\top} } \Sigma_w^{-\frac{1}{2}} \in \mathbb{R}^{d_s \times d_w}$. We note that $d_{\rms \wedge \rmw } = \norm{M}_{F}^{2}$.
    We define $\varepsilon(x) \defeq \gamma_{\rmw}(x) - M^{\top} \gamma_{\rms}(x) \in \mathbb{R}^{ d_w }$.
    Hence, we can write
    $\Gamma_{\rmw} = \Gamma_s M + E $, where the $i$-th row of the matrix $E$ is $\varepsilon(x_i)^{\top}$.
    We observe the following immediate properties of $\varepsilon(x)$:
    \begin{equation} \label{eq:expected_gamma_s_epsilon_zero}
        \E{ \gamma_{\rms} \varepsilon^{\top} }
        = \E{ \gamma_{\rms} \gamma_{\rmw}^{\top} } - \E{ \gamma_{\rms} \gamma_{\rms}^{\top} } M
        = M  - I_{d_s} M
        = 0 ,
    \end{equation}
    and
    \begin{eqnarray}
        \E{  \varepsilon   \varepsilon^{\top}  }
        &=& \E{ \paren{ \gamma_{\rmw} - M^{\top} \gamma_{\rms} } \paren{ \gamma_{\rmw} - M^{\top} \gamma_{\rms} }^{\top} } \nonumber \\
        &=& \E{ \gamma_{\rmw} \gamma_{\rmw}^{\top}-\gamma_{\rmw}\left(\gamma_{\rms}^{\top} M\right)-\left(M^{\top} \gamma_{\rms}\right) \gamma_{\rmw}^{\top}+\left(M^{\top} \gamma_{\rms}\right)\left(M^{\top} \gamma_{\rms}\right)^{\top} }
        =  I_{d_w} - M^{\top} M .
    \end{eqnarray}

Moreover, we have the following useful bounds on the fourth moments of $\gamma_{\rms}(x)$, $\varepsilon(x)$ and $\phi_{\rms}(x)$.
\begin{lemma} \label{lemma:bound_moments}
    Under Assumption~\ref{assumption:feature_joint_subgaussian}, we have the following facts:
    \begin{itemize}[leftmargin=5em]
        \item $ \E{ \norm{ \gamma_{\rms} (x) }_{2}^{4}  }  \lesssim d_{\rms}^{2} $
        \item $ \E{  \norm{\varepsilon(x)}_{2}^{4} }   \lesssim \paren{ d_{\rmw} - d_{ \rms \wedge \rmw } }^2 $
        \item $ \E{ \norm{ \phi_{\rms} (x) }_{2}^{4}  }  \lesssim \paren{ \tr{ \Sigma_s } }^2  $
    \end{itemize}
\end{lemma}

To handle the quantity $\mathbb{E}_{ \calB } \brackets{
        \tr{ \Gamma_{\rmw}^{\top} P_{\rms} \Gamma_{\rmw} }
}$, we need the following lemma.
\begin{lemma} \label{lemma:upper_bound_expected_tr_w2s_term}
    Under Assumption~\ref{assumption:feature_joint_subgaussian}, there exist constants $c, C > 0$ depending on the sub-gaussian norm of $\phi_{\rms}(x)$ and $\phi_{\rmw}(x)$, such that
    whenever $N \geq C d_{\rms}$, we have
    \begin{equation}
        \mathbb{E}_{ \calB } \brackets{
        \tr{ \Gamma_{\rmw}^{\top} P_{\rms} \Gamma_{\rmw} }
        }
        \leq N d_{\rms \wedge \rmw}
        + C d_{\rms}\left(d_{\rmw}-d_{\rms \wedge \rmw}\right)
        + C d_{\rms}\left(d_{\rmw}-d_{\rms \wedge \rmw}\right)   N \exp \paren{ - c N} .
    \end{equation}
\end{lemma}
The proof is deferred to Appendix~\ref{appendix:proof_lemma_upper_bound_expected_tr_w2s_term}.

\item
To control the quantity $\lambda_{\max} \paren{ \mathbb{E}_{ \tilde{\mathcal{B}}  } \brackets{   \Delta \Delta^{\top} }  }$ in \eqref{eq:final_expression_upper_bound_P_s_Phi_w_Delta}, we simply note that
$$
\lambda_{\max} \paren{ \mathbb{E}_{ \tilde{\mathcal{B}}  } \brackets{   \Delta \Delta^{\top} }  }
\leq  \tr{ \mathbb{E}_{ \tilde{\mathcal{B}}  } \brackets{   \Delta \Delta^{\top} } }
= \mathbb{E}_{ \tilde{\mathcal{B}}  } \brackets{ \tr{  \Delta \Delta^{\top} } }
= \mathbb{E}_{ \tilde{\mathcal{B}}  } \brackets{ \norm{ \Delta }_F^2 }
= \mathbb{E}_{ \tilde{\mathcal{B}}  } \brackets{ \norm{ \widehat{\Theta}_{\mathrm{w}} - \Theta_{\mathrm{w}}^{*}  }_F^2 } ,
$$
where the inequalities follow from the fact that $\mathbb{E}_{ \tilde{\mathcal{B}}  } \brackets{   \Delta \Delta^{\top} } $ is a psd matrix.

\end{itemize}

\medskip

When combining \eqref{eq:final_expression_upper_bound_P_s_Phi_w_Delta} with Lemma~\ref{lemma:upper_bound_expected_tr_w2s_term}, we absorb the exponentially small remainder in Lemma~\ref{lemma:upper_bound_expected_tr_w2s_term} into the term $d_{\rms}\left(d_{\rmw}-d_{\rms \wedge \rmw}\right)$ under $N\gtrsim d_{\rms}$.
The proof of \eqref{eq:main_objective_step_2_upper_bound_w2s} is complete by combining all the pieces together.

\section{Benchmark Bounds and W2S Performance}
\label{sec:comparison_bounds}
The main upper bound becomes most informative when it is compared with a direct strong-only benchmark and paired with a concrete control of the weak teacher. This section develops those auxiliary bounds and then combines them in a case study in Section~\ref{sec:key_discussion} that identifies sufficient conditions under which W2S improves over direct strong-model training.

\subsection{Strong-only lower bound}
\label{sec:lower_bound_strong}

We recall the setup of training the strong model.
We train the strong model by directly minimizing the empirical decision risk on the labeled data:
$
    \widehat{\Theta}_{\mathrm{s}} \in \argmin _{\Theta \in \mathbb{R}^{d_{\phi} \times K}, ~ \Theta \bfone = 0}
        \frac{1}{n} \sum_{i=1}^{n} c \left( \tilde{w}\left(P_{\Theta}^{\mathrm{s}} \mid \tilde{x}_i\right), \tilde{\xi}_i \right) \nonumber .
$
We assume that such a minimizer exists almost surely.

In this subsection, we study how to lower bound
$
\mathbb{E}_{ \tilde{\mathcal{B}} } \brackets{
            R(  \widehat{\pi}_{\mathrm{s}}   ) - R^*
        } .
$
To this end, we first introduce some notation.
Define the single-sample gradient vector
$$
\varphi \defeq \operatorname{vec} \left( \nabla_{\Theta} c\left(\tilde{w}\left(P_{\Theta_s^*}^{\mathrm{s}} (x)\right), \xi\right)  \right) \in \mathbb{R}^{d_{\phi} K} $$
with $(x, \xi) \sim \mathcal{D}$.
We denote $\Sigma_{\varphi} \defeq \operatorname{Var}(\varphi) \in \mathbb{R}^{d_{\phi} K \times d_{\phi} K}$.

\begin{theorem} \label{thm:lower_bound_w2s}
    In addition to Assumption~\ref{assumption:realizability} and \ref{assumption:cost_function},
    suppose the following assumptions hold:
    \begin{enumerate}
    \item
    Assume that there exists $\gamma > 0$ such that for all $x \in \mathcal{X}$, we have almost surely
    \begin{equation} \label{eq:w_tilde_lowerbounded_p}
        \left\|\tilde{w}\left(P_{\widehat{\Theta}_s}^s(x)\right)-\tilde{w}\left(P^*(x)\right)\right\| \geq \gamma\left\|P_{\widehat{\Theta}_s}^s(x)-P^*(x)\right\| .
    \end{equation}
    Moreover, we assume that there exists $B_{\eta} > 0$ such that
    \begin{equation} \label{eq:assumption_eta_two_ends_bounded}
        \norm{ \widehat{\Theta}_{\rms}^{\top} \phi_{\rms}(x)  }_{\infty} \leq B_{\eta}  , \quad  \norm{ \eta^{*}(x) }_{\infty} \leq B_{\eta} .
    \end{equation}

    \item There exists $L_H > 0$ such that, almost surely over $\tilde{\mathcal{B}}$,
    \begin{equation} \label{eq:assumption_hessian_c}
        - L_H I_{d_{\phi}K}  \preceq
        \nabla_{\Theta}^2
        \left[
        \frac{1}{n} \sum_{i=1}^{n} c\left(\tilde{w}\left(P_{\Theta}^{\mathrm{s}} (\tilde{x}_i)\right), \tilde{\xi}_i\right)
        \right]   \preceq L_H I_{d_{\phi}K}
    \end{equation}
    for any $\Theta$ on the line segment connecting $\Theta_{\rms}^{*}$ and $\widehat{\Theta}_{\rms}$, where $\nabla_{\Theta}^2$ is viewed as an operator on $\operatorname{vec}(\Theta)\in\mathbb{R}^{d_{\phi}K}$.

    \end{enumerate}
    Under the above assumptions, for excess decision risk of the strong model, we have
    \begin{equation}
        \mathbb{E}_{ \tilde{\mathcal{B}} } \brackets{
            R(  \widehat{\pi}_{\mathrm{s}}   ) - R^*
        }
    \gtrsim  \mu \gamma^2 \frac{e^{-4 B_{\eta}}}{K^2}  \paren{ \rho_{\rms} + \frac{\lambda_{\min }\left(\Lambda_{\rms}\right) \tr{\Sigma_{\varphi}} }{L_H^2}    \frac{1}{n} }.
    \end{equation}
\end{theorem}

This theorem shows that the excess decision risk of the strong-only benchmark admits two unavoidable sources of error.
The first is the approximation term $\rho_{\rms}$, which reflects the representational mismatch between the ground-truth logit map and the strong feature class.
The second term is the estimation error term, which scales as $\frac{\lambda_{\min }\left(\Lambda_{\rms}\right) \tr{\Sigma_{\varphi}} }{L_H^2}    \frac{1}{n} $.
We stress that this is an algorithm-specific lower bound for our strong-only benchmark training procedure.
The proof of Theorem~\ref{thm:lower_bound_w2s} is relatively standard, and is deferred to Appendix~\ref{appendix:proof_thm_lower_bound_strong}.

The W2S upper bound also depends on the estimation quality of the weak teacher through the term $\mathbb{E}_{ \tilde{\mathcal{B}}  } \brackets{  \norm{ \widehat{\Theta}_{\mathrm{w}} - \Theta_{\mathrm{w}}^{*}  }_F^2 }$. The next subsection gives a concrete control of this term under maximum likelihood training.

\subsection{Weak-model upper bound}
\label{sec:weak_model}
For the sake of completeness and concreteness, this subsection gives a maximum likelihood estimator as an example of a weak teacher training procedure and quantifies its performance. But the general W2S framework is not confined to this specific weak teacher training procedure.

We define $\calC_B = \braces{ \Theta \in \mathbb{R}^{ d_{\phi} \times K} ~\mid~ \Theta \bfone = 0, \norm{ \Theta }_F \leq B_{\Theta} }$ as the constrained parameter space for some $B_{\Theta} > 0$.
We consider the following maximum likelihood estimator based on the weak features:
\begin{equation} \label{eq:weak_mle}
    \widehat{\Theta}_{\rmw} = \argmin_{ \Theta \in  \calC_B }
    ~ - \frac{1}{n} \sum_{i=1}^{n} \ln \paren{ p_{\Theta}^{\rmw} \paren{ \tilde{\xi}_i \mid \tilde{x}_i } } ,
\end{equation}
where we denote $p_{\Theta}^{\rmw} \paren{ \tilde{\xi}_i \mid \tilde{x}_i }$ to be the likelihood of observing $\tilde{\xi}_i$ given input $\tilde{x}_i$ under the weak feature model parameterized by $\Theta$.
 Let $\mu_{\rmw} \defeq \frac{e^{-2B_{\Theta}B_{\phi}}}{K}\lambda_{\min}\paren{\Sigma_{\rmw}}$.

\begin{assumption}[identifiability of $\Theta_{\rmw}^{*}$] \label{assumption:identifiability_weak}
    The ground-truth weak model parameter $\Theta_{\rmw}^{*}$ is identifiable in the sense that $\rho_{\rmw} = 0$ and the minimizer $\Theta_{\rmw}^{*}$ is unique.
    Moreover, we assume that $\Theta_{\rmw}^{*} $ is attained in the interior of $\calC_B$, i.e., $\norm{ \Theta_{\rmw}^{*} }_F < B_{\Theta}$.
\end{assumption}

\begin{assumption} [bounded weak features] \label{assumption:bounded_weak_features}
    There exists a constant $B_{\phi} > 0$ such that for all $x \in \mathcal{X}$, we have
    $
        \norm{ \phi_{\rmw} (x) }_2 \leq B_{\phi} .
    $
\end{assumption}

\begin{theorem} \label{thm:weak_model_upper_bound}
    Under Assumption~\ref{assumption:realizability}, Assumption~\ref{assumption:identifiability_weak} and Assumption~\ref{assumption:bounded_weak_features},
    the MLE estimator $\widehat{\Theta}_{\rmw}$ defined in \eqref{eq:weak_mle} enjoys that
    \begin{equation}
        \EE{\tilde{\calB}}{ \norm{ \widehat{\Theta}_{\rmw} - \Theta_{\rmw}^{*} }_{F}^{2} }
        \leq   \frac{4}{\mu_{\rmw}^{2}} \frac{\tr{\Sigma_{\rmw}}}{n}
        +  4 B_{\Theta}^2   d_{\rmw} \exp \paren{ - \frac{ n \lambda_{\min} \paren{ \Sigma_{\rmw} } }{ 8 B_{\phi}^2 } }  ,
    \end{equation}
\end{theorem}

The proof of this theorem is deferred to Appendix~\ref{appendix:proof_thm_weak_model}.

\subsection{When does W2S outperform? }
\label{sec:key_discussion}

We now turn the earlier bounds into a sufficient condition under which W2S improves over direct strong-model training.
Combining the W2S upper bound in Theorem~\ref{thm:upper_bound_w2s} with the strong-only lower bound in Theorem~\ref{thm:lower_bound_w2s}, it suffices, up to universal constants, that
\begin{equation}
    \mathcal{E}_{\mathrm{imit}}
    +\mathcal{E}_{N}
    +\mathcal{E}_{\mathrm{teacher}}
    +\mathcal{E}_{\mathrm{approx}}
    \lesssim
    \mu\gamma^2\frac{e^{-4B_{\eta}}}{K^2}
    \paren{
        \rho_{\rms}
        +
        \frac{\lambda_{\min}\paren{\Lambda_{\rms}}\tr{\Sigma_{\varphi}}}{L_H^2 n}
    } .
    \label{eq:w2s_better_condition_refined}
\end{equation}
The left-hand side collects the costs paid by W2S: imitation, unlabeled-sample fluctuation, teacher estimation, and model misspecification.
The right-hand side is the strong model's approximation and labeled-sample statistical error.
Thus, the comparison asks whether the weak teacher and the unlabeled sample can offset the label scarcity faced by direct strong-model training.

We next plug the weak-model estimate from Theorem~\ref{thm:weak_model_upper_bound} into the weak-to-strong term $\mathcal{E}_{\mathrm{teacher}}$ defined after Theorem~\ref{thm:upper_bound_w2s}.
To obtain a transparent sufficient condition, we work in the large-\(n\) regime made explicit in Corollary~\ref{cor:w2s_outperformance_certificate}.
Also define
$
    C_{\mathrm{t}}
    \defeq
    \frac{4LKG^2}{\mu^2\mu_{\rmw}^2}
    \lambda_{\max}\paren{\Lambda_{\rmw}}
    \tr{\Sigma_{\rmw}} .
$
Then $\mathcal{E}_{\mathrm{teacher}}$ is bounded by
\begin{equation}
    \mathcal{E}_{\mathrm{teacher}}
    \lesssim
    \frac{C_{\mathrm{t}}}{n}
    \paren{
        d_{\rms\wedge\rmw}
        +
        \frac{d_{\rms}\paren{d_{\rmw}-d_{\rms\wedge\rmw}}}{N}
    } .
\end{equation}
In the decomposition following Theorem~\ref{thm:upper_bound_w2s}, all terms in the unlabeled-sample statistical error $\mathcal{E}_N$ scale as $N^{-1/2}$.
The imitation term $\mathcal{E}_{\mathrm{imit}}$, which contains $\EE{\tilde{\calB}}{\rho_{\rms \rightarrow \rmw}\paren{\widehat{\Theta}_{\mathrm{w}}}}$, is also independent of $N$.

Define the residual margin as
\begin{equation*}
    \mathcal{M}_n
    \defeq
    \mu\gamma^2\frac{e^{-4B_{\eta}}}{K^2}
    \paren{
        \rho_{\rms}
        +
        \frac{\lambda_{\min}\paren{\Lambda_{\rms}}\tr{\Sigma_{\varphi}}}{L_H^2 n}
    }
    -
    \mathcal{E}_{\mathrm{imit}}
    -
    \mathcal{E}_{\mathrm{approx}}
    -
    \frac{C_{\mathrm{t}}d_{\rms\wedge\rmw}}{n}.
\end{equation*}
With this definition, the condition \eqref{eq:w2s_better_condition_refined} can be written as
\begin{equation}
    \frac{C_N}{\sqrt{N}}
    +
    \frac{C_{\mathrm{t}}d_{\rms}\paren{d_{\rmw}-d_{\rms\wedge\rmw}}}{nN}
    \lesssim
    \mathcal{M}_n ,
    \label{eq:w2s_case_study_overlap_condition}
\end{equation}
which leads to the following observation.

\begin{corollary}[A sufficient condition for W2S outperformance] \label{cor:w2s_outperformance_certificate}
    Suppose the assumptions of Theorems~\ref{thm:upper_bound_w2s}, \ref{thm:lower_bound_w2s}, and \ref{thm:weak_model_upper_bound} hold.
    Assume the following three conditions hold.
    \begin{enumerate}[label=(\roman*)]
        \item The labeled sample size \(n\) is large enough that the exponentially small term in Theorem~\ref{thm:weak_model_upper_bound} is dominated by its \(1/n\) term:
        \(
            B_{\Theta}^2 d_{\rmw}
            \exp \paren{ - \frac{ n \lambda_{\min} \paren{ \Sigma_{\rmw} } }{ 8 B_{\phi}^2 } }
            \lesssim
            \frac{\tr{\Sigma_{\rmw}}}{\mu_{\rmw}^2 n}.
        \)
        \item The imitation error $\mathcal{E}_{\mathrm{imit}}$, the model-approximation error $\mathcal{E}_{\mathrm{approx}}$, and the overlap cost \(C_{\mathrm{t}}d_{\rms\wedge\rmw}/n\) are small enough relative to the strong-only benchmark term to leave a positive margin:
        \begin{equation}
        \begin{aligned}
            \mathcal{E}_{\mathrm{imit}}
            + \mathcal{E}_{\mathrm{approx}}
            + \frac{C_{\mathrm{t}}d_{\rms\wedge\rmw}}{n}
            &<
            \mu\gamma^2\frac{e^{-4B_{\eta}}}{K^2}
            \paren{
                \rho_{\rms}
                +
                \frac{\lambda_{\min}\paren{\Lambda_{\rms}}\tr{\Sigma_{\varphi}}}{L_H^2 n}
            } .
        \end{aligned}
            \label{eq:w2s_outperformance_positive_margin}
        \end{equation}
        Equivalently, \(\mathcal{M}_n>0\).
        \item The unlabeled sample size \(N\) is large enough:
        \begin{equation}
            N
            \gtrsim
            \max\braces{
            d_{\rms},
            \paren{
                \frac{
                    C_N+
                    \sqrt{
                        C_N^2+
                        4\mathcal{M}_n C_{\mathrm{t}}d_{\rms}\paren{d_{\rmw}-d_{\rms\wedge\rmw}}/n
                    }
                }{
                    2\mathcal{M}_n
                }
            }^2
            }.
            \label{eq:w2s_outperformance_sample_size}
        \end{equation}
    \end{enumerate}
    Under these conditions, the previous bounds certify, up to universal constants, that
    \begin{equation}
        \EE{\tilde{\calB},\calB}{R(\widehat{\pi}_{\mathrm{w2s}})-R^*}
        \leq
        \EE{\tilde{\calB}}{R(\widehat{\pi}_{\mathrm{s}})-R^*}.
    \end{equation}
\end{corollary}

Therefore, W2S can outperform  if the \(N\)-independent costs $\mathcal{E}_{\mathrm{imit}}$, $\mathcal{E}_{\mathrm{approx}}$, and \(C_{\mathrm{t}}d_{\rms\wedge\rmw}/n\) are small enough to leave a positive residual margin \(\mathcal{M}_n>0\).
This condition combines three requirements: the strong student must be able to imitate the weak teacher accurately, the two feature classes must have small approximation error, and the overlap dimension \(d_{\rms\wedge\rmw}\) must be sufficiently small relative to the labeled sample size \(n\).
In practice, this is a mild condition that is commonly satisfied.
Intuitively, a smaller overlap dimension means that the ways in which the teacher and student think are different enough, so the teacher's mistakes are more likely to be averaged out rather than directly inherited.
In \eqref{eq:w2s_case_study_overlap_condition}, once \(\mathcal{M}_n>0\), the finite-\(N\) terms \(C_N/\sqrt{N}\) and \(C_{\mathrm{t}}d_{\rms}\paren{d_{\rmw}-d_{\rms\wedge\rmw}}/(nN)\) decrease as \(N\) grows; with fixed problem-dependent constants, \eqref{eq:w2s_outperformance_sample_size} scales qualitatively as \(N\gtrsim \max\{d_{\rms},n^2\}\).

\section{Numerical Experiments}
\label{sec:numerical_experiments}

In this section, we present numerical experiments to validate our theoretical findings.
In Section~\ref{subsec:synthetic_experiments}, we evaluate the empirical performance of W2S in a contextual newsvendor problem, where the data-generating process can be precisely controlled to isolate and validate the theoretical mechanisms identified in the previous sections.
Furthermore, Section~\ref{subsec:comment_moderation_experiments} evaluates W2S on a real-world comment moderation dataset to test whether the predicted gains persist in a more realistic operational setting.

\subsection{Synthetic Experiments}
\label{subsec:synthetic_experiments}

We consider a synthetic contextual single-item newsvendor problem.

\paragraph{Setup.}

We take a finite discrete support $\Xi = \braces{z_1, \cdots, z_K} \subseteq \mathbb{R}$ where $z_1 < z_2 < \cdots < z_K$, and interpret $\xi \in \Xi$ as the random demand for a product.
The decision is an order quantity $w \in \mathbb{R}$ before the demand $\xi$ is realized.
We use the standard newsvendor cost function $c(w,\xi) = c_u (\xi - w)_+ + c_o (w - \xi)_+$, where $c_u > 0$ is the underage cost, $c_o > 0$ is the overage cost, and $(a)_+ = \max\{a,0\}$.
We note that such a cost function is neither strongly convex nor smooth in $w$, and hence does not satisfy Assumption~\ref{assumption:cost_function}.
It is well-known that the optimal order quantity given the context vector $x$ and the demand distribution $P(\cdot \mid x)$ is given by the quantile function:
\begin{equation} \label{eq:optimal_action_synthetic}
    \tilde{w}\paren{P \mid x } = \inf \braces{z \in \mathbb{R}: \prr{\xi \sim P(\cdot \mid x)}{ \xi \leq z} \geq \frac{c_u}{c_u+c_o} } .
\end{equation}
One can verify that $\tilde{w}\paren{P \mid x } \in \Xi$.
We work with a context vector $x$ sampled from $ \mathcal{N}(0, I_{d_x})$.
We consider the strong and weak feature models $\phi_{\rms}(x) = V_{\rms}^{\top} x$ and $\phi_{\rmw}(x) = V_{\rmw}^{\top} x$, where $V_{\rms} \in \mathbb{R}^{d_x \times d_{\phi}}$ and $V_{\rmw} \in \mathbb{R}^{d_x \times d_{\phi}}$ are fixed matrices with orthonormal columns.
We explicitly construct $V_{\rms}$ and $V_{\rmw}$ such that $\E{ \phi_{\rms}(x) \phi_{\rmw}(x)^\top } = \E{ V_{\rms}^{\top} x x^\top V_{\rmw} } = V_{\rms}^{\top} V_{\rmw}  = \left[\begin{array}{cc}
I_r & 0 \\
0 & 0
\end{array}\right]$.
First, we need $0 \leq r \leq \min\{d_{\rms}, d_{\rmw}\}$, $d_{\phi} \geq \max\{d_{\rms}, d_{\rmw}\}$ and $d_x \geq d_{\rms} + d_{\rmw} - r$.
We choose an orthonormal basis $Q = \left[q_1, \ldots, q_{d_x}\right] \in \mathbb{R}^{d_x \times d_x}$.
Then we set $U_{\rms} = \left[q_1, \ldots, q_r, q_{r+1}, \ldots, q_{d_s}\right] \in \mathbb{R}^{d_x \times d_s}$ and $U_{\rmw} = \left[q_1, \ldots, q_r, q_{d_s+1}, \ldots, q_{d_s+\left(d_w-r\right)}\right] \in \mathbb{R}^{d_x \times d_w}$.
This way, we have $U_s^{\top} U_w=\left[\begin{array}{cc}
I_r & 0 \\
0 & 0
\end{array}\right]$.
To construct $V_{\rms}$ and $V_{\rmw}$, it suffices to pad zeroes: $V_{\rms}=\left[U_s, 0_{d_x \times\left(d_\phi-d_s\right)}\right],  V_w=\left[U_w, 0_{d_x \times\left(d_\phi-d_w\right)}\right]$.
Hence, we have $\E{ \phi_{\rms}(x) \phi_{\rms}(x)^\top } = V_{\rms}^{\top} V_{\rms}  = \left[\begin{array}{cc}
I_{d_{\rms}} & 0 \\
0 & 0
\end{array}\right]$, and similarly for $\Sigma_{\rmw}$.
One can easily verify that $d_{s\wedge w} = \norm{ \Sigma_s^{-\frac{1}{2}} \E{ \phi_{\rms}(x) \phi_{\rmw}(x)^\top } \Sigma_w^{-\frac{1}{2}} }_F^2 =\|V_{\rms}^\top V_w\|_F^2=r$.

We make the ground-truth environment realizable by the strong model (hence there is no approximation error for the strong model, i.e., $\rho_{\rms} = 0$), by defining the ground-truth logits to be linear in the strong features: $\eta^{*}(x)=\Theta_*^{\top} \phi_{\rms}(x) \in \mathbb{R}^K$ and $P^*\left(\xi=z_k \mid x\right)=\operatorname{softmax}_k\left(\eta^{*}(x)\right)$.

As noted in \cite{qi2025integrated}, the W2S training objective
\[
    \widehat{\Theta}_{\mathrm{w2s}} \in \argmin _{ \Theta \in \mathbb{R}^{d_{\phi} \times K}, ~ \norm{\Theta}_F \leq B } \frac{1}{N}
    \sum_{j=1}^{N} \mathbb{E}_{\xi \sim P_{ \widehat{\Theta}_{\mathrm{w}}}^{\mathrm{w}}(\cdot \mid x_j ) }
    \left[ c\left(\tilde{w}\left(P^{\mathrm{s}}_\Theta \mid x_j \right), \xi_j\right) \right]
\]
is in general neither convex nor differentiable, which is also the case even in this synthetic newsvendor problem.
Indeed, the objective function is piecewise constant with respect to $\Theta$ because the optimal order quantity $\tilde{w}\left(P^{\mathrm{s}}_\Theta \mid x_j \right)$ only changes when the quantile of the distribution $P^{\mathrm{s}}_\Theta(\cdot \mid x_j )$ crosses one of the demand values in $\Xi$.
To this end, we work with a smoothed surrogate optimal action function instead of the exact optimal action function in \eqref{eq:optimal_action_synthetic}.
To be specific, the induced conditional pmf on $\Xi$ by the strong model is given by $P^{\mathrm{s}}_\Theta(\xi = z_k \mid x) = \frac{\exp \left(\Theta_k^{\top} V_{\rms}^{\top} x\right)}{\sum_{j=1}^K \exp \left(\Theta_j^{\top} V_{\rms}^{\top} x\right)}$.
For $\beta>0$, define $a_{\Theta, k}^{\beta} (x) = - \beta \paren{ \sum_{\ell=1}^{k} P^{\mathrm{s}}_\Theta(\xi = z_{\ell} \mid x) - \frac{c_u}{c_u+c_o} }^2$.
We then use
\begin{equation}  \label{eq:surrogate_oracle_synthetic}
    \widehat{w}_{\beta} \paren{ P_{\Theta}^{\rms} \mid x } = \sum_{k=1}^{K} \operatorname{softmax}_k\paren{a_{\Theta}^{\beta}(x)} z_k
\end{equation}
as the surrogate of $\tilde{w} \paren{ P_{\Theta}^{\rms} \mid x }$, where $\beta$ controls the sharpness of the approximation.
Moreover, we use $\tau \log \left(1+e^{t / \tau}\right)$ to approximate $(t)_+$ for some small smoothing parameter $\tau > 0$.
With all these in hand, we use a first-order method to approximately solve for $\widehat{\Theta}_{\mathrm{w2s}}$.
Again following the spirit of \cite{qi2025integrated}, in the numerical implementation reported below, we use the following regularized version of the surrogate objective:
\begin{equation}
\label{eq:w2s_surrogate_objective}
    \widehat{\Theta}_{\mathrm{w2s}} \in \argmin_{\Theta \in \mathbb{R}^{d_{\phi} \times K}}
    \frac{1}{N}\sum_{j=1}^{N}\mathbb{E}_{\xi \sim P_{\widehat{\Theta}_{\mathrm{w}}}^{\mathrm{w}}(\cdot \mid x_j)}
    \left[
    c\left(\widehat{w}_{\beta}\left(P_{\Theta}^{\mathrm{s}}\mid x_j\right),\xi\right)
    \right]
    +\lambda_{\mathrm{w2s}}\left\|\Theta-\Theta_{\mathrm{s}}^{0}\right\|_F^2,
\end{equation}
where $\Theta_{\mathrm{s}}^{0}$ denotes the initial strong-model parameter used to initialize W2S and $\lambda_{\mathrm{w2s}}=5\times 10^{-4}$.
We note that we only use the smoothed surrogate objective for numerical optimization; the evaluation of decision risk is still based on the original newsvendor cost function without smoothing.

\begin{figure}[htbp]
    \centering

    \begin{subfigure}[t]{0.49\linewidth}
        \centering
        \includegraphics[width=\linewidth]{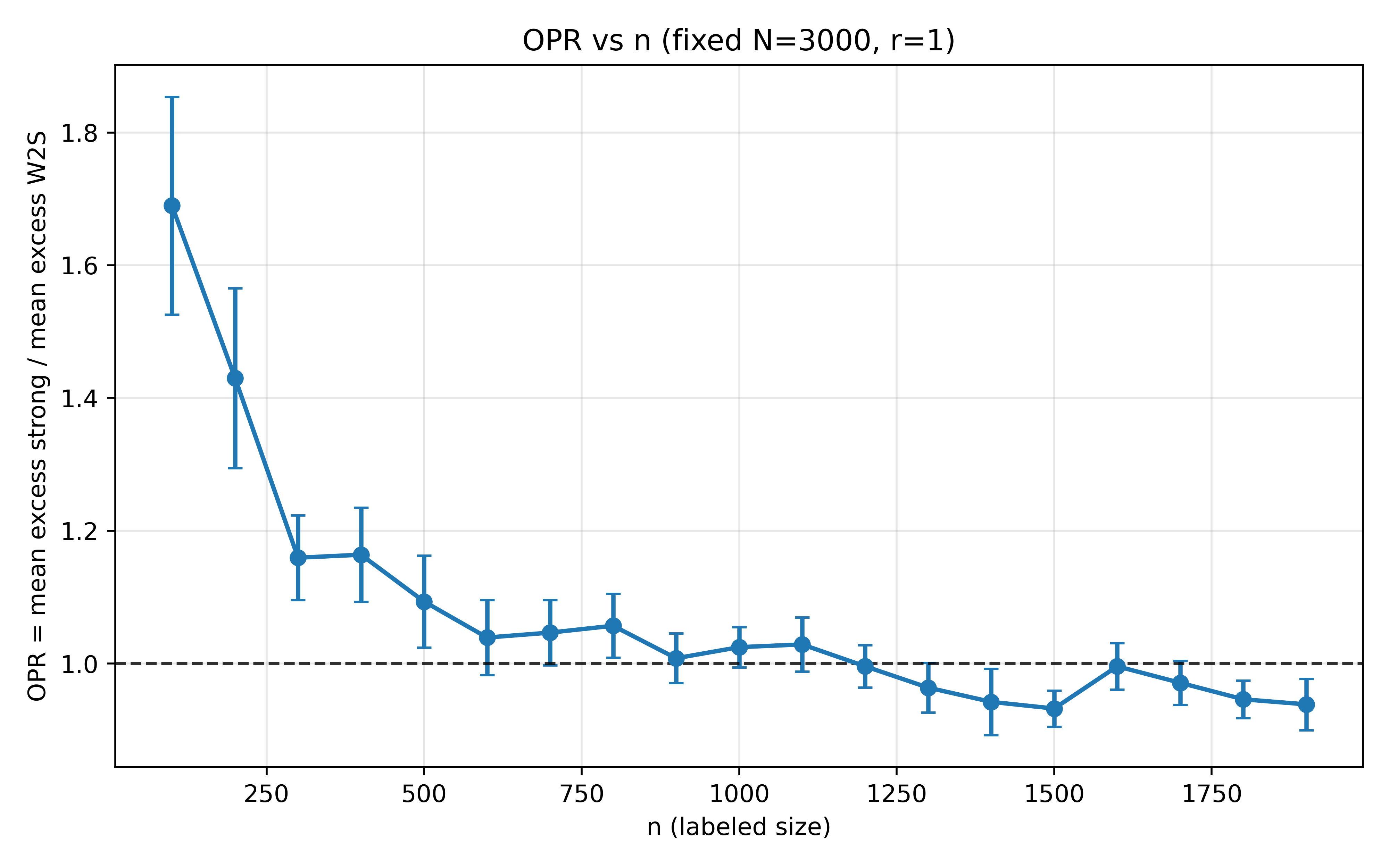}
        \caption{\textbf{OPR vs.\ labeled size $n$} with fixed unlabeled size $N=3000$ and overlap $d_{\rms \wedge \rmw}=1$.}
        \label{fig:opr_vs_n_r1}
    \end{subfigure}
    \hfill
    \begin{subfigure}[t]{0.49\linewidth}
        \centering
        \includegraphics[width=\linewidth]{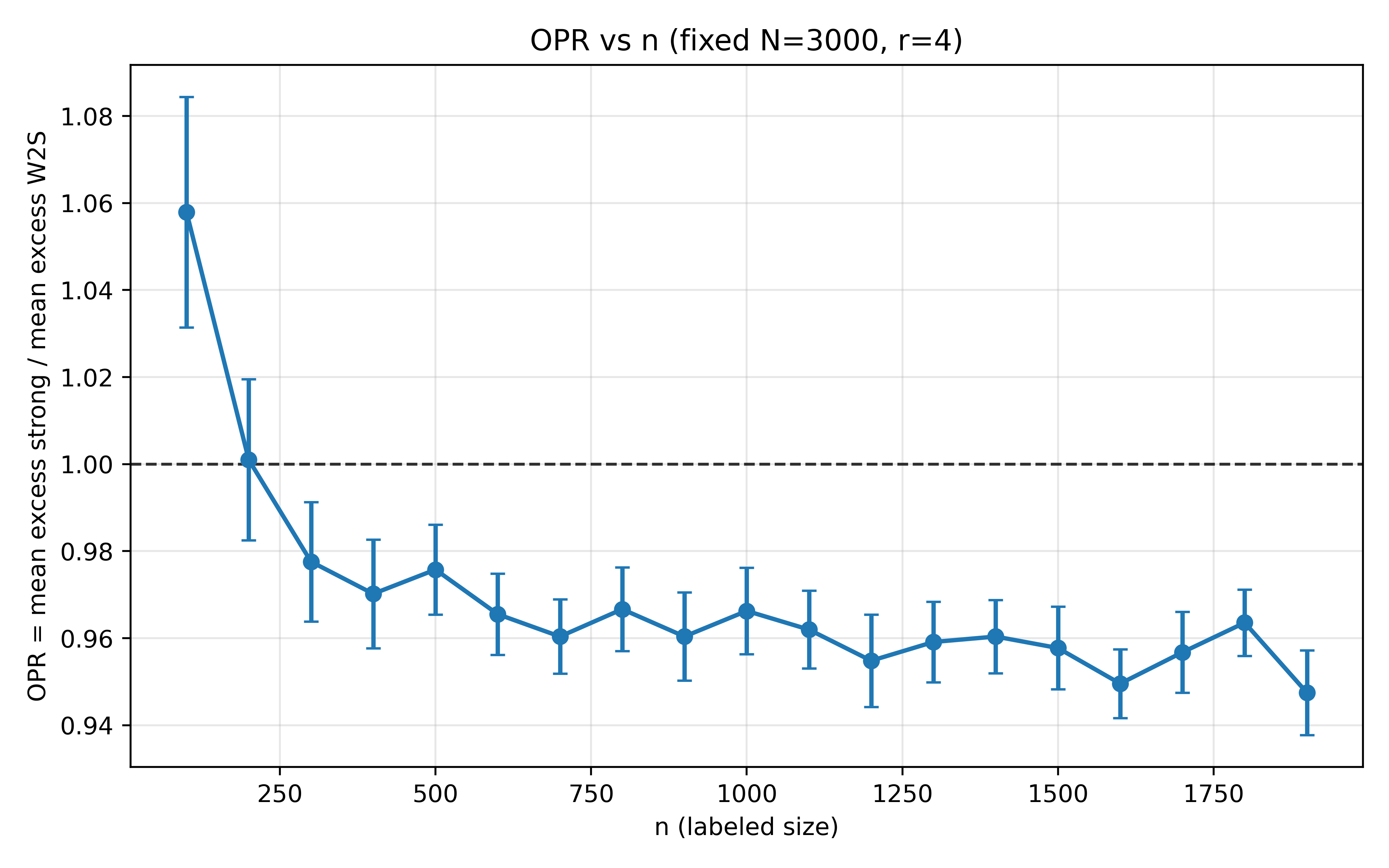}
        \caption{\textbf{OPR vs.\ labeled size $n$} with fixed unlabeled size $N=3000$ and overlap $d_{\rms \wedge \rmw}=4$.}
        \label{fig:opr_vs_n_r4}
    \end{subfigure}

    \vspace{0.5em}

    \begin{subfigure}[t]{0.49\linewidth}
        \centering
        \includegraphics[width=\linewidth]{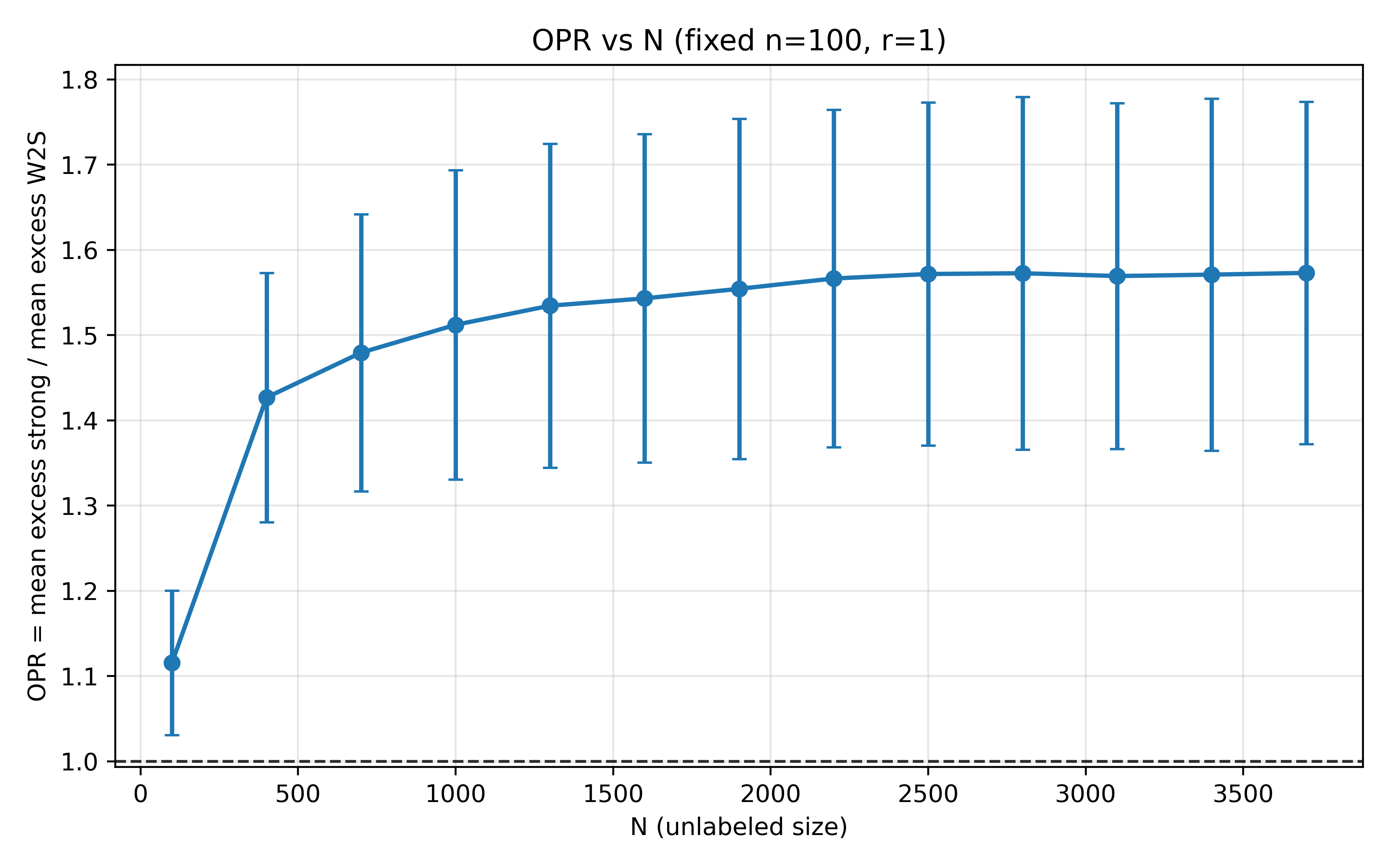}
        \caption{\textbf{OPR vs.\ unlabeled size $N$} with fixed labeled size $n=100$ and overlap $d_{\rms \wedge \rmw}=1$.}
        \label{fig:opr_vs_N_r1}
    \end{subfigure}
    \hfill
    \begin{subfigure}[t]{0.49\linewidth}
        \centering
        \includegraphics[width=\linewidth]{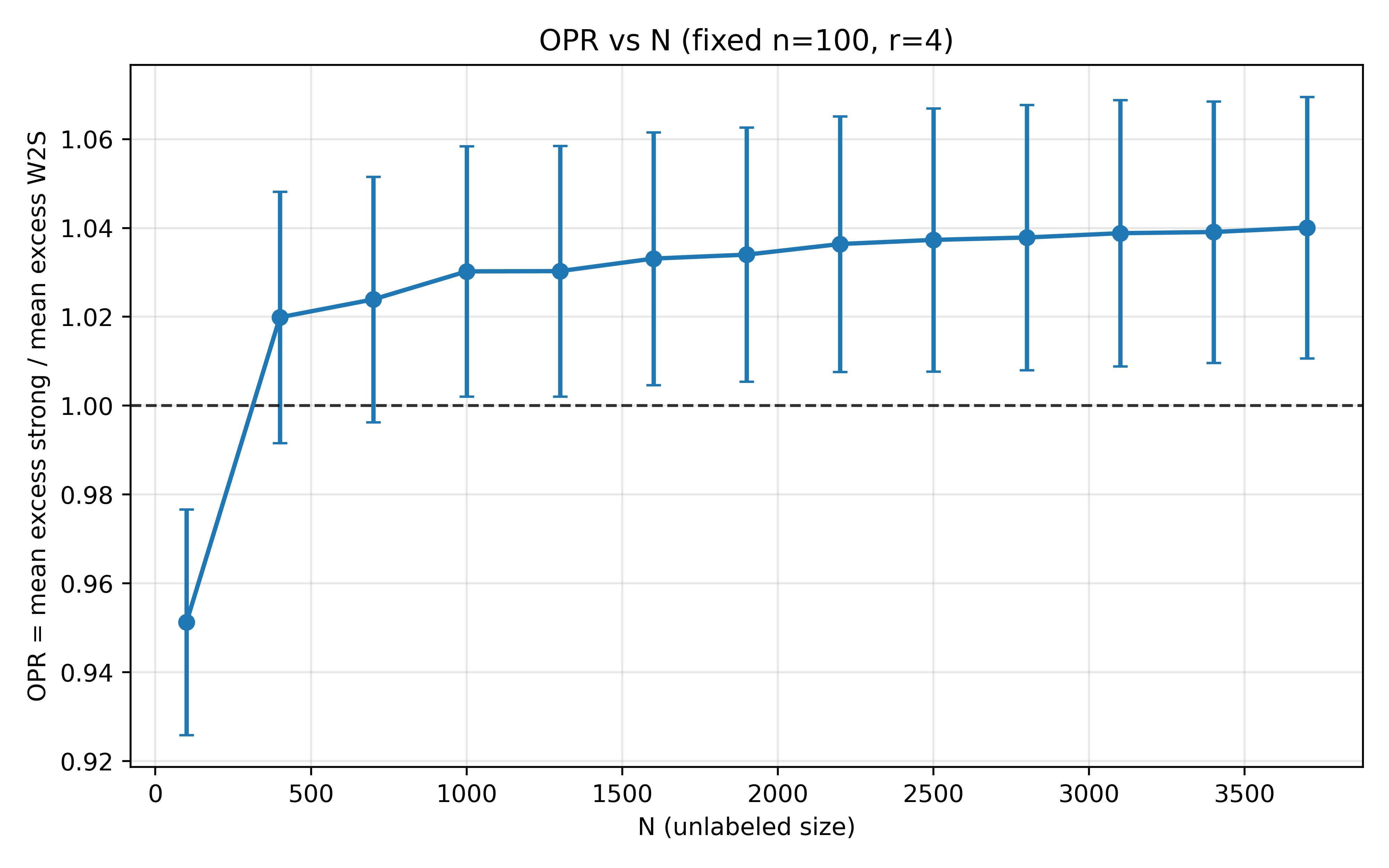}
        \caption{\textbf{OPR vs.\ unlabeled size $N$} with fixed labeled size $n=100$ and overlap $d_{\rms \wedge \rmw}=4$.}
        \label{fig:opr_vs_N_r4}
    \end{subfigure}

    \caption[Sensitivity of the optimization performance ratio to sample sizes and overlap dimension]{
    Sensitivity of the \emph{optimization performance ratio} (OPR) to labeled and unlabeled sample sizes and to the overlap dimension.
    OPR is defined as $\mathrm{OPR} := \frac{\text{ExcessRisk(Strong Benchmark)}}{\text{ExcessRisk(W2S)}} = \frac{ \EE{ \tilde{\calB} }{ R( \widehat{ \pi }_{\mathrm{s}} ) - R^* }   }{ \EE{ \tilde{\calB}, \calB }{ R( \widehat{ \pi }_{\mathrm{w2s}} ) - R^* } }$, where excess risk is measured relative to the oracle using the ground-truth conditional distribution.
    Each plotted value is the ratio of mean excess risks across 60 independent trials; error bars show approximate 95\% normal confidence bands computed as $1.96$ times a bootstrap standard error for this ratio-of-means estimator.
    Parameters used for the experiment: $d_x=80, \Xi=\{0,0.5,1,\cdots,10\}, d_{\rms}=10, d_{\rmw}=48, c_o=1, c_u=3, \beta=35,\tau=0.5$.
    }
    \label{fig:opr_labeled_unlabeled_overlap}
\end{figure}

\medskip

\paragraph{Results.}

We can make a few observations on Figure~\ref{fig:opr_labeled_unlabeled_overlap}.

\begin{enumerate}
    \item W2S provides the largest gains when labeled data are scarce, especially under low overlap. In the top-left panel ($d_{\rms \wedge \rmw}=1, \text{fixed } N=3000$), the OPR starts well above 1 and then decreases toward 1 as $n$ grows.
    This suggests that when the strong model is data-limited, leveraging abundant unlabeled data through the weak teacher can materially reduce excess decision risk compared to training the strong model on labeled data alone.

    \item The benefit of W2S diminishes as overlap increases. Comparing $d_{\rms \wedge \rmw}=1$ vs $d_{\rms \wedge \rmw}=4$, the curves are systematically closer to 1 when $d_{\rms \wedge \rmw}=4$.
    In the top-right panel, OPR is only slightly above (or near) 1 for small $n$ and drifts below (or around) 1 as $n$ increases.
    Namely, when the weak and strong representations share more directions, the extra information conveyed by the weak teacher is less distinctive, so W2S yields only modest improvement.

    \item More unlabeled data improves W2S up to a plateau, with stronger effects at low overlap. For example, in the bottom-left panel ($d_{\rms \wedge \rmw}=1, \text{fixed } n=100$), OPR increases as $N$ grows and then levels off, indicating that additional unlabeled samples help W2S learn a better policy until the benefit saturates.
\end{enumerate}

\begin{figure}[htbp]
    \centering
    \includegraphics[width=0.7\linewidth]{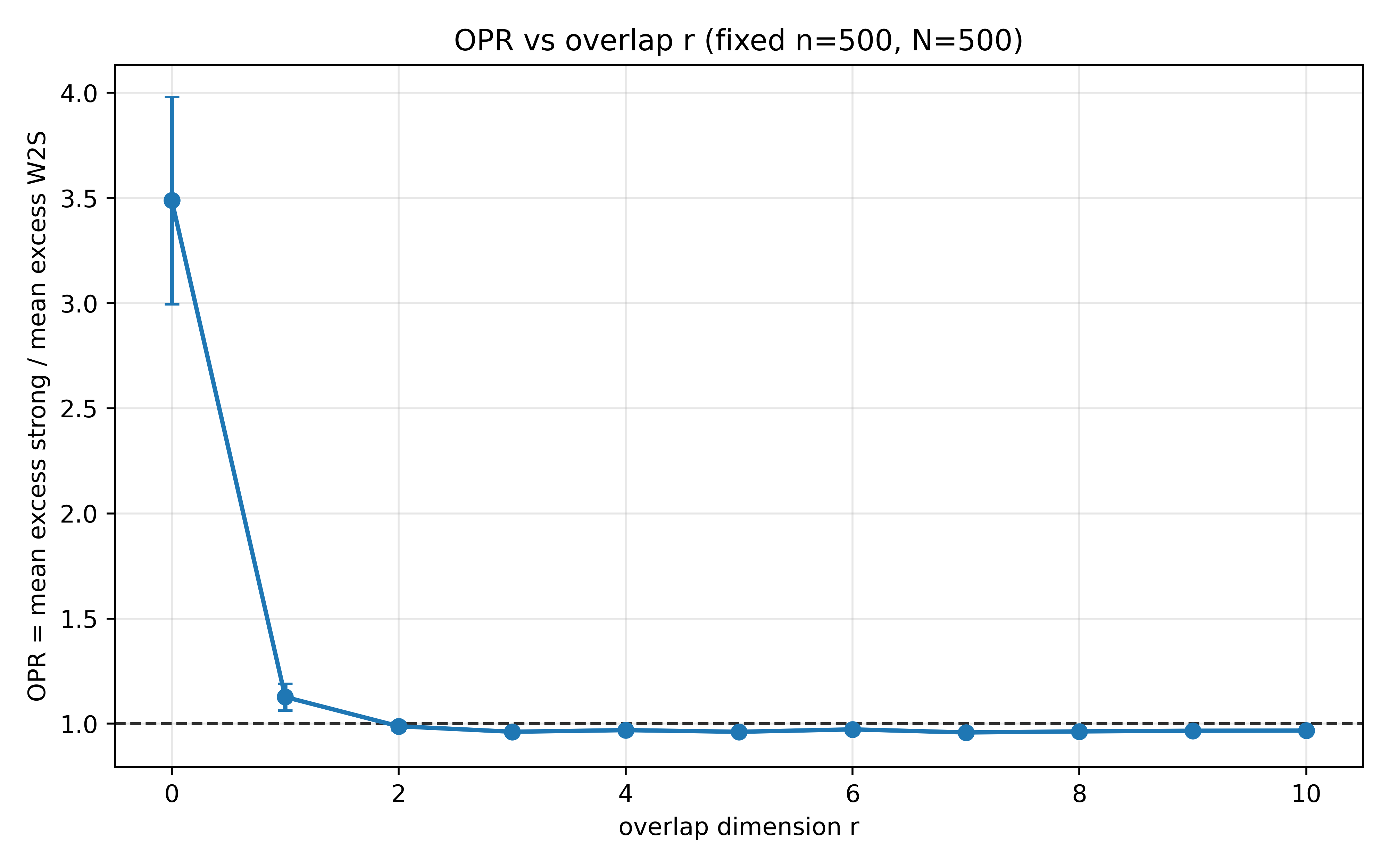}
    \caption{
    Effect of the strong--weak feature overlap on the OPR.
    We fix the labeled and unlabeled sample sizes to $n=500$ and $N=500$, and vary the overlap dimension $d_{\rms \wedge \rmw}$.
    The large OPR at small $d_{\rms \wedge \rmw}$ indicates that W2S substantially reduces excess decision risk when the weak and strong representations share little (or no) subspace, whereas the curve approaching $\mathrm{OPR}\approx 1$ for larger $d_{\rms \wedge \rmw}$ indicates that the benefit of W2S diminishes as the two feature spaces become more aligned.
    }
    \label{fig:opr_vs_overlap_r}
\end{figure}

In Figure~\ref{fig:opr_vs_overlap_r}, we further examine the effect of the overlap dimension $d_{\rms \wedge \rmw}$ on the OPR, fixing $n=500$ and $N=500$.
We observe that the OPR decreases as $d_{\rms \wedge \rmw}$ increases, confirming that W2S is most beneficial when the weak and strong feature representations share little subspace.
When $d_{\rms \wedge \rmw}$ is small, W2S can leverage information from directions that are inaccessible to the strong model alone, leading to significant reductions in excess decision risk.

\begin{figure}[htbp]
    \centering
    \includegraphics[width=0.85\linewidth]{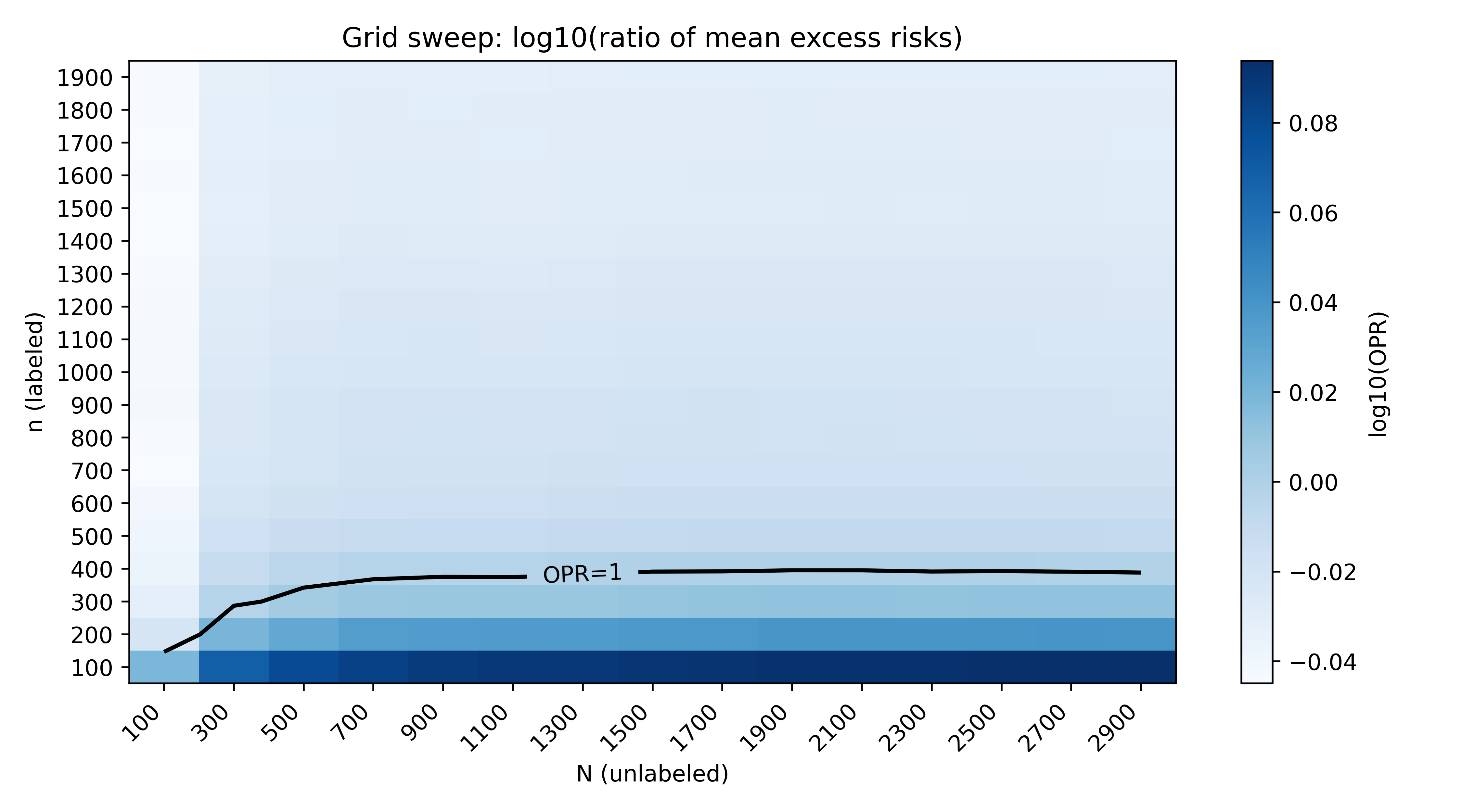}
    \caption{
    Grid sweep of the OPR over labeled and unlabeled sample sizes.
    Each cell reports $\log(\mathrm{OPR})$, where OPR is computed as the ratio of mean excess risks across 60 independent trials, at overlap $d_{\rms \wedge \rmw}=2$.
    Darker (more positive) values indicate $\mathrm{OPR}>1$, i.e., W2S achieves lower excess decision risk than the strong-only baseline; lighter (more negative) values indicate $\mathrm{OPR}<1$.
    The overlaid contour highlights the level set $\mathrm{OPR}=1$, separating the two regimes.
    }
    \label{fig:opr_heatmap_grid}
\end{figure}

Figure~\ref{fig:opr_heatmap_grid} presents an exhaustive grid sweep of the OPR over labeled and unlabeled sample sizes at fixed overlap $d_{\rms \wedge \rmw}=2$.
It reinforces the earlier observations that W2S excels when labeled data are limited and unlabeled data are abundant, e.g., for the region $n \leq 300$, we see the largest gains.
At moderate labeled sizes $n\approx 500$, the improvement is smaller but still visible.
The solid contour marks the boundary $\mathrm{OPR}=1$, indicating that beyond roughly $n\sim 10^3$ the advantage of W2S becomes marginal.

For ease of exposition, we focus in this section on the unconstrained single-item newsvendor problem. In Appendix~\ref{appendix:two_product_newsvendor}, we consider a multi-item newsvendor problem with a capacity constraint, and see that the empirical results remain consistent with the theory, in the constrained case as well.

\subsection{Comment Moderation Experiments}
\label{subsec:comment_moderation_experiments}

In this section, we evaluate W2S on a more realistic comment moderation task, a canonical application in AI-assisted service systems \citep{lee2024design}.
Different from the synthetic experiment, this experiment does not enforce any of the assumptions used in the theory.

\paragraph{Background.}
An online platform receives a stream of comments: most are harmless, some are toxic, and a small fraction are severely toxic \citep{borkan2019nuanced,kumar2021designing,siegelmann2024mico}.
The operational task is therefore to select, for each comment, one of several moderation actions: approve it, remove it, or route it to a designated review queue. These queues represent different levels of human attention and urgency, ranging from standard review and delayed holding to priority review.
{
More broadly, this routing structure reflects the human-in-the-loop design of modern AI agent systems, in which automated agents handle routine cases while ambiguous or high-risk cases are escalated to human reviewers with different priority levels.
}
The central tradeoff is between the harm caused by exposing users to severely toxic content and the friction created by incorrectly removing harmless content. Accordingly, the value of a predictor is determined not only by its classification accuracy but also by the downstream cost of the moderation action it induces. This setting naturally gives rise to a labeled–unlabeled data split: raw comments are abundant in platform traffic, whereas reliable toxicity-severity labels require costly expert judgments.

\paragraph{Setup.}
We use the Jigsaw Unintended Bias/Civil Comments corpus, which contains online comments with crowd-sourced toxicity and identity annotations \citep{borkan2019nuanced,jigsaw2019unintended}.
The context $x$ is the comment text.
The raw data reports a toxicity score for each comment, which we discretize into three severity classes $\Xi=\{0,1,2\}$ using thresholds $0.5$ and $0.8$: non-toxic, toxic, and severely toxic.
The feasible action set is
\[
    \mathcal{A}
    =
    \{\text{allow},\text{remove},\text{standard review},\text{hold review},\text{priority review}\}.
\]
Here, \textit{allow} means that the comment is automatically approved and remains visible on the platform.
\textit{Remove} means that the comment is automatically suppressed without human review.
\textit{Standard review} sends the comment to a regular human-review queue while the comment remains visible during the review process.
\textit{Hold review} also sends the comment to human review, but the comment is held off the platform until the review is completed.
\textit{Priority review} is a faster human-review route for comments that may be more urgent; it has higher review cost but shorter delay.
These actions trade off different types of operational cost: exposure cost from leaving harmful comments visible, user-friction cost from suppressing benign comments, reviewer cost, and delay cost from human review.

The original data set is highly imbalanced, with approximately 14\% toxic comments and only 2.6\% severely toxic comments.
To address this imbalance, we augment the data by prompting Claude Sonnet, using real severe-toxicity comments as style anchors, to generate additional plausible CivilComments-style comments.
In the augmented data set, the severely toxic class is increased to about 10\% of the total comments.
Then, we further partition the augmented comment data into separate splits for (1) task pre-training, (2) downstream W2S training with labeled and unlabeled samples, and (3) held-out evaluation.

Putting in the notation of Section~\ref{sec:setting}, the random outcome is the severity class $\xi\in\Xi$, and the action is the moderation route $a\in\mathcal{A}$.
We encode the cost of assigning route $a$ to a comment with severity $\xi$ by a route-severity cost matrix $C$, so that $c(a,\xi)=C_{a,\xi}$.
The exact entries of $C$ are reported in Appendix~\ref{appendix:moderation_cost_matrix}; they are chosen to reflect the operational tradeoffs described above.
Given a conditional model $P:\mathcal{X}\to\Delta(\Xi)$, the induced plug-in route follows the decision rule used throughout the paper:
\(
    \tilde{w}(P(x))
    \in
    \arg\min_{a\in\mathcal{A}}
    \sum_{\xi\in\Xi} C_{a,\xi}P(\xi\mid x).
\)
We refer to this as the \emph{hard-route} decision.

\paragraph{Pre-training.}
To obtain the two feature maps $\phi_{\rmw}$ and $\phi_{\rms}$, we start from two generic language-model backbones with different capacities: a compact two-layer BERT encoder for the weak model and a DistilBERT encoder for the strong model \citep{devlin2019bert,sanh2019distilbert}.
To map each encoder representation to the three severity classes, we attach a newly initialized classification head to each model and task pre-train the resulting models end-to-end on labeled comments from the pre-training split using cross-entropy loss; sample-size and optimizer-level details are reported in Appendix~\ref{appendix:moderation_model_pretraining}.
To align the experiment with the weak-to-strong regime \citep{burns2024weak}, we restrict the task-specific pretraining compute allocated to the larger DistilBERT model to prevent it from already dominating the smaller model before weak supervision.

\paragraph{Training protocol.}
To validate the theoretical findings in Section~\ref{sec:comparison_bounds}, we compare W2S with the strong-only baseline across a range of labeled and unlabeled sample sizes. For each experimental setting, we draw a labeled sample $\tilde{\mathcal{B}}$ and an unlabeled sample $\mathcal{B}$ uniformly without replacement from their respective pools.
The strong-only benchmark is initialized from the strong pre-trained model and trained on $\tilde{\mathcal B} $ using the supervised decision-aware objective in \eqref{eq:strong_training_procedure}, with the optimal action oracle $\tilde{w}$ replaced by a differentiable soft-route surrogate to enable gradient-based optimization, in the same spirit as \eqref{eq:surrogate_oracle_synthetic}.
We emphasize here that in evaluation, we still use the hard-route decision.
It is optimized using AdamW with weight decay.
For W2S, we initialize the weak teacher from the weak pre-trained model and fine-tune it on $\tilde{\calB}$ by cross-entropy.
We then implement the W2S training in the same way as in \eqref{eq:w2s_surrogate_objective}.
It is optimized with AdamW with a cosine learning-rate scheduler.

\begin{figure}[!htbp]
    \centering
    \begin{subfigure}[t]{0.49\linewidth}
        \centering
        \includegraphics[width=\linewidth]{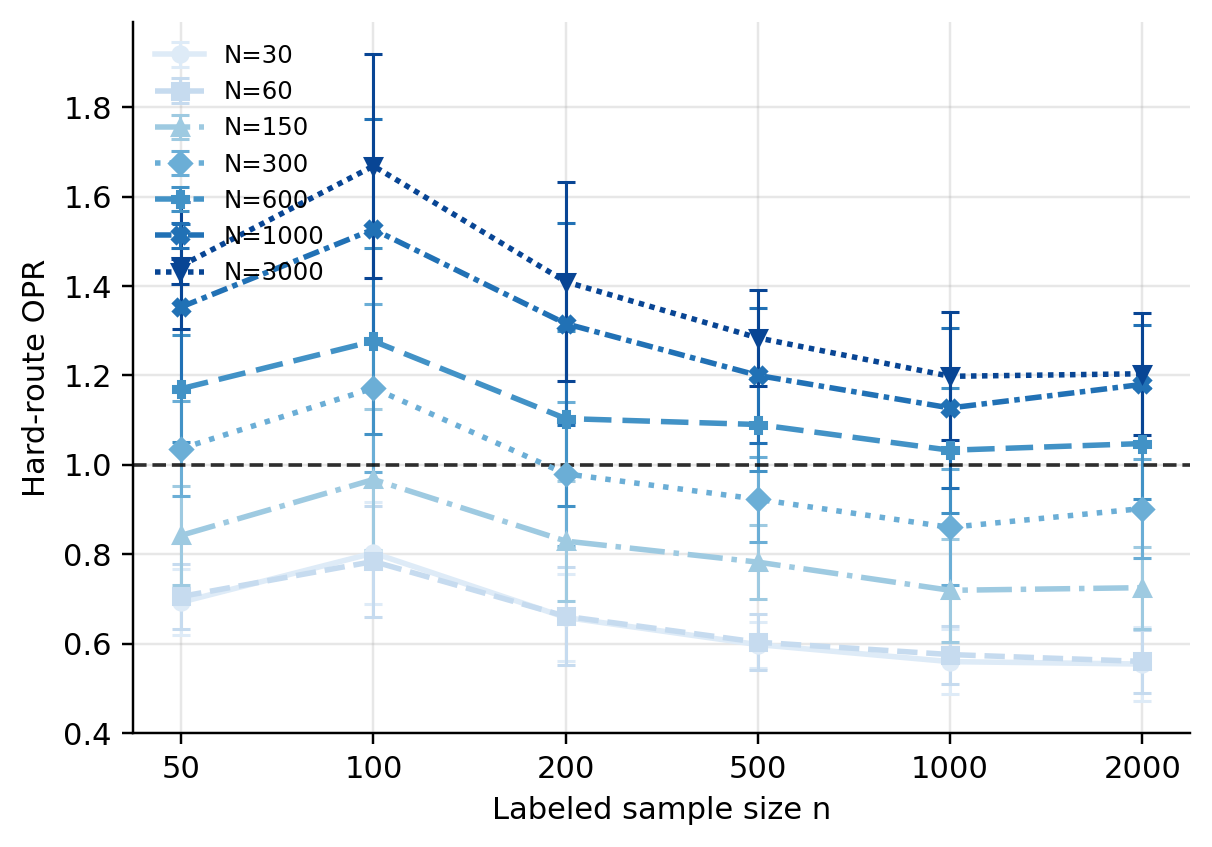}
        \caption{\textbf{OPR vs.\ labeled size $n$}.}
    \end{subfigure}
    \hfill
    \begin{subfigure}[t]{0.49\linewidth}
        \centering
        \includegraphics[width=\linewidth]{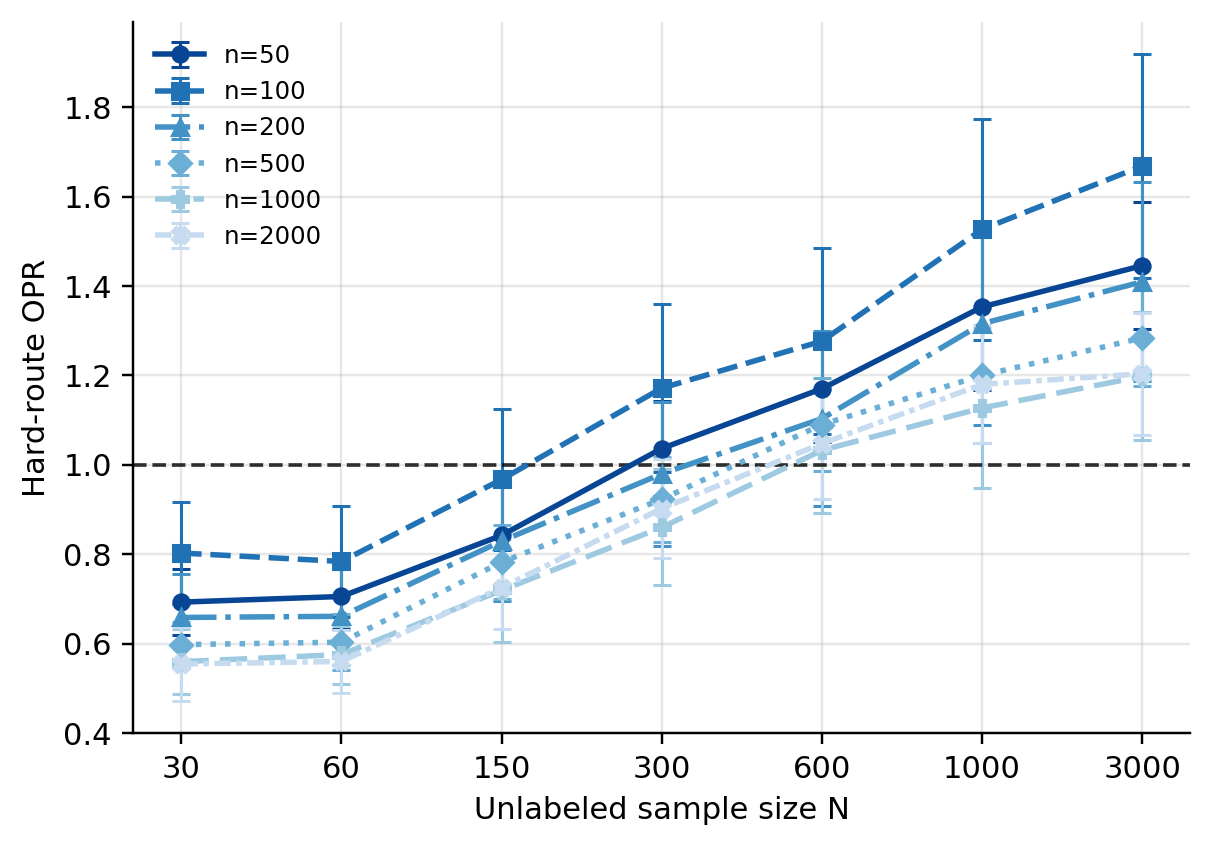}
        \caption{\textbf{OPR vs.\ unlabeled size $N$}.}
    \end{subfigure}
    \caption{
    Panels (a)--(b) plot empirical OPR across $30$ trials with $1.96$ bootstrap standard-error bars.
    }
    \label{fig:moderation_hard_route_opr}
\end{figure}

\paragraph{Results.}
Since we do not have access to the ground-truth distribution for computing the OPR in \eqref{eq:opr_def}, we compute the empirical OPR using the held-out split of the data set.
Figure~\ref{fig:moderation_hard_route_opr} shows how the labeled and unlabeled sample sizes affect the OPR in this experiment. We observe that, consistent with the qualitative implication of Corollary~\ref{cor:w2s_outperformance_certificate}, W2S does not improve over the strong-only baseline when the unlabeled sample size $N$ is small, but achieves OPR above one once $N$ is greater than 600.
Another finding consistent with the synthetic experiment is that the advantage of W2S is most pronounced when the labeled sample size $n$ is small, and the gain narrows as $n$ increases.

\section{Concluding Remarks}

This paper takes a first step toward understanding weak-to-strong learning as a decision-making problem.
In the contextual stochastic optimization setting, we show that weak supervision can improve downstream decision performance when labeled data are scarce and unlabeled contexts are abundant, with the overlap between weak and strong representations determining how much teacher error is inherited rather than averaged out.
The numerical experiments support this mechanism in both controlled simulations and a comment moderation routing task.
A natural next step is to move beyond one-shot decisions to sequential decision making, including reinforcement learning and online control. In such settings, weak supervision may affect not only the final decision rule, but also exploration, state distribution shift, and the accumulation of decision errors over time.

\newpage

\begin{APPENDICES}
\renewcommand{\theHsection}{appendix.\Alph{section}}
\renewcommand{\theHsubsection}{appendix.\Alph{section}.\arabic{subsection}}
\renewcommand{\theHsubsubsection}{appendix.\Alph{section}.\arabic{subsection}.\arabic{subsubsection}}
\renewcommand{\theHequation}{appendix.\arabic{equation}}
\renewcommand{\theHfigure}{appendix.\arabic{figure}}
\renewcommand{\theHtable}{appendix.\arabic{table}}

\section{Omitted Proofs in Section~\ref{sec:setting}}

\begin{lemma} \label{lemma:overlap_dimension_range}
    We always have $0 \leqslant d_{\rms \wedge \rmw} \leqslant \min \left\{d_{\rms}, d_{\rmw}\right\}$.
\end{lemma}

\begin{myproof}
        We define $\gamma_{\rms}(x) = \Lambda_{\rms}^{-\frac{1}{2}} V_{\rms}^{\top} \phi_{\rms}(x)$ and $\gamma_{\rmw}(x) = \Lambda_{\rmw}^{-\frac{1}{2}} V_{\rmw}^{\top} \phi_{\rmw}(x)$. One can verify that they are zero-mean random vectors with identity covariance matrix.
        Then by definition $ M \defeq \E{ \gamma_{\rms}(x) \gamma_{\rmw}(x)^{\top} }
        = \E{ \Lambda_{\rms}^{-\frac{1}{2}} V_{\rms}^{\top} \phi_{\rms}(x) \phi_{\rmw}(x)^{\top} V_{\rmw} \Lambda_{\rmw}^{-\frac{1}{2}} } = \Lambda_{\rms}^{-\frac{1}{2}} V_{\rms}^{\top}  \E{  \phi_{\rms}(x) \phi_{\rmw}(x)^{\top} } V_{\rmw} \Lambda_{\rmw}^{-\frac{1}{2}} $.

        We next show that $\normtwo{M} \leq 1$.
        First of all, by definition of the 2-norm and by invoking Cauchy-Schwarz inequality, we have $\normtwo{M} = \sup_{\normtwo{v} = 1} \normtwo{M v} = \sup_{ \normtwo{u} = \normtwo{v} = 1} \abs{u^{\top} M v}$.
        For any vectors $u \in \mathbb{R}^{d_{\rms}}$ and $v \in \mathbb{R}^{d_{\rmw}}$, we have
        $$
        u^{\top} M v = u^{\top} \E{ \gamma_{\rms} \gamma_{\rmw}^{\top} } v = \E{ u^{\top} \gamma_{\rms} \gamma_{\rmw}^{\top} v } .
        $$
        By Cauchy-Schwarz inequality, we have $\abs{ \E{ u^{\top} \gamma_{\rms} \gamma_{\rmw}^{\top} v } } \leq \sqrt{\E{ \paren{u^{\top} \gamma_{\rms}}^2 }} \sqrt{ \E{ \paren{\gamma_{\rmw}^{\top} v }^2 } }$.
        Because both $\gamma_{\rms}$ and $\gamma_{\rmw}$ are isotropic, we have $\sqrt{\E{ \paren{u^{\top} \gamma_{\rms}}^2 }} = \normtwo{u}$ and $\sqrt{ \E{ \paren{\gamma_{\rmw}^{\top} v }^2 } } = \normtwo{v}$.
        Hence, we conclude that $\normtwo{M} = \sup_{ \normtwo{u} = \normtwo{v} = 1} \E{ u^{\top} \gamma_{\rms} \gamma_{\rmw}^{\top} v } \leq 1$.

        Now, we proceed to upper bound the overlap dimension.
        We have
        \begin{eqnarray}
            d_{\rms \wedge \rmw} &=& \norm{ \Sigma_{\rms}^{- \frac{1}{2}} \E{ \phi_{\rms}(x) \phi_{\rmw}(x)^{\top} } \Sigma_{\rmw}^{- \frac{1}{2}} }_F^2 \nonumber \\
            &=& \norm{ V_{\rms} \Lambda_{\rms}^{-\frac{1}{2}} V_{\rms}^{\top} \E{ \phi_{\rms}(x) \phi_{\rmw}(x)^{\top} } V_{\rmw} \Lambda_{\rmw}^{-\frac{1}{2}} V_{\rmw}^{\top} }_F^2 \label{eq:using_Sigma_half} \\
            &=& \norm{ V_{\rms} M V_{\rmw}^{\top}  }_F^2 \nonumber \\
            &=& \tr{ V_{\rms} M V_{\rmw}^{\top} V_{\rmw} M^{\top} V_{\rms}^{\top} } \label{eq:using_F_norm_tr}  \\
            &=& \tr{ M M^{\top} V_{\rms}^{\top} V_{\rms} }  \nonumber \\
            &=& \tr{ M M^{\top}  }  .
        \end{eqnarray}
        Equation~\eqref{eq:using_Sigma_half} follows from the definition of $\Sigma_{\rms}^{-\frac{1}{2}}$ and $\Sigma_{\rmw}^{-\frac{1}{2}}$.
        Equation~\eqref{eq:using_F_norm_tr} is due to the fact that $\norm{A}_F^2 = \tr{A A^{\top}}$ for any matrix $A$.
        The last equation follows from the fact that $V_{\rms}^{\top} V_{\rms} = I$.

        Finally, we note that $\tr{M M^{\top}} = \norm{M}_F^2 = \sum_{i=1}^{\min \{d_{\rms}, d_{\rmw}\}} \sigma_i^2$, where $\sigma_i$ is the $i$-th singular value of $M$.
        Because $\normtwo{M} \leq 1$, we have $\sigma_i \leq 1$ for all $i$.
        Hence, the proof is complete.
 \end{myproof}

\section{Omitted Proofs in Section~\ref{sec:upper_bound_w2s}}

\subsection{Proof of Lemma~\ref{lemma:convex_smooth_expected_cost}}
\label{appendix:proof_lemma_convex_smooth_expected_cost}
\begin{myproof}
    First of all, since $c$ is $\mu$-strongly convex and $L$-smooth in its first argument, we have
    \begin{equation}    \label{eq:a_a_prime_strong_convex_smooth}
        \frac{\mu}{2} \norm{ a' - a }_2^2
        \leq c( a' , z ) - c( a , z ) - \inner{ \nabla_a c( a , z ) , a' - a }
        \leq \frac{L}{2} \norm{ a' - a }_2^2
    \end{equation}
    for every $z \in \Xi$ and $a,a' \in \calA$.
    Multiplying by \eqref{eq:a_a_prime_strong_convex_smooth} by $P( \xi = z_k \mid x )$ and summing over $k=1,2,\ldots,K$, we have
    \begin{eqnarray}
        \frac{\mu}{2}\left\|a^{\prime}-a\right\|_2^2
        \leq \sum_{k=1}^K P\left(\xi=z_k \mid x\right)\left[c\left(a^{\prime}, z_k\right)-c\left(a, z_k\right)-\left\langle\nabla_a c\left(a, z_k\right), a^{\prime}-a\right\rangle\right]
        \leq \frac{L}{2}\left\|a^{\prime}-a\right\|_2^2 .
    \end{eqnarray}
    The middle term can be written as
    \begin{eqnarray}
        && \sum_{k=1}^K P\left(\xi=z_k \mid x\right)\left[c\left(a^{\prime}, z_k\right)-c\left(a, z_k\right)\right]
        - \left\langle\sum_{k=1}^K P\left(\xi=z_k \mid x\right) \nabla_a c\left(a, z_k\right), a^{\prime}-a\right\rangle  \nonumber \\
        &=&   \mathbb{E}_{ \xi \sim P  (\cdot \mid x ) } \brackets{
            c \paren{ a' , \xi } - c \paren{ a , \xi }
        }
        - \left\langle
            \sum_{k=1}^K P\left(\xi=z_k \mid x\right) \nabla_a c\left(a, z_k\right), a^{\prime}-a
        \right\rangle \nonumber
    .
    \end{eqnarray}
    Taking $a' = \tilde{w} ( P' \mid x )$ and $a = \tilde{w} ( P \mid x )$, we have
    \begin{eqnarray}
        && \frac{\mu}{2} \norm{ \tilde{w} ( P '\mid x ) - \tilde{w} ( P \mid x ) }_2^2 \nonumber \\
        &\leq& \mathbb{E}_{ \xi \sim P  (\cdot \mid x ) } \brackets{
        c \paren{ \tilde{w}\left( P' \mid x \right), \xi }
        - c \paren{ \tilde{w} \paren{ P \mid x } , \xi }
    }   \nonumber \\
        && - \inner{ \sum_{k=1}^K P ( \xi = z_k \mid x ) \nabla_w c( \tilde{w}\left( P \mid x \right) , z_k ) , \tilde{w}\left( P' \mid x \right) - \tilde{w}\left( P \mid x \right) }  \nonumber \\
        &\leq&  \frac{L}{2} \norm{ \tilde{w} ( P '\mid x ) - \tilde{w} ( P \mid x ) }_2^2  .
    \end{eqnarray}
    To conclude the proof, we recall the definition of $\tilde{w} ( P\mid x )$, which is given by
    $
    \tilde{w} ( P \mid x ) = \operatorname{argmin}_{ w \in \calA } \sum_{k=1}^K P ( \xi = z_k \mid x ) c( w , z_k ) .
    $
    The first-order optimality condition implies that
    \begin{equation}
        \sum_{k=1}^K P ( \xi = z_k \mid x ) \nabla_w c( \tilde{w} ( P \mid x ) , z_k ) = 0 .
    \end{equation}

\end{myproof}

\subsection{Proof of Lemma~\ref{lemma:decision_to_logit_continuity}}
\label{appendix:proof_lemma_decision_to_logit_continuity}

\begin{myproof}
Fix $x \in \mathcal{X}$. For brevity, write
$w=\tilde{w}(P_{\eta}\mid x)$, $w'=\tilde{w}(P_{\eta'}\mid x)$,
$p_k=P_{\eta}(\xi=z_k\mid x)$, and $p'_k=P_{\eta'}(\xi=z_k\mid x)$.
Since $w$ and $w'$ minimize their respective probability-weighted costs over the convex set $\mathcal{A}$, the variational inequalities give
\begin{equation} \label{eq:vi_revised_decision_to_logit}
    \inner{\sum_{k=1}^K p_k \nabla_w c(w,z_k), w'-w} \geq 0
    \quad\text{and}\quad
    \inner{\sum_{k=1}^K p'_k \nabla_w c(w',z_k), w-w'} \geq 0 .
\end{equation}
By the $\mu$-strong convexity of each $c(\cdot,z_k)$ and \eqref{eq:vi_revised_decision_to_logit},
\begin{align}
    \sum_{k=1}^K p_k \left[c(w',z_k)-c(w,z_k)\right]
    &\geq \inner{\sum_{k=1}^K p_k\nabla_w c(w,z_k),w'-w}
    +\frac{\mu}{2}\norm{w'-w}_2^2 \nonumber\\
    &\geq \frac{\mu}{2}\norm{w'-w}_2^2 .
    \label{eq:strong_convexity_eta_to_eta_prime}
\end{align}
Similarly, by the $\mu$-strong convexity of each $c(\cdot,z_k)$ and \eqref{eq:vi_revised_decision_to_logit},
\begin{align}
    \sum_{k=1}^K p'_k \left[c(w,z_k)-c(w',z_k)\right]
    &\geq \inner{\sum_{k=1}^K p'_k\nabla_w c(w',z_k),w-w'}
    +\frac{\mu}{2}\norm{w'-w}_2^2 \nonumber\\
    &\geq \frac{\mu}{2}\norm{w'-w}_2^2 .
    \label{eq:strong_convexity_eta_prime_to_eta}
\end{align}
Adding \eqref{eq:strong_convexity_eta_to_eta_prime} and \eqref{eq:strong_convexity_eta_prime_to_eta}, and using Part~(2) of Assumption~\ref{assumption:cost_function}, we obtain
\begin{align}
    \mu\norm{w'-w}_2^2
    &\leq \sum_{k=1}^K (p_k-p'_k)\left[c(w',z_k)-c(w,z_k)\right] \nonumber\\
    &\leq \sum_{k=1}^K \abs{p_k-p'_k}\abs{c(w',z_k)-c(w,z_k)} \nonumber\\
    &\leq G\norm{P_{\eta}(\cdot\mid x)-P_{\eta'}(\cdot\mid x)}_1\norm{w'-w}_2 .
    \label{eq:probability_to_action_revised}
\end{align}
If $w=w'$, the first desired inequality is immediate. Otherwise, dividing \eqref{eq:probability_to_action_revised} by $\mu\norm{w'-w}_2$ gives
\begin{equation} \label{eq:action_probability_lipschitz_revised}
    \norm{w'-w}_2
    \leq \frac{G}{\mu}\norm{P_{\eta}(\cdot\mid x)-P_{\eta'}(\cdot\mid x)}_1 .
\end{equation}
By Cauchy--Schwarz and the fact that the softmax map is $1/2$-Lipschitz from $\ell_2$ to $\ell_2$ \citep{nair2025softmax},
\begin{align}
    \norm{P_{\eta}(\cdot\mid x)-P_{\eta'}(\cdot\mid x)}_1
    &\leq \sqrt{K}\norm{P_{\eta}(\cdot\mid x)-P_{\eta'}(\cdot\mid x)}_2 \\
    &\leq \frac{\sqrt{K}}{2}\norm{\eta(x)-\eta'(x)}_2 .
    \label{eq:probability_logit_lipschitz_revised}
\end{align}
Combining \eqref{eq:action_probability_lipschitz_revised} and \eqref{eq:probability_logit_lipschitz_revised} proves the result.
\end{myproof}

\subsection{Statement and proof of Lemma~\ref{lemma:w2s_action_btw_w2s_and_weak}}
\label{appendix:proof_lemma_w2s_action_btw_w2s_and_weak}

\begin{lemma} \label{lemma:w2s_action_btw_w2s_and_weak}
    We have
    \begin{eqnarray}
        && \EE{ \calB, \tilde{\calB} }{ \mathbb{E}_{x} \brackets{ \normtwo{
        \tilde{w} \paren{ P^{\mathrm{s}}_{\widehat{\Theta}_{\mathrm{w2s}}} \mid x  }
        - \tilde{w} \paren{ P^{\mathrm{w}}_{ \widehat{\Theta}_{\mathrm{w}}} \mid x  }
        }^2 } } \nonumber \\
        &\leq& \frac{1}{4} \frac{ L K G^2 }{\mu^3 }
        \EE{ \tilde{\calB} , \calB }{
            \mathbb{E}_{x} \brackets{
                \norm{ \eta^{\mathrm{s}}_{ \widehat{\Theta}_{\mathrm{ls}}} (x) - \eta^{\mathrm{w}}_{\widehat{\Theta}_{\mathrm{w}}} (x) }_2^2
            }
        } + 8 \sqrt{2} \frac{ K G^2 B \sqrt{\tr{\Sigma_{\rms}}} }{ \mu^2}  \frac{1}{ \sqrt{N} }
        + \frac{4 C_{\mathrm{unif}}}{\mu} \frac{1}{\sqrt{N}}  .
    \end{eqnarray}

\end{lemma}

\begin{myproof}

To proceed, we define the following population and empirical estimated costs associated with the W2S procedure:
\begin{equation}
    \mathcal{L}_{\mathrm{w2s}}^{ \widehat{\Theta}_{\mathrm{w}}  }(\Theta)
    \defeq \mathbb{E}_x \mathbb{E}_{\xi \sim P_{\widehat{\Theta}_{\mathrm{w}}}^{\mathrm{w}}(\cdot \mid x)}\left[c\left(\tilde{w}\left(P_{\Theta}^{\mathrm{s}} \mid x\right), \xi\right)\right] ,
\end{equation}
and
\begin{equation}
    \widehat{\mathcal{L}}_{\mathrm{w2s}}^{ \widehat{\Theta}_{\mathrm{w}}  }(\Theta)
    \defeq \frac{1}{N}
    \sum_{ j=1 }^{N} \mathbb{E}_{\xi \sim P_{ \widehat{\Theta}_{\mathrm{w}}}^{\mathrm{w}}(\cdot \mid x_j)}
    \left[c\left(\tilde{w}\left(P^{\mathrm{s}}_\Theta \mid x_j\right), \xi\right)\right] .
\end{equation}
Let
\[
    \Theta_{\mathrm{w2s}}
    \in \argmin_{\Theta\in\mathcal{H}_B}
    \mathcal{L}_{\mathrm{w2s}}^{\widehat{\Theta}_{\mathrm{w}}}(\Theta)
\]
denote a population minimizer of the W2S objective.

Conditioned on the labeled data $\tilde{\calB}$ used to train the weak teacher $\widehat{\Theta}_{\mathrm{w}}$, we first note that by Lemma~\ref{lemma:convex_smooth_expected_cost}, we have
\begin{eqnarray}
    && \EE{\calB \mid \tilde{\calB}} {\frac{\mu}{2} \mathbb{E}_{x} \brackets{ \norm{ \tilde{w} ( P^{\mathrm{s}}_{ \widehat{\Theta}_{\mathrm{w2s}} } \mid x ) - \tilde{w} ( P^{\mathrm{w}}_{\widehat{\Theta}_{\mathrm{w}}} \mid x ) }_2^2 } \mid \tilde{\calB} } \nonumber \\
    &\leq& \EE{\calB \mid \tilde{\calB}} { \mathbb{E}_{x} \brackets{
        \mathbb{E}_{ \xi \sim P^{\mathrm{w}}_{\widehat{\Theta}_{\mathrm{w}}} (\cdot \mid x ) } \brackets{
    c \paren{ \tilde{w}\left(P^{\mathrm{s}}_{\widehat{\Theta}_{\mathrm{w2s}}} \mid x \right), \xi }
    - c \paren{ \tilde{w} \paren{ P^{\mathrm{w}}_{\widehat{\Theta}_{\mathrm{w}}} \mid x } , \xi }
    } } \mid \tilde{\calB} } \\
    &=& \EE{\calB \mid \tilde{\calB}}{ \mathcal{L}_{\mathrm{w2s}}^{\widehat{\Theta}_{\mathrm{w}}} \paren{ \widehat{\Theta}_{\mathrm{w2s}} }
    -  \mathbb{E}_{x} \brackets{ \mathbb{E}_{ \xi \sim P^{\mathrm{w} }_{\widehat{\Theta}_{\mathrm{w}}} (\cdot \mid x ) } \brackets{ c \paren{ \tilde{w} \paren{ P^{\mathrm{w}}_{\widehat{\Theta}_{\mathrm{w}}} \mid x } , \xi } } } \mid \tilde{\calB} } .
\end{eqnarray}

Moreover, in view of Lemma~\ref{lemma:w2s_population_empirical_gap}, we can upper bound the quantity $ \mathcal{L}_{\mathrm{w2s}}^{\widehat{\Theta}_{\mathrm{w}}} \paren{ \widehat{\Theta}_{\mathrm{w2s}} }
    - \mathcal{L}_{\mathrm{w2s}}^{\widehat{\Theta}_{\mathrm{w}}} \paren{  {\Theta}_{\mathrm{w2s}} }$.
Using the uniform absolute-deviation bound in Lemma~\ref{lemma:w2s_population_empirical_gap}, we obtain
\begin{eqnarray}
    &&  \EE{ \calB \mid \tilde{\calB}} { \mathcal{L}_{\mathrm{w2s}}^{\widehat{\Theta}_{\mathrm{w}}} \paren{ \widehat{\Theta}_{\mathrm{w2s}} }
    - \mathcal{L}_{\mathrm{w2s}}^{\widehat{\Theta}_{\mathrm{w}}} \paren{  {\Theta}_{\mathrm{w2s}} } } \nonumber \\
    &=& \EE{ \calB \mid \tilde{\calB}} {
         \mathcal{L}_{\mathrm{w2s}}^{\widehat{\Theta}_{\mathrm{w}}} \paren{ \widehat{\Theta}_{\mathrm{w2s}} }
    - \widehat{\mathcal{L}}_{\mathrm{w2s}}^{\widehat{\Theta}_{\mathrm{w}}} \paren{ \widehat{\Theta}_{\mathrm{w2s}} }
    + \widehat{\mathcal{L}}_{\mathrm{w2s}}^{\widehat{\Theta}_{\mathrm{w}}} \paren{ \widehat{\Theta}_{\mathrm{w2s}} }
    - \widehat{\mathcal{L}}_{\mathrm{w2s}}^{\widehat{\Theta}_{\mathrm{w}}} \paren{  {\Theta}_{\mathrm{w2s}} }
    + \widehat{\mathcal{L}}_{\mathrm{w2s}}^{\widehat{\Theta}_{\mathrm{w}}} \paren{  {\Theta}_{\mathrm{w2s}} }
    - \mathcal{L}_{\mathrm{w2s}}^{\widehat{\Theta}_{\mathrm{w}}} \paren{  {\Theta}_{\mathrm{w2s}} }
    } \nonumber \\
    &\leq& 2 \EE{ \calB \mid \tilde{\calB}} { \sup_{\Theta \in \mathcal{H}_B} \abs{ \mathcal{L}_{\mathrm{w2s}}^{ \widehat{\Theta}_{\mathrm{w}} } \paren{  {\Theta}  } - \widehat{\mathcal{L}}_{\mathrm{w2s}}^{ \widehat{\Theta}_{\mathrm{w}} } \paren{  {\Theta}  } } } \nonumber \\
    &\leq& 2 \sqrt{ \frac{ V_0(\widehat{\Theta}_{\mathrm{w}}) }{N} }
    + 4 \sqrt{2} \frac{G^2 KB}{ \mu} \frac{1}{ \sqrt{N} } \sqrt{  \tr{ \Sigma_{\rms} }   } \nonumber \\
    &\leq& \frac{2 C_{\mathrm{unif}}}{\sqrt{N}}
    + 4 \sqrt{2} \frac{G^2 KB}{ \mu} \frac{1}{ \sqrt{N} } \sqrt{  \tr{ \Sigma_{\rms} }   } .
\end{eqnarray}
Here, the first inequality uses the empirical optimality of $\widehat{\Theta}_{\mathrm{w2s}}$, and the final inequality uses $V_0(\widehat{\Theta}_{\mathrm{w}}) \leq C_{\mathrm{unif}}^2$.

On the other hand, in view of the optimality of $\Theta_{\mathrm{w2s}}$ with respect to the population objective $\mathcal{L}_{\mathrm{w2s}}^{\widehat{\Theta}_{\mathrm{w}}} (\cdot) $, we have
\begin{eqnarray}
    && \EE{\calB \mid \tilde{\calB}}{ \mathcal{L}_{\mathrm{w2s}}^{\widehat{\Theta}_{\mathrm{w}}} \paren{ {\Theta}_{\mathrm{w2s}}  }
    - \mathbb{E}_{x} \brackets{ \mathbb{E}_{ \xi \sim P^{\mathrm{w} }_{\widehat{\Theta}_{\mathrm{w}}} (\cdot \mid x ) } \brackets{ c \paren{ \tilde{w} \paren{ P^{\mathrm{w}}_{\widehat{\Theta}_{\mathrm{w}}} \mid x } , \xi } } } \mid \tilde{\calB} }  \nonumber \\
    &\leq& \EE{\calB \mid \tilde{\calB}}{ \mathcal{L}_{\mathrm{w2s}}^{\widehat{\Theta}_{\mathrm{w}}} \paren{ \widehat{\Theta}_{\mathrm{ls}}  }
- \mathbb{E}_{x} \brackets{ \mathbb{E}_{ \xi \sim P^{\mathrm{w} }_{\widehat{\Theta}_{\mathrm{w}}} (\cdot \mid x ) } \brackets{ c \paren{ \tilde{w} \paren{ P^{\mathrm{w}}_{\widehat{\Theta}_{\mathrm{w}}} \mid x } , \xi } } } \mid \tilde{\calB} } \label{eq:apply_opt_w2s_main} \\
    &\leq& \EE{\calB \mid \tilde{\calB}}{  \frac{L}{2} \mathbb{E}_{x} \brackets{ \norm{ \tilde{w} ( P^{\mathrm{s}}_{ \widehat{\Theta}_{\mathrm{ls}}} \mid x ) - \tilde{w} ( P^{\mathrm{w}}_{\widehat{\Theta}_{\mathrm{w}}} \mid x ) }_2^2 }  \mid \tilde{\calB}  }  \label{eq:apply_smooth_expected_cost_main}
\end{eqnarray}
where \eqref{eq:apply_smooth_expected_cost_main} again follows from Lemma~\ref{lemma:convex_smooth_expected_cost}.

Combining all above and invoking Lemma~\ref{lemma:decision_to_logit_continuity}, we get
\begin{eqnarray}
    && \EE{ \calB \mid \tilde{\calB} }{
         \mathbb{E}_{x} \brackets{ \normtwo{
        \tilde{w} \paren{ P^{\mathrm{s}}_{\widehat{\Theta}_{\mathrm{w2s}}} \mid x  }
        - \tilde{w} \paren{ P^{\mathrm{w}}_{ \widehat{\Theta}_{\mathrm{w}}} \mid x  }
        }^2 }  \mid \tilde{\calB}
    }  \nonumber \\
    &\leq& \frac{1}{4} \frac{L K G^2 }{\mu^3}
    \EE{ \calB \mid \tilde{\calB} }{
          \mathbb{E}_{x} \brackets{
            \norm{ \eta^{\mathrm{s}}_{ \widehat{\Theta}_{\mathrm{ls}}} (x) - \eta^{\mathrm{w}}_{\widehat{\Theta}_{\mathrm{w}}} (x) }_2^2
        }  \mid \tilde{\calB}
    } + 8 \sqrt{2} \frac{ K G^2 B \sqrt{\tr{\Sigma_{\rms}}} }{ \mu^2}  \frac{1}{ \sqrt{N} }
    + \frac{4 C_{\mathrm{unif}}}{\mu} \frac{1}{\sqrt{N}} .
\end{eqnarray}
Finally, taking expectation over $\tilde{\calB}$ on both sides concludes the proof.

\end{myproof}

\subsection{Statement and proof of Lemma~\ref{lemma:w2s_population_empirical_gap}}
\label{appendix:proof_lemma_w2s_population_empirical_gap}

\begin{lemma} \label{lemma:w2s_population_empirical_gap}
    Define
    \(
        C_{\mathrm{unif}}
        \defeq
        \max_{k\in[K]}
        \abs{c\paren{\tilde{w}\paren{\operatorname{softmax}(0)},z_k}} .
    \)
    For any fixed $\widehat{\Theta}_{\mathrm{w}}$, we have
    \begin{equation}
        \EE{ \calB \mid \tilde{\calB}} {\sup_{\Theta \in \mathcal{H}_B}
        \abs{ \mathcal{L}_{\mathrm{w2s}}^{ \widehat{\Theta}_{\mathrm{w}} } \paren{  {\Theta}  } - \widehat{\mathcal{L}}_{\mathrm{w2s}}^{ \widehat{\Theta}_{\mathrm{w}} } \paren{  {\Theta}  } } }
        \leq \frac{C_{\mathrm{unif}}}{\sqrt N}
        + 2\sqrt{2}  \frac{G^2 KB}{  \mu} \frac{1}{ \sqrt{N} } \sqrt{  \tr{ \Sigma_{\rms} }   } .
    \end{equation}
\end{lemma}

\begin{myproof}

For brevity, we let $\phi(x) = \phi_{\rms}(x)$ be the strong feature mapping.
Given the weak teacher $P^{\mathrm{w}}_{\widehat{\Theta}_{\mathrm{w}}}$, denote $Q_k(x) = P^{\mathrm{w}}_{\widehat{\Theta}_{\mathrm{w}}} (\xi = z_k \mid x )$.
Given $\Theta$, the function $g_{\Theta}: \mathcal{X} \rightarrow \mathbb{R}$ defined by
\begin{eqnarray*}
g_{\Theta} (x)
&\defeq& \sum_{k=1}^{K} Q_k(x)
c \paren{ \tilde{w} \paren{ \operatorname{softmax} \paren{ \Theta^{\top} \phi(x) }  } , z_k } \\
&=& \EE{ \xi \sim P^{\rmw}_{\widehat{\Theta}_{\rmw}}(\cdot \mid x) }{ c \paren{ \tilde{w} \paren{ \operatorname{softmax} \paren{ \Theta^{\top} \phi(x) }  } , \xi  }  }
\end{eqnarray*}
is $\frac{G^2 \sqrt{K} }{2 \mu}  \norm{\phi (x) }_2$-Lipschitz with respect to $\Theta$.
Indeed, for any $\Theta, \Theta'$, denote $w = \tilde{w} \paren{ \operatorname{softmax} \paren{ \Theta^{\top} \phi(x) } }$, $w' = \tilde{w} \paren{ \operatorname{softmax} \paren{ (\Theta')^{\top} \phi(x) } }$.
Then we have
\begin{eqnarray}
    \abs{ g_{\Theta} (x) - g_{\Theta'} (x) } &=& \abs{ \sum_{k=1}^{K} Q_k(x) \paren{ c (w, z_k) - c(w', z_k) } }  \nonumber \\
    &\leq& \sum_{k=1}^{K} Q_k(x) \abs{ c (w, z_k) - c(w', z_k) } \nonumber \\
    &\leq& G \sum_{k=1}^{K} Q_k(x) \norm{ w - w' }_2  \label{eq:use_c_to_action_in_gen_w2s} \\
    &\leq& \frac{G^2 \sqrt{K}}{2 \mu} \norm{ \paren{ \Theta - \Theta'}^{\top} \phi(x)  }_2 \label{eq:use_action_to_logits_in_gen_w2s} \\
    &\leq& \frac{G^2 \sqrt{K}}{2 \mu} \norm{\phi (x) }_2 \norm{ \Theta - \Theta' }_2 .
\end{eqnarray}
Inequality~\eqref{eq:use_c_to_action_in_gen_w2s} follows since $\normtwo{ \nabla_w c( w , z ) } \leq G$, and Inequality~\eqref{eq:use_action_to_logits_in_gen_w2s} is due to Lemma~\ref{lemma:decision_to_logit_continuity}.

By definition, we have
$
\mathcal{L}_{\mathrm{w2s}}^{ \widehat{\Theta}_{\mathrm{w}} } \paren{  {\Theta}  } - \widehat{\mathcal{L}}_{\mathrm{w2s}}^{ \widehat{\Theta}_{\mathrm{w}} } \paren{  {\Theta}  }
= \EE{x}{ g_{\Theta} (x) } - \frac{1}{N} \sum_{j=1}^{N} g_{\Theta} (x_j)
$.
For the proof, define
\[
g_0(x)
\defeq g_{\Theta=0}(x)
= \sum_{k=1}^K P^{\mathrm{w}}_{\widehat{\Theta}_{\mathrm{w}}}(\xi=z_k\mid x)c(\tilde{w}(\operatorname{softmax}(0)),z_k),
\qquad
V_0
\defeq \operatorname{Var}_{x}\paren{g_0(x)} ,
\]
and set $h_{\Theta}(x)=g_{\Theta}(x)-g_0(x)$.
Then
\begin{eqnarray}
    && \EE{ \calB \mid \tilde{\calB}} {\sup_{\Theta \in \mathcal{H}_B}
        \abs{ \EE{x}{ g_{\Theta} (x) } - \frac{1}{N} \sum_{j=1}^{N} g_{\Theta} (x_j) } } \nonumber \\
    &\leq& \EE{ \calB \mid \tilde{\calB}} {\abs{ \EE{x}{ g_0 (x) } - \frac{1}{N} \sum_{j=1}^{N} g_0 (x_j) }}
    + \EE{ \calB \mid \tilde{\calB}} {\sup_{\Theta \in \mathcal{H}_B}
        \abs{ \EE{x}{ h_{\Theta} (x) } - \frac{1}{N} \sum_{j=1}^{N} h_{\Theta} (x_j) } } \nonumber \\
    &\leq& \sqrt{ \frac{V_0}{N} }
    + \EE{ \calB \mid \tilde{\calB}} {\sup_{\Theta \in \mathcal{H}_B}
        \abs{ \EE{x}{ h_{\Theta} (x) } - \frac{1}{N} \sum_{j=1}^{N} h_{\Theta} (x_j) } } .
    \label{eq:w2s_pop_emp_gap_centering}
\end{eqnarray}
The second inequality follows from Cauchy's inequality, since
\[
\EE{ \calB \mid \tilde{\calB}} {\abs{ \EE{x}{ g_0 (x) } - \frac{1}{N} \sum_{j=1}^{N} g_0 (x_j) }}
\leq
\sqrt{
\EE{ \calB \mid \tilde{\calB}} {
\paren{ \EE{x}{ g_0 (x) } - \frac{1}{N} \sum_{j=1}^{N} g_0 (x_j) }^2
}}
= \sqrt{ \frac{V_0}{N} } .
\]

By the standard symmetrization inequality \citep[Chapter~4]{wainwright2019high}, applied to the centered class $\braces{h_{\Theta}:\Theta\in\mathcal H_B}$, we have
\begin{equation}
    \mathbb{E}_{\calB \mid \tilde{\calB}}
    \left[\sup _{\Theta \in \mathcal{H}_B}\left|\mathbb{E}_x\left[h_{\Theta}(x)\right]-\frac{1}{N} \sum_{j=1}^N h_{\Theta}\left(x_j\right)\right|\right]
    \leq 2 \mathbb{E}_{\calB, \sigma \mid \tilde{\calB}}\left[\sup _{\Theta \in \mathcal{H}_B} \left|\frac{1}{N}\sum_{j=1}^N \sigma_j h_{\Theta}\left(x_j\right)\right| \right] .
\label{eq:w2s_pop_emp_gap_symmetrization_normalized}
\end{equation}
The class $\braces{h_\Theta:\Theta\in\mathcal H_B}$ contains the zero function, because $h_0\equiv0$. Thus
\begin{equation}
\mathbb{E}_{\calB, \sigma \mid \tilde{\calB}}\left[\sup _{\Theta \in \mathcal{H}_B} \left|\frac{1}{N}\sum_{j=1}^N \sigma_j h_{\Theta}\left(x_j\right)\right| \right]
\leq
2\mathbb{E}_{\calB, \sigma \mid \tilde{\calB}}\left[\sup _{\Theta \in \mathcal{H}_B} \frac{1}{N}\sum_{j=1}^N \sigma_j h_{\Theta}\left(x_j\right) \right] .
\label{eq:w2s_pop_emp_gap_abs_to_one_sided}
\end{equation}
For each fixed $x_j$, the inner map $u\mapsto \sum_{k=1}^{K}Q_k(x_j)c(\tilde w(\operatorname{softmax}(u)),z_k)-g_0(x_j)$ is $\frac{G^2\sqrt K}{2\mu}$-Lipschitz in $u$ and vanishes at $u=0$.
By virtue of the vector contraction inequality (Lemma~\ref{lemma:vector_contraction_inequality}), we conclude that
\begin{eqnarray}
    \mathbb{E}_{\calB, \sigma \mid \tilde{\calB}}
    \brackets{ \sup _{\Theta \in \mathcal{H}_B} \frac{1}{N}\sum_{j=1}^N \sigma_j h_{\Theta}\left(x_j\right)  }
    &\leq& \sqrt{2} \frac{G^2 \sqrt{K}}{ 2 \mu}
    \mathbb{E}_{\calB, \sigma \mid \tilde{\calB}} \brackets{ \sup _{\Theta \in \mathcal{H}_B} \frac{1}{N}\sum_{j=1}^N \sum_{k=1}^{K} \sigma_{j,k} \brackets{ \Theta^{\top} \phi (x_j) }_k  } \nonumber \\
    &\leq& \sqrt{2} \frac{G^2 \sqrt{K}}{ 2 \mu}  \frac{B \sqrt{K}}{\sqrt{N}} \sqrt{  \tr{ \EE{}{ \phi(x) \phi(x)^{\top} } }   } , \label{eq:w2s_pop_emp_gap_contraction_normalized}
\end{eqnarray}
where the second inequality follows from Lemma~\ref{lemma:vector_rademacher_complexity_linear_feature}.
Combining \eqref{eq:w2s_pop_emp_gap_centering}, \eqref{eq:w2s_pop_emp_gap_symmetrization_normalized}, \eqref{eq:w2s_pop_emp_gap_abs_to_one_sided}, and \eqref{eq:w2s_pop_emp_gap_contraction_normalized} gives
\begin{eqnarray}
&& \EE{ \calB \mid \tilde{\calB}} {\sup_{\Theta \in \mathcal{H}_B}
\abs{ \mathcal{L}_{\mathrm{w2s}}^{ \widehat{\Theta}_{\mathrm{w}} } \paren{  {\Theta}  } - \widehat{\mathcal{L}}_{\mathrm{w2s}}^{ \widehat{\Theta}_{\mathrm{w}} } \paren{  {\Theta}  } } } \nonumber \\
&\leq& \sqrt{\frac{V_0}{N}}
+2\sqrt{2}\frac{G^2KB}{\mu}\sqrt{\frac{\tr{\Sigma_{\rms}}}{N}} .
\label{eq:w2s_pop_emp_gap_combined_with_v0}
\end{eqnarray}
Since $Q_k(x)\geq0$, $\sum_{k=1}^K Q_k(x)=1$, and $\abs{c(\tilde{w}(\operatorname{softmax}(0)),z_k)}\leq C_{\mathrm{unif}}$ for every $k$,
\[
\abs{g_0(x)}
\leq
\sum_{k=1}^K Q_k(x) C_{\mathrm{unif}}
= C_{\mathrm{unif}} .
\]
Therefore $V_0=\operatorname{Var}_{x}\paren{g_0(x)}\leq \EE{x}{g_0(x)^2}\leq C_{\mathrm{unif}}^2$. Substituting $V_0\leq C_{\mathrm{unif}}^2$ into \eqref{eq:w2s_pop_emp_gap_combined_with_v0} yields the claimed inequality.

\end{myproof}

\subsection{Proof of Lemma~\ref{lemma:bound_moments}}

\begin{myproof}
    \begin{itemize}
        \item We note that $\gamma_{\rms}(x)$ is a linear map of $\phi(x) = \left[\begin{array}{l}
    \phi_{\rms}(x) \\
    \phi_{\rmw}(x)
    \end{array}\right] $, with $\gamma_{\rms} = B_s \phi $, where
    $B_s =
    \left[\begin{array}{lll}
    \Lambda_{\rms}^{-1 / 2} V_{\rms}^{\top} & ~0
    \end{array}\right] $.
    Hence, for any $u \in \mathbb{R}^{d_s}$, we have
    \begin{equation}
        \E{ e^{ u^{\top} \gamma_{\rms}(x) }  } = \E{  e^{\left(B_s^{\top} u\right)^{\top} \phi} }
        \leq \exp \left(\frac{1}{2} c^2 u^{\top}\left(B_s \Sigma_\phi B_s^{\top}\right) u\right) .
    \end{equation}
    But $B_s \Sigma_\phi B_s^{\top}=\operatorname{Cov}\left(\gamma_{\rms}\right)$ and we know that $\operatorname{Cov}\left(\gamma_{\rms}\right)=I_{d_s}$.
    Lemma~\ref{lemma:joint_subgaussian_cov_fourth_moment} then implies that $\E{ \norm{ \gamma_{\rms} (x) }_{2}^{4}  }  \lesssim \paren{  \tr{ I_{d_s} } }^2 = d_s^2  $.

    \item We note that $\varepsilon(x)$ is a linear map of $\phi(x) $. To be precise, we have $\varepsilon = \gamma_{\rmw} - M^{\top} \gamma_{\rms} = \paren{ \left[\begin{array}{lll}
    0 & ~ \Lambda_{\rmw}^{-\frac{1}{2}} V_w^{\top}
    \end{array}\right]  - M^{\top} \left[\begin{array}{lll}
    \Lambda_{\rms}^{-1 / 2} V_{\rms}^{\top} & ~0
    \end{array}\right]
      } \phi
    \defeq B_{\varepsilon} \phi  $.
    Hence, for any $u \in \mathbb{R}^{d_w}$, we have
    \begin{equation}
        \E{ e^{u^{\top} \varepsilon} } = \E{  e^{u^{\top} B_{\varepsilon} \phi} }
        = \E{ e^{\left(B_{\varepsilon}^{\top} u\right)^{\top} \phi} }
        \leq \exp \left(\frac{1}{2} c^2 u^{\top}\left(B_{\varepsilon} \Sigma_\phi B_{\varepsilon}^{\top}\right) u\right) .
    \end{equation}
    We notice that $B_{\varepsilon} \Sigma_\phi B_{\varepsilon}^{\top} = \operatorname{Cov}(\varepsilon) = I_{d_w} - M^{\top} M \preceq I_{d_w}$, which implies that $\E{  \norm{\varepsilon(x)}_{2}^{4} }   \lesssim  \paren{ d_w - \norm{M}_F^2 }^2 $.

    \item We note that $\phi_{\rms}(x)$ is a linear map of $\phi(x)$, with $\phi_{\rms} = B_{\phi_{\rms}} \phi$, where $B_{\phi_{\rms}} = \left[\begin{array}{lll}I_{d_s} & ~0
\end{array}\right] $.
    Following the same argument as above suffices.
    \end{itemize}

\end{myproof}

\subsection{Proof of Lemma~\ref{lemma:upper_bound_expected_tr_w2s_term}}
\label{appendix:proof_lemma_upper_bound_expected_tr_w2s_term}

\begin{myproof}

    To start off the analysis, by algebraic calculations, we obtain that
    \begin{eqnarray}
        \tr{ \Gamma_{\rmw}^{\top} P_s \Gamma_{\rmw} } &=&  \tr{ \Gamma_{\rmw}^{\top} P_s^{\top} P_s \Gamma_{\rmw} } \nonumber \\
        &=& \norm{ P_s \Gamma_{\rmw} }_F^2 \label{eq:using_P_s_projection_PP_P}\\
        &=& \norm{ P_s \Gamma_s M + P_s E }_F^2 \nonumber \\
        &\leq& 2 \norm{  \Gamma_s M }_F^2 + 2 \norm{ P_s E }_F^2  \label{eq:gamma_w_Ps_gamma_w}  .
    \end{eqnarray}
    Equation~\eqref{eq:using_P_s_projection_PP_P} holds since $P_s$ is an orthogonal projection matrix.
    The last inequality follows from the fact that $\norm{A+B}_F^2 \leq 2 \norm{A}_F^2 + 2 \norm{B}_F^2$ for any matrices $A, B$ of the same dimension, and the fact that $P_s$ is an orthogonal projection matrix, which implies $\norm{P_s A}_F \leq \norm{A}_F$ for any matrix $A$.

    For the first term in \eqref{eq:gamma_w_Ps_gamma_w}, taking expectation over $\calB$ yields that
    \begin{eqnarray}
        \E{ \norm{  \Gamma_s M }_F^2 }
        = \E{ \tr{ M^{\top} \Gamma_s^{\top} \Gamma_s M } }
        =  \tr{ M^{\top} \E{\Gamma_s^{\top} \Gamma_s} M }
        = N \tr{ M^{\top} M }  \label{eq:Gamma_s_M_F_norm}
    \end{eqnarray}
    where the last step follows from the fact that $\E{ \Gamma_s^{\top} \Gamma_s } = \sum_{j=1}^{N} \E{ \gamma_{\rms} (x_j) \gamma_{\rms}(x_j)^{\top} } = N I_{d_s}$.

    To deal with the second term in \eqref{eq:gamma_w_Ps_gamma_w}, we need to handle the randomness of $P_s$. To this end, we define the good event to be
    \begin{equation}
        \calG = \braces{ \Gamma_{s}^{\top} \Gamma_{s} \succeq \frac{N}{4} I_{d_{\rms}}  } .
    \end{equation}
    The good event $\calG$ ensures that the matrix $\Gamma_s^{\top} \Gamma_s$ is invertible. It also happens with high probability.
    To be precise, there exist constants $c_{\text{NG}}, C_{\text{NG}} > 0$ depending on the sub-gaussian norm of $\phi_{\rms}(x)$, such that
    whenever $N \geq C_{\text{NG}} d_{\rms}$, we have
    \begin{equation} \label{eq:tr_bound_good_complement}
        \pr{ \calG^{\complement} } \leq 2 \exp \paren{ - c_{\text{NG}} N } .
    \end{equation}

    To see this, recall that the $i$-th row of $\Gamma_s$ is
$\gamma_{\rms}(x_i)^\top$. Since $x_1,\ldots,x_N$ are i.i.d.,
the rows of $\Gamma_s$ are i.i.d. sub-Gaussian random vectors.
Moreover, their second-moment matrix is
\[
\begin{aligned}
\E{\gamma_{\rms}(x_i)\gamma_{\rms}(x_i)^\top}
&=
\Lambda_{\rms}^{-1/2}V_{\rms}^\top
\Sigma_{\rms}
V_{\rms}\Lambda_{\rms}^{-1/2}
= I_{d_{\rms}}.
\end{aligned}
\]
Therefore, the rows of $\Gamma_s$ are isotropic.
    By applying Theorem~5.39 in \cite{vershynin2010introduction} (cf. Lemma~\ref{lemma:matrix_concentration_2_norm}) to the matrix $\Gamma_s^{\top} \Gamma_s$, we know that there exist absolute constants $c_{\text{NG}}$, $C_{\text{NG}}'$ depending on the sub-gaussian norm of the rows of $\Gamma_s$ such that for any $t > 0$, with probability at least $1 - 2 \exp(-c_{\text{NG}} t^2)$, we have
    \begin{equation}
        \normtwo{ \frac{1}{N} \Gamma_s^{\top} \Gamma_s - I_{d_{\rms}} } \leq \max \paren{ \delta , \delta^2} ,~ \delta = C_{\text{NG}}' \sqrt{ \frac{d_{\rms}}{N} } + \frac{t}{\sqrt{N}} .
    \end{equation}
    This implies that if $ \max \paren{ \delta , \delta^2} \leq \frac{3}{4}$, then
    \begin{equation}
        \frac{1}{N} \Gamma_s^{\top} \Gamma_s \succeq \frac{1}{4} I_{d_{\rms}}
    \end{equation}
    holds, which means the event $\calG$ happens.
    In order to ensure $ \max \paren{ \delta , \delta^2} \leq \frac{3}{4}$, it suffices to let $\delta \leq \frac{3}{4}$. This can be achieved by choosing $t = \sqrt{N}/4$ and $N \geq 4 C_{\text{NG}}' d_{\rms}$.
    The proof of \eqref{eq:tr_bound_good_complement} is now complete.

    Under event $\calG$, the matrix $\Gamma_s^{\top} \Gamma_s$ is invertible and hence we have
    \begin{equation} \label{eq:PsE_G}
        \norm{ P_s E}_F^2
        = \tr{ E^{\top} P_s^{\top} P_s E }
        = \tr{ E^{\top} \Gamma_s \paren{ \Gamma_s^{\top} \Gamma_s }^{-1} \Gamma_s^{\top} E }
        \leq \frac{4}{N} \norm{ \Gamma_s^{\top} E }_{F}^{2} .
    \end{equation}
    For the case when the event $\calG$ does not happen, we have
    \begin{equation} \label{eq:PsE_G_complement}
        \E{ \norm{ P_s E}_F^2 \one{\calG^{\complement}}}
        \leq  \E{ \norm{  E}_F^2 \one{\calG^{\complement}}}
        \leq \sqrt{ \E{ \norm{E}_F^4 } \pr{\calG^{\complement}} } .
    \end{equation}
    The first inequality follows from the fact that $\norm{P_s A}_F \leq \norm{A}_F$ for any matrix $A$ as $P_s$ is a projection matrix, and the second inequality is by Cauchy-Schwarz inequality.
    Combining \eqref{eq:PsE_G} and \eqref{eq:PsE_G_complement}, we obtain that
    \begin{eqnarray}
        \EE{\calB}{ \norm{ P_s E }_F^2   }
        &=& \EE{\calB}{ \norm{ P_s E }_F^2  \one{\calG} } + \EE{\calB}{ \norm{ P_s E }_F^2  \one{\calG^{\complement}} }  \nonumber \\
        &\leq&  \frac{4}{N} \EE{\calB}{  \norm{ \Gamma_s^{\top} E }_{F}^{2}   }
        + \sqrt{ \EE{\calB}{ \norm{E}_F^4 } \pr{\calG^{\complement}} }   \label{eq:Ps_E_decomposition}.
    \end{eqnarray}
    We proceed to control the two terms in \eqref{eq:Ps_E_decomposition} respectively.
    \begin{enumerate}
        \item For the first term, by algebraic calculations, we have
    \begin{equation}
        \norm{ \Gamma_s^{\top} E }_{F}^{2}
        =  \norm{ \sum_{j=1}^{N} \gamma_{\rms} (x_j) \varepsilon(x_j)^{\top} }_{F}^{2}
        =  \sum_{i=1}^{N} \sum_{j=1}^{N} \gamma_{\rms} (x_i)^{\top}  \gamma_{\rms}(x_j)\varepsilon(x_j)^{\top} \varepsilon(x_i) .
    \end{equation}
    For terms in the sum with $i \neq j$, taking expectation, we have
    \begin{eqnarray}
        && \E{ \gamma_{\rms} (x_i)^{\top}  \gamma_{\rms}(x_j) \varepsilon(x_j)^{\top} \varepsilon(x_i) } \nonumber \\
        &=& \E{ \sum_{a=1}^{d_s} [\gamma_{\rms} (x_i)]_a   [\gamma_{\rms}(x_j)]_a   \sum_{b=1}^{d_w}  [\varepsilon(x_j)]_b [\varepsilon(x_i)]_b  } \nonumber \\
        &=& \sum_{a=1}^{d_s} \sum_{b=1}^{d_w} \E{ [\gamma_{\rms} (x_i)]_a  [\varepsilon(x_i)]_b } \E{  [\gamma_{\rms}(x_j)]_a  [\varepsilon(x_j)]_b  }  \label{eq:using_independence_gamma_epsilon} \\
        &=& \sum_{a=1}^{d_s} \sum_{b=1}^{d_w} \paren{ \E{ [\gamma_{\rms} (x)]_a  [\varepsilon(x)]_b } }^2 \label{eq:using_identical_distribution_gamma_epsilon}   \\
        &=& \norm{ \E{ \gamma_{\rms} \varepsilon^{\top}} }_{F}^{2} \nonumber \\
        &=&  0 .
    \end{eqnarray}
    Equation~\eqref{eq:using_independence_gamma_epsilon} holds since $x_i$ and $x_j$ are independent for $i \neq j$.
    Equation~\eqref{eq:using_identical_distribution_gamma_epsilon} is due to the fact that $x_i$ and $x_j$ are identically distributed for any $i, j \in [N]$.
    The last step follows from \eqref{eq:expected_gamma_s_epsilon_zero}.
    Therefore, we conclude that
    \begin{equation}
        \E{\norm{ \Gamma_s^{\top} E }_{F}^{2} }
        = \sum_{i=1}^{N} \E{   \gamma_{\rms} (x_i)^{\top}  \gamma_{\rms}(x_i) \varepsilon(x_i)^{\top} \varepsilon(x_i) }
        = N \E{ \norm{ \gamma_{\rms} (x) }_{2}^{2} \norm{\varepsilon(x)}_{2}^{2} }  .
    \end{equation}
    After applying Cauchy-Schwarz inequality,
    $\E{ \norm{ \gamma_{\rms} (x) }_{2}^{2} \norm{\varepsilon(x)}_{2}^{2} } \leq \sqrt{ \E{ \norm{ \gamma_{\rms} (x) }_{2}^{4}  } } \sqrt{ \E{\norm{\varepsilon(x)}_{2}^{4} } }$, we can further use Lemma~\ref{lemma:bound_moments}.

    \item For the second term in \eqref{eq:Ps_E_decomposition}, we observe that since $\norm{E}_F^2 = \sum_{i=1}^{N} \norm{ \varepsilon(x_i) }_{2}^{2}$, it follows that
    $   \norm{ E }_{F}^{4} =  \paren{ \sum_{i=1}^{N} \norm{ \varepsilon(x_i) }_{2}^{2} }^2  = \sum_{i=1}^{N} \norm{ \varepsilon(x_i) }_{2}^{4} + 2 \sum_{i < j} \norm{ \varepsilon(x_i) }_{2}^{2} \norm{ \varepsilon(x_j) }_{2}^{2}   $.
    Hence, taking expectation yields that
    \begin{equation}
        \E{ \norm{ E }_{F}^{4} } = N \E{ \norm{ \varepsilon(x) }_{2}^{4} } + 2 N (N-1) \paren{ \E{ \norm{ \varepsilon(x) }_{2}^{2} } }^2 .
    \end{equation}
    Furthermore, we can calculate $\E{ \norm{ \varepsilon(x) }_{2}^{2} }$ as follows
    \begin{equation}
        \E{ \norm{\varepsilon(x)}_{2}^{2} } = \tr{ \E{ \varepsilon(x) \varepsilon(x)^{\top} } } = \tr{ I_{d_w} - M^{\top} M } = d_w - d_{\rms \wedge \rmw} .
    \end{equation}
    Then, it suffices to note that the event $\calG^{\complement}$ happens with probability at most $2 \exp \paren{ - c_{\text{NG}} N }$ as shown in \eqref{eq:tr_bound_good_complement}.

    \end{enumerate}

    Now, combining \eqref{eq:Gamma_s_M_F_norm} and \eqref{eq:Ps_E_decomposition} yields that
    \begin{eqnarray}
        && \EE{\calB}{ \tr{ \Gamma_{\rmw}^{\top} P_{\rms} \Gamma_{\rmw} } } \nonumber \\
        &\leq& 4 N \tr{ M^{\top} M } + \frac{2}{N} N \sqrt{ \E{ \norm{ \gamma_{\rms} (x) }_{2}^{4}  } } \sqrt{ \E{\norm{\varepsilon(x)}_{2}^{4} } } \nonumber \\
        &&  + 2 \sqrt{ N \E{ \norm{ \varepsilon(x) }_{2}^{4} } + 2 N (N-1) \paren{ \E{ \norm{ \varepsilon(x) }_{2}^{2} } }^2  } \sqrt{ 2 \exp(- c_{\text{NG}} N) }  \label{eq:using_moments_in_tr} \\
        &\lesssim&  N d_{\rms \wedge \rmw}  +  d_s  \paren{d_w - d_{\rms \wedge \rmw}  } + N  d_s  \paren{d_w - d_{\rms \wedge \rmw}  } \exp \paren{- \frac{1}{2} c_{\text{NG}} N }  ,
    \end{eqnarray}
    where we use Lemma~\ref{lemma:bound_moments} to conclude Inequality~\eqref{eq:using_moments_in_tr}.

\end{myproof}

\subsection{Statement and proof of Lemma~\ref{lemma:vector_rademacher_complexity_linear_feature} }

\begin{lemma} \label{lemma:vector_rademacher_complexity_linear_feature}
    Given a set $\mathcal{X}$, let $\phi: \mathcal{X} \rightarrow \mathbb{R}^d$.
    Let $ \mathcal{H}_B = \braces{ \Theta \in \mathbb{R}^{d \times K} : \norm{ \Theta }_F \leq B }$ for some constant $B>0$.
    Consider the vector-valued function class $ \mathcal{F} = \braces{ \Theta^{\top} \phi(x) : \Theta \in \mathcal{H}_B } $.
    Let $ \calB = \braces{ x_j }_{j=1}^{N} $ be $N$ i.i.d. samples from $\mathcal{D}_x$.
    Then we have
    \begin{equation}
        \mathbb{E}_{ \calB , \bfsigma} \brackets{ \sup_{ \Theta \in \mathcal{H}_B }
        \frac{1}{N} \sum_{j=1}^{N} \sum_{k=1}^{K} \sigma_{j, k} [\Theta^{\top} \phi(x_j)]_k
        }
        \leq \frac{B \sqrt{K}}{\sqrt{N}} \sqrt{  \tr{ \EE{\calD_x}{ \phi(x) \phi(x)^{\top} } }   } .
    \end{equation}
\end{lemma}

\begin{myproof}

    Recalling the algebraic fact that $\inner{A, u v^{\top}}_F = \tr{A^{\top} u v^{\top} }$ for any matrix $A$ and vectors $u,v$, we have
    $$
    \sum_{j=1}^N \sum_{k=1}^{K} \sigma_{j, k} [\Theta^{\top} \phi(x_j)]_k
    = \sum_{j=1}^{N} \sum_{k=1}^{K} \sigma_{j,k} e_k^{\top} \Theta^{\top} \phi(x_j)
    = \sum_{j=1}^{N} \phi(x_j)^{\top} \Theta \bfsigma_j
    = \sum_{j=1}^{N} \inner{\Theta, \phi(x_j) \bfsigma_j^{\top}}_F .
    $$
    Hence, we can rewrite the empirical Rademacher complexity of $\mathcal{F}$ as
    \begin{eqnarray}
        \widehat{\Re}_{N} \paren{ \mathcal{F} \mid \calB }  &=& \mathbb{E}_{\sigma } \brackets{
            \sup_{\Theta \in \mathcal{H}_B} \frac{1}{N} \sum_{j=1}^{N}
            \inner{\Theta, \phi(x_j) \bfsigma_j^{\top}}_F } \\
        &=& \frac{1}{N} \mathbb{E}_{\sigma } \brackets{
            \sup_{\Theta \in \mathcal{H}_B}
            \inner{\Theta, \sum_{j=1}^{N} \phi(x_j) \bfsigma_j^{\top}}_F } \nonumber \\
        &=& \frac{B}{N} \mathbb{E}_{\sigma } \brackets{
            \norm{ \sum_{j=1}^{N} \phi(x_j) \bfsigma_j^{\top} }_{F} } \label{eq:empirical_Rad_complexity_F_equal_to} .
    \end{eqnarray}
    Indeed, the last equality is due to the following fact.
    Let $(\mathbb{H},\langle\cdot, \cdot\rangle)$ be a real inner-product space with induced norm $\|h\|=\sqrt{\langle h, h\rangle}$. Let $a \in \mathbb{H}$ and $B \geq 0$. Then we have
    $
    \sup _{\|h\| \leq B}\langle h, a\rangle=B\|a\| .
    $
    Next, by Jensen's inequality, we have
    \begin{eqnarray}
         \mathbb{E}_{\sigma } \brackets{
            \norm{ \sum_{j=1}^{N} \phi(x_j) \bfsigma_j^{\top} }_{F} }
        &\leq& \paren{ \mathbb{E}_{\sigma } \brackets{ \norm{ \sum_{j=1}^{N} \phi(x_j) \bfsigma_j^{\top} }_{F}^{2}   }  }^{\frac{1}{2}}  \nonumber \\
        &=& \paren{ \sum_{j=1}^{N} \norm{  \phi(x_j) }_{2}^{2} \mathbb{E}_{\sigma } \brackets{ \norm{ \bfsigma_j }^{2}_{2} }   }^{\frac{1}{2}} \label{eq:1_step_after_jensen} \\
        &=&  \paren{ \sum_{j=1}^{N} \norm{  \phi(x_j)  }_{2}^{2}  K }^{\frac{1}{2}} . \label{eq:2_step_after_jensen}
    \end{eqnarray}
    Specifically, Equation~\ref{eq:1_step_after_jensen} is due to the following calculation:
    $$
    \mathbb{E}_{\sigma } \brackets{ \norm{ \sum_{j=1}^{N} \phi(x_j) \bfsigma_j^{\top} }_{F}^{2}   }
    =
        \sum_{a=1}^{d} \sum_{k=1}^{K}  \sum_{i=1}^{N} \sum_{j=1}^{N} \mathbb{E}_{\sigma } \brackets{ \phi(x_j)_a \sigma_{j,k}   \phi(x_i)_a \sigma_{i,k}
    }
    = \sum_{j=1}^{N} \sum_{k=1}^{K} \sum_{a=1}^{d} \phi(x_j)_a^2
    = K \sum_{j=1}^{N} \norm{  \phi(x_j) }_{2}^{2} .
    $$
    Equation~\ref{eq:2_step_after_jensen} is due to the independence of Rademacher random variables $\sigma_{j,k}$ across different $j$ and $k$, and the fact that $ \mathbb{E}_{\sigma } \brackets{ \norm{ \bfsigma_j }^{2}_{2}  }= \sum_{k=1}^{K} \E{ \sigma_{j,k}^2 }  = K$.

    Therefore, combining \eqref{eq:empirical_Rad_complexity_F_equal_to} and \eqref{eq:2_step_after_jensen}, we have
    \begin{eqnarray}
        \widehat{\Re}_{N} \paren{ \mathcal{F} \mid \calB }
        &\leq&  \frac{B}{N} \paren{ \sum_{j=1}^{N} \norm{  \phi(x_j)  }_{2}^{2}  K }^{\frac{1}{2}} .  \label{eq:last_step_before_expectation}
    \end{eqnarray}
    Invoking Jensen's inequality, we have
    \begin{equation}
        \EE{ \calB }{ \paren{ \sum_{j=1}^{N} \norm{  \phi(x_j)  }_{2}^{2}   }^{\frac{1}{2}} }
        \leq \sqrt{ \EE{\calB}{ \sum_{j=1}^{N} \norm{  \phi(x_j)  }_{2}^{2} } }
        = \sqrt{ N \tr{ \EE{\calD_x}{ \phi(x) \phi(x)^{\top} } } } .
    \end{equation}
    Taking expectation over $\calB$ on both sides of \eqref{eq:last_step_before_expectation} concludes the proof.

\end{myproof}

\subsection{Proof of Lemma~\ref{lemma:generalization_error_ell}}
\label{appendix:proof_lemma_generalization_error_ell}

\begin{myproof}
    Conditioned on the labeled data $\tilde{\calB}$, the teacher logit map $x \mapsto \eta_{\widehat{\Theta}_{\mathrm{w}}}^{\mathrm{w}}(x)$ is fixed. For notational simplicity, in this proof, we denote $\phi(x)=\phi_{\mathrm{s}}(x)$, $g(x)=\eta_{\widehat{\Theta}_{\mathrm{w}}}^{\mathrm{w}}(x)$, $L(\Theta)=L_{\mathrm{ls}}(\Theta)$, $\widehat{L}_N(\Theta)=\widehat{L}_{\mathrm{ls}}(\Theta)$,
    \[
        \ell(x,\Theta)=\norm{\Theta^{\top}\phi(x)-g(x)}_2^2,\qquad
        \ell_0(x)=\ell(x,0)=\norm{g(x)}_2^2,
    \]
    and $\bar{\ell}(x,\Theta)=\ell(x,\Theta)-\ell_0(x)$. Thus
    \[
        \bar{\ell}(x,\Theta)
        =
        \norm{\Theta^{\top}\phi(x)}_2^2
        -2g(x)^{\top}\Theta^{\top}\phi(x).
    \]
    All expectations in this proof are conditioned on $\tilde{\calB}$.

    The decomposition below follows directly from the identity
    $\ell(x,\Theta)=\ell_0(x)+\bar\ell(x,\Theta)$. Indeed, using
    $L(\Theta)=\EE{x}{\ell(x,\Theta)}$ and
    $\widehat L_N(\Theta)=N^{-1}\sum_{j=1}^N\ell(x_j,\Theta)$, subtracting the
    empirical average from the population average yields, for every $\Theta\in\mathcal H_B$,
    \begin{eqnarray}
        L(\Theta)-\widehat L_N(\Theta)
        &=&
        \EE{x}{\ell_0(x)+\bar\ell(x,\Theta)}
        -\frac1N\sum_{j=1}^N\paren{\ell_0(x_j)+\bar\ell(x_j,\Theta)}
        \nonumber\\
        &=&
        \paren{\EE{x}{\ell_0(x)}-\frac1N\sum_{j=1}^N\ell_0(x_j)}
        +
        \paren{\EE{x}{\bar\ell(x,\Theta)}-\frac1N\sum_{j=1}^N\bar\ell(x_j,\Theta)} .
        \label{eq:generalization_error_ell_uniform_decomposition}
    \end{eqnarray}
    Therefore
    \begin{eqnarray}
        && \mathbb{E}_{\calB}\brackets{\sup_{\Theta\in\mathcal H_B}
        \abs{L(\Theta)-\widehat L_N(\Theta)}} \nonumber\\
        &\leq&
        \mathbb{E}_{\calB}\brackets{\abs{\EE{x}{\ell_0(x)}-\frac1N\sum_{j=1}^N\ell_0(x_j)}}
        +
        \mathbb{E}_{\calB}\brackets{\sup_{\Theta\in\mathcal H_B}
        \abs{\EE{x}{\bar\ell(x,\Theta)}-\frac1N\sum_{j=1}^N\bar\ell(x_j,\Theta)}} .
        \label{eq:generalization_error_ell_full_to_centered}
    \end{eqnarray}
    The baseline term is bounded by Cauchy-Schwarz as
    \begin{equation}
        \mathbb{E}_{\calB}\brackets{\abs{\EE{x}{\ell_0(x)}-\frac1N\sum_{j=1}^N\ell_0(x_j)}}
        \leq
        \sqrt{\frac{\operatorname{Var}_{x}(\ell_0(x))}{N}}
        \leq
        \frac{1}{\sqrt N}\paren{\EE{x}{\norm{g(x)}_2^4}}^{1/2}.
        \label{eq:generalization_error_ell_baseline_fourth}
    \end{equation}
    Applying the standard symmetrization inequality to the function class
    $\braces{x\mapsto \bar{\ell}(x,\Theta):\Theta\in\mathcal H_B}$ gives
    \begin{equation}
        \mathbb{E}_{\calB}\brackets{\sup_{\Theta\in\mathcal H_B}
        \abs{\EE{x}{\bar\ell(x,\Theta)}-\frac1N\sum_{j=1}^N\bar\ell(x_j,\Theta)}}
        \leq
        2\mathbb{E}_{\calB,\sigma}\brackets{\sup_{\Theta\in\mathcal H_B}
        \abs{\frac1N\sum_{j=1}^N\sigma_j\bar\ell(x_j,\Theta)}} .
        \label{eq:generalization_error_ell_centered_symm}
    \end{equation}
    Since $\bar\ell(x,0)\equiv0$, the function class $\braces{x\mapsto \bar{\ell}(x,\Theta):\Theta\in\mathcal H_B}$ contains the zero function. Hence, using the symmetry of the Rademacher variables,
    \begin{equation}
        \mathbb{E}_{\calB,\sigma}\brackets{\sup_{\Theta\in\mathcal H_B}
        \abs{\frac1N\sum_{j=1}^N\sigma_j\bar\ell(x_j,\Theta)}}
        \leq
        2\mathbb{E}_{\calB,\sigma}\brackets{\sup_{\Theta\in\mathcal H_B}
        \frac1N\sum_{j=1}^N\sigma_j\bar\ell(x_j,\Theta)} .
        \label{eq:generalization_error_ell_abs_to_one_sided}
    \end{equation}
    Moreover, expanding $\bar\ell$ gives
    \begin{eqnarray}
        && \mathbb{E}_{\calB,\sigma}\brackets{\sup_{\Theta\in\mathcal H_B}
        \frac1N\sum_{j=1}^N\sigma_j\bar\ell(x_j,\Theta)} \nonumber \\
        &\leq&
        \mathbb{E}_{\sigma, \calB } \brackets{
        \sup_{\Theta \in \mathcal{H}_B} { \frac{1}{N} \sum_{j=1}^N \sigma_j   \normtwo{ \Theta^{\top} \phi(x_j) }^2
         } }
         + 2 \mathbb{E}_{\sigma, \calB } \brackets{
        \sup_{\Theta \in \mathcal{H}_B} { \abs{ \frac{1}{N} \sum_{j=1}^N \sigma_j   g(x_j)^{\top} \Theta^{\top} \phi(x_j)  }
         } } .
         \label{eq:two_terms_rademacher_complexity_ell}
     \end{eqnarray}

    For the first term in \eqref{eq:two_terms_rademacher_complexity_ell}, keeping the normalization from \eqref{eq:two_terms_rademacher_complexity_ell}, we have
    \begin{eqnarray}
        && \mathbb{E}_{\sigma, \calB } \brackets{
        \sup_{\Theta \in \mathcal{H}_B} {   \frac{1}{N}\sum_{j=1}^N \sigma_j   \normtwo{ \Theta^{\top} \phi(x_j) }^2
         } } \nonumber \\
        &=& \frac{1}{N}\mathbb{E}_{\sigma, \calB } \brackets{
        \sup_{\Theta \in \mathcal{H}_B} {   \tr{ \Theta^{\top} \paren{ \sum_{j=1}^{N} \sigma_j \phi(x_j) \phi(x_j)^{\top} }  \Theta }
         } } \nonumber  \\
        &\leq& \frac{1}{N}\mathbb{E}_{\sigma, \calB } \brackets{
        \sup_{\Theta \in \mathcal{H}_B} {   \norm{ \Theta }_F^2 \norm{ \sum_{j=1}^{N} \sigma_j \phi(x_j) \phi(x_j)^{\top} }_F
         } }  \label{eq:generalization_error_ell_trace_upperbound} \\
        &=& \frac{B^2}{N} \mathbb{E}_{\sigma, \calB } \brackets{
           \norm{ \sum_{j=1}^{N} \sigma_j \phi(x_j) \phi(x_j)^{\top} }_F
          } \nonumber  \\
        &\leq& \frac{B^2}{N} \sqrt{ \EE{\sigma, \calB}{ \norm{ \sum_{j=1}^{N} \sigma_j \phi(x_j) \phi(x_j)^{\top} }_F^2 } } \label{eq:generalization_error_ell_cauchy_3} \\
        &=&  \frac{B^2}{N}\sqrt{ \EE{\calB}{ \sum_{j=1}^{N} \norm{ \phi(x_j) \phi(x_j)^{\top} }_F^2 } } \label{eq:clean_up_Rad_2}  \\
        &=& \frac{B^2}{\sqrt{N}}\paren{\EE{x \sim \mathcal{D}_x }{ \norm{ \phi(x) }_2^4 }}^{\frac{1}{2}} \label{eq:two_terms_rademacher_complexity_ell_1} .
    \end{eqnarray}
    To arrive at \eqref{eq:generalization_error_ell_trace_upperbound}, we recall the fact that for any symmetric matrix $A$ compatible with matrix $\Theta$, $ \tr{\Theta^{\top} A \Theta } \leq \normtwo{A} \norm{ \Theta }_F^2 \leq \norm{ A }_F \norm{ \Theta }_F^2 $.
    Inequality~\eqref{eq:generalization_error_ell_cauchy_3} holds due to Cauchy-Schwarz inequality.
    Equality~\eqref{eq:clean_up_Rad_2} follows from the zero-mean property and the independence of Rademacher variables $\sigma_j$ across different $j$.

    For the second term in \eqref{eq:two_terms_rademacher_complexity_ell}, the Frobenius identity
    \(g(x_j)^\top\Theta^\top\phi(x_j)=\inner{\Theta,\phi(x_j)g(x_j)^\top}_F\)
    gives the normalized bound
    \begin{eqnarray}
        \mathbb{E}_{\sigma, \calB } \brackets{
        \sup_{\Theta \in \mathcal{H}_B} { \abs{  \frac{1}{N}\sum_{j=1}^N \sigma_j   g(x_j)^{\top} \Theta^{\top} \phi(x_j)  }
         } }
        &=&  \frac{1}{N}\mathbb{E}_{\sigma, \calB } \brackets{
        \sup_{\Theta \in \mathcal{H}_B} \abs{    \inner{ \Theta, \sum_{j=1}^{N} \sigma_j \phi(x_j)g(x_j)^{\top} }_F
         } } \\
        &\leq& \frac{B}{N}\mathbb{E}_{\sigma, \calB } \brackets{
             \norm{ \sum_{j=1}^{N} \sigma_j \phi(x_j)g(x_j)^{\top} }_F
        } \nonumber \\
        &\leq&  \frac{B}{N}\EE{ \calB }{ \sqrt{ \EE{\sigma}{ \norm{ \sum_{j=1}^{N} \sigma_j \phi(x_j)g(x_j)^{\top} }_F^2 \mid \calB} } } \nonumber \\
        &=& \frac{B}{N}\EE{ \calB }{ \sqrt{ \sum_{j=1}^{N} \norm{ \phi(x_j)g(x_j)^{\top} }_F^2  } } \label{eq:clean_up_Rad} \\
        &=& \frac{B}{N}\EE{ \calB }{ \sqrt{ \sum_{j=1}^{N} \norm{ g(x_j) }_2^2 \norm{ \phi(x_j) }_2^2  } } \label{eq:F_norm_algebra}\\
        &\leq& \frac{B}{\sqrt{N}}\paren{\EE{x}{ \norm{ g(x) }_2^2 \norm{ \phi(x) }_2^2 }}^{\frac{1}{2}}  \label{eq:generalization_error_ell_cauchy_1} \\
        &\leq& \frac{B}{\sqrt{N}}  \paren{ \E{ \norm{g(x)}_2^4 } }^{\frac{1}{4}} \paren{ \E{ \norm{\phi(x)}_2^4 }  }^{\frac{1}{4}}  \label{eq:generalization_error_ell_cauchy_2} .
    \end{eqnarray}
    In \eqref{eq:clean_up_Rad}, the cross terms vanish by the independence and zero-mean property of the Rademacher variables. Equality~\eqref{eq:F_norm_algebra} uses $\norm{uv^{\top}}_F=\norm{u}_2\norm{v}_2$, and \eqref{eq:generalization_error_ell_cauchy_1} follows from Cauchy-Schwarz inequality.

    The first term in \eqref{eq:generalization_error_ell_full_to_centered} is bounded by \eqref{eq:generalization_error_ell_baseline_fourth}. For the second term in \eqref{eq:generalization_error_ell_full_to_centered}, the symmetrization step and the one-sided reduction reduce the problem to the two Rademacher terms in \eqref{eq:two_terms_rademacher_complexity_ell}: the quadratic term is bounded by \eqref{eq:two_terms_rademacher_complexity_ell_1}, and the bilinear term is bounded by \eqref{eq:generalization_error_ell_cauchy_1}. Therefore
    \begin{equation}
        \mathbb{E}_{ \calB } \brackets{ \sup_{\Theta\in\mathcal H_B}\abs{L(\Theta)-\widehat L_N(\Theta)} }
        \leq
        \frac{1}{\sqrt N}\paren{\EE{x}{\norm{g(x)}_2^4}}^{1/2}
        +\frac{4B^2}{\sqrt N}\paren{\EE{x}{\norm{\phi(x)}_2^4}}^{1/2}
        +\frac{8B}{\sqrt N}\paren{\EE{x}{\norm{g(x)}_2^2\norm{\phi(x)}_2^2}}^{1/2}.
        \label{eq:generalization_error_ell_generic_bound}
    \end{equation}

    Finally, under the boundedness condition \(\norm{\widehat{\Theta}_{\rmw}}_F \leq B_{\Theta}^{\rmw}\) a.s., for every \(x\),
    \[
        \left\|\eta_{\widehat{\Theta}_{\mathrm{w}}}^{\mathrm{w}}(x)\right\|_2
        =
        \left\|\widehat{\Theta}_{\mathrm{w}}^{\top}\phi_{\mathrm{w}}(x)\right\|_2
        \leq
        B_{\Theta}^{\rmw}\left\|\phi_{\mathrm{w}}(x)\right\|_2 .
    \]
    Hence, with \(g(x)=\eta_{\widehat{\Theta}_{\mathrm{w}}}^{\mathrm{w}}(x)\) and \(\phi(x)=\phi_{\mathrm{s}}(x)\), the two terms involving \(g\) in \eqref{eq:generalization_error_ell_generic_bound} satisfy
    \[
        \paren{\mathbb{E}_{x}\norm{g(x)}_2^4}^{1/2}
        \leq
        \paren{B_{\Theta}^{\rmw}}^2\paren{\mathbb{E}_{x}\norm{\phi_{\mathrm{w}}(x)}_2^4}^{1/2},
        \qquad
        \paren{\mathbb{E}_{x}\norm{g(x)}_2^2\norm{\phi_{\mathrm{s}}(x)}_2^2}^{1/2}
        \leq
        B_{\Theta}^{\rmw}\paren{\mathbb{E}_{x}\norm{\phi_{\mathrm{w}}(x)}_2^2\norm{\phi_{\mathrm{s}}(x)}_2^2}^{1/2}.
    \]
    Substituting these two inequalities into \eqref{eq:generalization_error_ell_generic_bound} proves Lemma~\ref{lemma:generalization_error_ell}.

\end{myproof}

\subsection{Statement and proof of Lemma~\ref{lemma:joint_subgaussian_cov_fourth_moment}}

\begin{lemma} \label{lemma:joint_subgaussian_cov_fourth_moment}
    Let $X \in \mathbb{R}^d$ be mean-zero with covariance $\Sigma=\operatorname{Cov}(X) \succeq 0$. Assume there exists $c>0$ such that for all $u \in \mathbb{R}^d$,
    \begin{equation} \label{eq:joint_subgaussian_cov_mgf_bound_condition}
        \E{ \exp \paren{  u^{\top} X  } } \leq \exp \paren{ \frac{1}{2} c^2 u^{\top} \Sigma u }   .
    \end{equation}
    Then
    $$
    \mathbb{E}\|X\|_2^4 \leq 16 c^4( \tr{ \Sigma } )^2 .
    $$
\end{lemma}

\begin{myproof}
    We start from the following algebraic identity
    \begin{equation}
        \normtwo{X}^4 = \paren{ \sum_{i=1}^{d} X_i^2 }^2
        = \sum_{i=1}^{d} \sum_{j=1}^{d} X_i^2 X_j^2 .  \nonumber
    \end{equation}
    Taking expectation on both sides and applying Cauchy-Schwarz inequality, we have
    \begin{equation} \label{eq:fourth_moment_X}
        \E{ \normtwo{X}^4 }
        = \sum_{i=1}^{d} \sum_{j=1}^{d} \E{ X_i^2 X_j^2 }
        \leq \sum_{i=1}^{d} \sum_{j=1}^{d} \sqrt{\E{ X_i^4 } \E{ X_j^4 }}
        = \left( \sum_{i=1}^{d} \sqrt{\E{ X_i^4 }} \right)^2 .
    \end{equation}
    It suffices to upper bound $\E{ X_i^4 }$ for each $i \in [d]$.
    To this end, we recall the following well-known tail moment identity for any non-negative random variable $Z$: for any $p > 0$, we have
    \begin{equation}
        \E{ Z^p } = p \int_{0}^{\infty} s^{p-1} \pr{ Z > s } ds . \nonumber
    \end{equation}
    Applying the above identity to $Z = |X_i|$ and $p=4$, we have
    \begin{equation} \label{eq:fourth_moment_X_i_tail_identity}
        \E{ X_i^4 } = 4 \int_{0}^{\infty} s^{3} \pr{ |X_i| > s } ds .
    \end{equation}
    To proceed, we use the Chernoff method.
    We take $u = t e_i$ in \eqref{eq:joint_subgaussian_cov_mgf_bound_condition}, where $e_i$ is the $i$-th standard basis vector in $\mathbb{R}^d$ and $t >0$ is a free parameter to be determined later.
    Then
    \begin{equation}
        \E{ e^{t X_i} } \leq \exp \paren{ \frac{1}{2} c^2 t^2 \Sigma_{ii} } . \nonumber
    \end{equation}
    For any $t > 0$ and $s \geq 0$, we have
    \begin{equation}
        \pr{ X_i \geq s } = \pr{ e^{t X_i} \geq e^{t s} }
        \leq e^{-ts} \E{ e^{t X_i} }
        \leq \exp \paren{ \frac{1}{2} c^2 t^2 \Sigma_{ii} - t s } , \nonumber
    \end{equation}
    where the first inequality is by Markov's inequality.
    Taking $t = \frac{s}{c^2 \Sigma_{ii}}$, we have
    $
        \pr{ X_i \geq s } \leq \exp \paren{ - \frac{s^2}{2 c^2 \Sigma_{ii}} } . \nonumber
    $
    Hence, we conclude that
    \begin{equation} \label{eq:abs_X_i_tail}
        \pr{ \abs{ X_i } \geq s } \leq 2 \exp \paren{ - \frac{s^2}{2 c^2 \Sigma_{ii}} } .
    \end{equation}
    Therefore, combining \eqref{eq:fourth_moment_X}, \eqref{eq:fourth_moment_X_i_tail_identity} and \eqref{eq:abs_X_i_tail} yields that
    \begin{eqnarray}
        \E{ \normtwo{X}^4 }
        &\leq& \left( \sum_{i=1}^{d} \sqrt{\E{ X_i^4 }} \right)^2  \nonumber \\
        &=& 4 \left( \sum_{i=1}^{d} \sqrt{ \int_{0}^{\infty} s^{3} \pr{ |X_i| > s } ds } \right)^2  \nonumber \\
        &\leq& 4 \left( \sum_{i=1}^{d} \sqrt{ \int_{0}^{\infty} s^{3} 2 \exp \paren{ - \frac{s^2}{2 c^2 \Sigma_{ii}} } ds } \right)^2  \nonumber \\
        &=& 4 \left( \sum_{i=1}^{d}  2 c^2 \Sigma_{ii}    \right)^2 \label{eq:using_gamma_integral} \\
        &=& 16 c^4 ( \tr{ \Sigma } )^2 ,
    \end{eqnarray}
    where \eqref{eq:using_gamma_integral} follows from the fact that $\int_{0}^{\infty} s^{3} \exp \paren{ - \frac{s^2}{a} } ds = \frac{1}{2} a^2$ for $a>0$.

\end{myproof}

\subsection{Statement and proof of Lemma~\ref{lemma:lower_bound_tr_sandwich}}
\label{appendix:proof_lemma_lower_bound_tr_sandwich}
\begin{lemma} \label{lemma:lower_bound_tr_sandwich}
    Let $A \succeq 0$ be symmetric, $\Pi$ an orthogonal projection matrix, and $C \succeq 0$ symmetric such that $C = \Pi C \Pi$. Assume that $A - \alpha \Pi \succeq 0$ for some $\alpha > 0$. Then we have
    \begin{equation}
        \tr{ A C A } \geq \alpha^2 \tr{C} .
    \end{equation}
\end{lemma}

\begin{myproof}
    We denote $H = A - \alpha \Pi \succeq 0$. We can write
    \begin{eqnarray}
        ACA &=& \paren{ \alpha \Pi + H } C \paren{ \alpha \Pi + H }   \nonumber \\
        &=& \alpha^2 \Pi C \Pi + \alpha \paren{ \Pi C H + H C \Pi } + H C H  \nonumber  .
    \end{eqnarray}
    Since $C = \Pi C \Pi$ and $\Pi$ is an orthogonal projection, we have $\Pi C = C$ and $C \Pi = C$. Therefore,
    \begin{equation}
        ACA = \alpha^2 C + \alpha \paren{ C H + H C } + H C H .
    \end{equation}
    Taking trace on both sides, we have
    \begin{equation}
        \tr{ACA} = \alpha^2 \tr{C} + 2 \alpha \tr{CH} + \tr{HCH } \nonumber .
    \end{equation}
    Since $H \succeq 0$ and $C \succeq 0$, we have $\tr{HC} = \tr{C^{\frac{1}{2}} H C^{\frac{1}{2} }} \geq 0$ and $\tr{HCH} = \tr{C^{\frac{1}{2}} H^2 C^{\frac{1}{2}}} \geq 0$. Hence, we have $\tr{ACA} \geq \alpha^2 \tr{C}$, which concludes the proof.

\end{myproof}

\subsection{A no-free-lunch bound for zero overlap}
\label{appendix:proof_lemma_case_study_overlap_lower_bound}
The following lemma shows that, under exact realizability, the overlap dimension cannot be arbitrarily small.
\begin{lemma} \label{lemma:case_study_overlap_lower_bound}
    When $\rho_{\rms} = \rho_{\rmw} = 0$, we have
    \begin{equation}
        d_{\rms \wedge \rmw} \geq \frac{1}{\lambda_{\max} \paren{\Sigma_{\rmw}}} \frac{ \norm{\Sigma_{\rms}^{\frac{1}{2} } \Theta_{\rms}^{*}}_F^2 } { \norm{ \Theta_{\rmw}^{*} }_F^2 } .
    \end{equation}
\end{lemma}
\begin{myproof}
    Since $\rho_{\rms} = \rho_{\rmw} = 0$, we have $ \eta^{*}(x) = \paren{\Theta_{\rms}^*}^{\top} \phi_{\rms}(x) = \paren{\Theta_{\rmw}^*}^{\top} \phi_{\rmw}(x)  $ almost surely with respect to $x$.
    We denote $\E{ \phi_{\rms} \phi_{\rmw}^{\top} } = \Sigma_{\rms \rmw}$.
    Hence, we can compute $ \E{ \phi_{\rms} (\eta^{*})^{\top} } $ in two ways:
    $$
    \E{  \phi_{\rms} (\eta^{*})^{\top} } = \E{ \phi_{\rms} \phi_{\rms}^{\top} } \Theta_{\rms}^* = \Sigma_{\rms} \Theta_{\rms}^*
    \quad \text{and} \quad
    \E{  \phi_{\rms} (\eta^{*})^{\top} } = \E{ \phi_{\rms} \phi_{\rmw}^{\top} } \Theta_{\rmw}^* = \Sigma_{\rms \rmw} \Theta_{\rmw}^* .
    $$
    Hence, we have $ \Sigma_{\rms} \Theta_{\rms}^* = \Sigma_{\rms \rmw} \Theta_{\rmw}^*  $.
    Left multiplying both sides by $\Sigma_{\rms}^{-\frac{1}{2}}$, we have $\Sigma_{\rms}^{\frac{1}{2}} \Theta_{\rms}^* = \Sigma_{\rms}^{-\frac{1}{2}} \Sigma_{\rms \rmw} \Theta_{\rmw}^*$.
    Denote $\Pi_{\rmw} = V_{\rmw} V_{\rmw}^{\top}$.
    Since $\phi_{\rmw}(x)=\Pi_{\rmw}\phi_{\rmw}(x)$ almost surely, replacing $\Theta_{\rmw}^{*}$ by $\Pi_{\rmw}\Theta_{\rmw}^{*}$ leaves the weak logits unchanged, so we can assume without loss of generality that $\Theta_{\rmw}^{*} = \Pi_{\rmw} \Theta_{\rmw}^{*} $. Then we have
    $ \Sigma_{\rms}^{\frac{1}{2}} \Theta_{\rms}^* = \Sigma_{\rms}^{-\frac{1}{2}} \Sigma_{\rms \rmw} \Pi_{\rmw} \Theta_{\rmw}^{*} $. Denote $B = \Sigma_{\rms}^{-\frac{1}{2}} \Sigma_{\rms \rmw} \Pi_{\rmw}$. Then the fact that Frobenius norm admits submultiplicativity implies that
    \begin{equation} \label{eq:Sigma_s_Theta_s_upper_bound}
        \norm{\Sigma_{\rms}^{\frac{1}{2}} \Theta_{\rms}^*}_F \leq \norm{ B }_F \norm{\Theta_{\rmw}^{*}}_F  .
    \end{equation}

    On the other hand, we can compute the overlap dimension $d_{s \wedge w}$ as follows:
    \begin{equation} \label{eq:overlap_calculation_exact}
        d_{s \wedge w} = \norm{ \Sigma_{\rms}^{-\frac{1}{2}} \Sigma_{\rms \rmw} \Sigma_{\rmw}^{-\frac{1}{2}} }_F^2
        = \norm{ \Sigma_{\rms}^{-\frac{1}{2}} \Sigma_{\rms \rmw} \Pi_{\rmw} \Sigma_{\rmw}^{-\frac{1}{2}} }_F^2
        = \norm{ B \Sigma_{\rmw}^{-\frac{1}{2}}  }_F^2
        = \tr{ \Sigma_{\rmw}^{-\frac{1}{2}} B^{\top} B \Sigma_{\rmw}^{-\frac{1}{2}} } ,
    \end{equation}
    where the second equality follows from the fact that
    \begin{eqnarray}
        \Pi_{\rmw} \Sigma_{\rmw}^{-\frac{1}{2}}
        = V_{\rmw} V_{\rmw}^{\top}  V_{\rmw} \Lambda_{\rmw}^{-\frac{1}{2}} V_{\rmw}^{\top}
        = V_{\rmw} \Lambda_{\rmw}^{-\frac{1}{2}} V_{\rmw}^{\top}
        = \Sigma_{\rmw}^{-\frac{1}{2}} , \nonumber
    \end{eqnarray}
    by simply noting that $V_{\rmw}^{\top} V_{\rmw} = I_{d_{\rmw}}$.
    Next, we can verify that $\Sigma_{\rmw}^{-\frac{1}{2}} \succeq \frac{1}{\sqrt{\lambda_{\max} \paren{\Sigma_{\rmw}}}} \Pi_{\rmw}$.
    Indeed, by writing $\Sigma_{\rmw}^{-\frac{1}{2}} - \frac{1}{ \sqrt{ \lambda_{\max} \paren{\Sigma_{\rmw}} } } \Pi_{\rmw}
    = V_{\rmw} \Lambda_{\rmw}^{-\frac{1}{2}} V_{\rmw}^{\top} - \frac{1}{ \sqrt{ \lambda_{\max} \paren{\Sigma_{\rmw}} } } V_{\rmw} V_{\rmw}^{\top}
    = V_{\rmw} \paren{ \Lambda_{\rmw}^{-\frac{1}{2}} - \frac{1}{ \sqrt{ \lambda_{\max} \paren{\Sigma_{\rmw}} } } I_{d_{\rmw}} } V_{\rmw}^{\top}$, we can verify that $\Lambda_{\rmw}^{-\frac{1}{2}} - \frac{1}{\sqrt{ \lambda_{\max} \paren{\Sigma_{\rmw}} }} I_{d_{\rmw}}$ is a diagonal matrix with non-negative entries, and hence is positive semidefinite.

    Therefore, invoking Lemma~\ref{lemma:lower_bound_tr_sandwich} on \eqref{eq:overlap_calculation_exact}, we have
    \begin{eqnarray}
        d_{s \wedge w}  \geq  \frac{1}{\lambda_{\max} \paren{\Sigma_{\rmw}}} \norm{ B }_F^2
        \geq  \frac{1}{\lambda_{\max} \paren{\Sigma_{\rmw}}} \frac{ \norm{\Sigma_{\rms}^{\frac{1}{2} } \Theta_{\rms}^{*}}_F^2 } { \norm{ \Theta_{\rmw}^{*} }_F^2 } .
    \end{eqnarray}
    The last inequality follows from \eqref{eq:Sigma_s_Theta_s_upper_bound}.
\end{myproof}

\section{Omitted Proofs in Section~\ref{sec:comparison_bounds}}

\subsection{Proof of Theorem~\ref{thm:lower_bound_w2s}}
\label{appendix:proof_thm_lower_bound_strong}
\begin{myproof}

\textbf{Step 1: lower bound the decision risk by $\Omega \paren{ \rho_{\rms} + \lambda_{\min} \paren{\Lambda_{\rms} } \normtwo{\Delta \theta }^2}$. }
By Lemma~\ref{lemma:convex_smooth_expected_cost}, we have for any $x$,
\begin{equation} \label{eq:strong_model_lower_bound_step_1_using_Lemma_1}
    \mathbb{E}_{ \xi \sim P^*(\cdot \mid x) } \brackets{
        c \paren{ \tilde{w} \paren{  P^{\rms}_{ \widehat{\Theta}_s } \mid x } , \xi }
      - c \paren{ \tilde{w} \paren{ P^{*} \mid x } , \xi }
    }
    \geq \frac{\mu}{2} \norm{ \tilde{w} \paren{  P^{\rms}_{ \widehat{\Theta}_s } \mid x } - \tilde{w} \paren{ P^{*} \mid x } }_2^2 .
\end{equation}
We claim that under Assumption (1), we have
\begin{equation} \label{eq:strong_model_lower_bound_step_1}
    \norm{ \tilde{w} \paren{  P^{\rms}_{ \widehat{\Theta}_s } \mid x } - \tilde{w} \paren{ P^{*} \mid x } }_2
    \geq \gamma \frac{e^{-2 B_{\eta}}}{K} \norm{ \eta^{\rms}_{ \widehat{\Theta}_s } (x) - \eta^{*}(x) }_2 .
\end{equation}
To see this, for brevity, for any logit vector $\eta \in \mathbb{R}^K$, we denote
    \begin{equation}
        s(\eta)_k \defeq \frac{e^{\eta_k}}{\sum_{j=1}^{K} e^{\eta_j}} \text{ for } k \in [K]  \nonumber
    \end{equation}
    to be the $k$th coordinate of the softmax function evaluated at $\eta$.
    If $\|\eta\|_{\infty} \leq B_{\eta}$, then for any $k$,
    $
    s(\eta)_k=\frac{e^{\eta_k}}{\sum_j e^{\eta_j}} \geq \frac{e^{-B_{\eta}}}{K e^{B_{\eta}}}=\frac{e^{-2 B_{\eta}}}{K} .
    $
    If $\left\|\eta_0\right\|_{\infty} \leq B_{\eta}$ and $\left\|\eta_1\right\|_{\infty} \leq B_{\eta}$, then the entire segment $\eta(t) = (1-t) \eta_0 + t \eta_1$ also satisfies $\|\eta(t)\|_{\infty} \leq B_{\eta}$ for $t \in [0,1]$, hence
    $
    \min_{t \in[0,1]} \min _{k \in[K]} s(\eta(t))_k \geq \frac{e^{-2 B_{\eta}}}{K} .
    $
    Then \eqref{eq:assumption_eta_two_ends_bounded} implies that
    \begin{equation}
        \min_{t \in[0,1]} \min _{k \in[K]}
        s( \eta^{*} (x) + t \paren{ \eta_{\widehat{\Theta}_{\rms}}(x) - \eta^{*}(x) } )_k \geq \frac{e^{-2 B_{\eta}}}{K}  .
    \end{equation}
    To proceed, we invoke the following lemma.
    \begin{lemma} \label{lemma:lower_bound_softmax_cushion_implies_continuity}
        Given $\eta_0, \eta_1 \in \mathbb{R}^K$ satisfying $\bfone^{\top} \eta_0 = 0$ and $\bfone^{\top} \eta_1 = 0$.
        Along the segment $\eta(t) = (1-t) \eta_0 + t \eta_1$ for $t \in [0,1]$, the induced probability distributions satisfy
        \begin{equation}
            \min_{k \in [K]} s \paren{ \eta(t) }_k \geq \alpha , \quad \forall t \in [0,1]
        \end{equation}
        for some constant $\alpha > 0$.
        Then we have
        \begin{equation}
            \norm{ s(\eta_1) - s(\eta_0) }_2 \geq \alpha \norm{ \eta_1 - \eta_0 }_2 .
        \end{equation}
    \end{lemma}

    \begin{myproof}
        Let $v = \eta_1 - \eta_0$, so $\bfone^{\top} v = 0$. Define $p(t) = s(\eta(t))$. By the fundamental theorem of calculus, we have
        \begin{equation}
            s(\eta_1) - s(\eta_0) = \int_0^1 \frac{d}{dt} s(\eta(t)) dt = \int_0^1 J(\eta(t)) v \, dt,
        \end{equation}
        where $J(\eta) = \operatorname{Diag}(p) - p p^{\top}$ is the Jacobian matrix of the softmax function at $\eta$.

        Recall $\Pi = I_K - \frac{1}{K} \bfone \bfone^{\top}$ is the projection onto the subspace orthogonal to $\bfone$.
        Next, we claim that for every $t \in [0,1]$,
        \begin{equation} \label{eq:J_lower_bounded_by_alpha_Pi}
            J(\eta(t)) \succeq \alpha \Pi .
        \end{equation}
        Suppose the claim holds for now. This immediately implies that $M \defeq \int_{0}^{1} J(\eta(t)) dt \succeq \alpha \Pi$.
        Since $v^{\top} \bfone = 0$, we have $\Pi v = v$. Hence, $v^{\top} \Pi v = v^{\top} v = \norm{v}_2^2$.
        Then $ M \succeq \alpha \Pi$ implies that $ v^{\top} M v \geq \alpha v^{\top} \Pi v = \alpha \norm{v}_2^2$.
        By Cauchy-Schwarz inequality, we have
        \begin{equation}
            \normtwo{ M  v  }
            \geq \frac{v^{\top} M v }{\normtwo{v}}
            \geq \alpha \norm{v}_2 .
        \end{equation}
        Noting that $s(\eta_1) - s(\eta_0) = M v$, the claim then follows.

        Now, we return to prove \eqref{eq:J_lower_bounded_by_alpha_Pi}.
        In fact, for any arbitrary $x \in \mathbb{R}^K$, we decompose it as $x = \Pi x + \frac{1}{K} (\bfone^{\top} x) \bfone$.
        Then, we notice that
        \begin{equation}
            x^{\top} J \paren{ \eta(t) } x
            = (\Pi x + \frac{1}{K} (\bfone^{\top} x) \bfone)^{\top} J \paren{ \eta(t) } (\Pi x + \frac{1}{K} (\bfone^{\top} x) \bfone)
            = (\Pi x )^{\top} J \paren{ \eta(t) } (\Pi x ) .
        \end{equation}
        The last step holds since $J(\eta(t)) \bfone = \operatorname{diag}(p) \bfone - p p^{\top} \bfone = p - p = 0$.
        Let $m \defeq \min_{k \in [K]} s \paren{ \eta(t) }_k \geq \alpha$.
        Using Lemma~\ref{lemma:softmax_J_dominates_projection_under_cushion} and the fact that $\bfone^{\top} \Pi x = 0$, we have
        \begin{equation}
            x^{\top} J \paren{ \eta(t) } x
            = (\Pi x )^{\top} J \paren{ \eta(t) } (\Pi x )
            \geq m \norm{ \Pi x }_2^2
            \geq \alpha x^{\top} \Pi x ,
        \end{equation}
        which is exactly \eqref{eq:J_lower_bounded_by_alpha_Pi}.

    \end{myproof}

In view of Lemma~\ref{lemma:lower_bound_softmax_cushion_implies_continuity} and the condition given in \eqref{eq:w_tilde_lowerbounded_p}, the proof of \eqref{eq:strong_model_lower_bound_step_1} is now complete.
Combining \eqref{eq:strong_model_lower_bound_step_1_using_Lemma_1} and \eqref{eq:strong_model_lower_bound_step_1} yields that
\begin{equation}
    \mathbb{E}_{ \xi \sim P^*(\cdot \mid x) } \brackets{
        c \paren{ \tilde{w} \paren{  P^{\rms}_{ \widehat{\Theta}_s } \mid x } , \xi }
      - c \paren{ \tilde{w} \paren{ P^{*} \mid x } , \xi }
    }
    \geq \frac{1}{2} \mu \paren{\gamma \frac{e^{-2 B_{\eta}}}{K}}^2 \norm{ \eta^{\rms}_{ \widehat{\Theta}_s } (x) - \eta^{*}(x) }_2^2 .
\end{equation}

Next, we proceed to show that
\begin{equation} \label{eq:lower_bound_approx_estimation_decomposition}
    \mathbb{E}_{ \tilde{\mathcal{B}} , x } \brackets{ \norm{ \eta^{\rms}_{ \widehat{\Theta}_s } (x) - \eta^{*}(x) }_2^2 }
    \geq \rho_{\rms} + \mathbb{E}_{ \tilde{\mathcal{B}} , x }
    \brackets{  \normtwo{ \paren{ \widehat{\Theta}_s - \Theta_{\rms}^{*} }^{\top} \phi_{\rms}(x)  }^2 } .
\end{equation}

Indeed, we note that
\begin{eqnarray}
    \norm{ \eta^{\rms}_{ \widehat{\Theta}_s } (x) - \eta^{*}(x) }_2^2
    &=& \norm{  \paren{ \widehat{\Theta}_s - \Theta_{\rms}^{*} }^{\top} \phi_{\rms}(x) }_2^2
        + \norm{ (\Theta_{\rms}^{*})^{\top} \phi_{\rms}(x) - \eta^{*}(x) }_2^2 \nonumber \\
    && + 2 \inner{  \paren{ \widehat{\Theta}_s - \Theta_{\rms}^{*} }^{\top} \phi_{\rms}(x) ,
    (\Theta_{\rms}^{*})^{\top} \phi_{\rms}(x) - \eta^{*}(x) } .
\end{eqnarray}
For the cross term, taking expectation over $x$ and $\tilde{\calB}$ , we further have
 \begin{eqnarray*}
    && \EE{ \tilde{\calB}}{ \EE{x \mid \tilde{\calB} }{
        \inner{  \paren{ \widehat{\Theta}_s - \Theta_{\rms}^{*} }^{\top} \phi_{\rms}(x) ,
    (\Theta_{\rms}^{*})^{\top} \phi_{\rms}(x) - \eta^{*}(x) }
       \mid \tilde{\calB}  } }   \\
    &=& \EE{ \tilde{\calB}}{ \EE{x \mid \tilde{\calB} }{
        \inner{    \widehat{\Theta}_s - \Theta_{\rms}^{*}    ,
    \phi_{\rms}(x)  \paren{ (\Theta_{\rms}^{*})^{\top} \phi_{\rms}(x) - \eta^{*}(x) }^{\top} }_F
       \mid \tilde{\calB}  } } \\
    &=& \EE{ \tilde{\calB}}{
        \inner{    \widehat{\Theta}_s - \Theta_{\rms}^{*}    ,
    \EE{x \mid \tilde{\calB} }{ \phi_{\rms}(x)  \paren{ (\Theta_{\rms}^{*})^{\top} \phi_{\rms}(x) - \eta^{*}(x) }^{\top} \mid \tilde{\calB}  }
        }_F } \\
    &=& \EE{ \tilde{\calB}}{
        \inner{    \widehat{\Theta}_s - \Theta_{\rms}^{*}    ,
    \EE{x}{ \phi_{\rms}(x)  \paren{ (\Theta_{\rms}^{*})^{\top} \phi_{\rms}(x) - \eta^{*}(x) }^{\top}  }
        }_F }  .
\end{eqnarray*}
Now we recall that $\Theta_{\rms}^{*}$ is defined as the minimizer of
\begin{equation}
    \Theta_{\rms}^{*} = \argmin_{ \Theta \in \mathbb{R}^{ d_{\phi} \times K}, \Theta \bfone = 0 }
    L(\Theta)  ,
\end{equation}
where $L(\Theta) = \EE{x}{ \norm{ \Theta^{\top} \phi_{\rms}(x) - \eta^{*}(x) }_{2}^{2} } $ is a convex function of $\Theta$.
We denote $\calC = \{ \Theta \in \mathbb{R}^{ d_{\phi} \times K} : \Theta \bfone = 0 \}$.
By the first-order optimality condition of convex optimization, we have
\begin{equation}
    \inner{ \nabla L( \Theta_{\rms}^{*} ) , \Theta - \Theta_{\rms}^{*} }_F \geq 0 , ~ \forall~ \Theta \in \calC .
\end{equation}
By noting that
\begin{equation}
    \nabla L (\Theta) =  2 \EE{x}{ \phi_{\rms}(x)  \paren{ \Theta^{\top} \phi_{\rms}(x) - \eta^{*}(x) }^{\top}  }   \nonumber ,
\end{equation}
we have thus proved \eqref{eq:lower_bound_approx_estimation_decomposition}.

Denote $\Delta \Theta = \widehat{\Theta}_{\rms}-\Theta_{\rms}^* \in \mathbb{R}^{d_{\phi} \times K}$ and $\Delta \theta = \operatorname{vec} \paren{ \widehat{\Theta}_{\rms}-\Theta_{\rms}^* } \in \mathbb{R}^{d_{\phi} K}$.

We note that by the cyclic property and linearity property of trace, we have
\begin{eqnarray}
    \mathbb{E}\left[\left\|\Delta \Theta^{\top} \phi_{\rms}(x)\right\|_2^2 \mid \tilde{\mathcal{B}}\right]
    &=& \mathbb{E}_{x} \left[\phi_{\rms}(x)^{\top} \Delta \Theta \Delta \Theta^{\top} \phi_{\rms}(x) \mid \tilde{\mathcal{B}}\right] \nonumber \\
    &=& \mathbb{E}_{x} \left[ \tr{ \phi_{\rms}(x)^{\top} \Delta \Theta \Delta \Theta^{\top} \phi_{\rms}(x) } \mid \tilde{\mathcal{B}}\right]  \nonumber \\
    &=& \mathbb{E}_{x} \left[ \tr{  \Delta \Theta \Delta \Theta^{\top} \phi_{\rms}(x) \phi_{\rms}(x)^{\top} } \mid \tilde{\mathcal{B}}\right]  \nonumber \\
    &=&  \tr{  \Delta \Theta \Delta \Theta^{\top} \mathbb{E}_{x} \left[ \phi_{\rms}(x) \phi_{\rms}(x)^{\top} \mid \tilde{\mathcal{B}}\right] }  \nonumber \\
    &=& \tr{  \Delta \Theta \Delta \Theta^{\top} \Sigma_{\rms} } .
\end{eqnarray}
The last equation holds since $x$ is sampled independent of $\tilde{\calB}$.
Recall that for any compatible matrices $A,B,X$, we have
$
\operatorname{tr}\left(A X B X^{\top}\right)
= \operatorname{vec}(X)^{\top}\left(B^{\top} \otimes A\right) \operatorname{vec}(X) .
$
Taking $X = \Delta \Theta$, $A=\Sigma_{\rms}$, $B=I_K$, we have
\begin{eqnarray}
    \mathbb{E}\left[\left\|\Delta \Theta^{\top} \phi_{\rms}(x)\right\|_2^2 \mid \tilde{\mathcal{B}}\right]
    &=& \Delta \theta^{\top} \left(  I_K \otimes \Sigma_{\rms} \right) \Delta \theta \nonumber \\
    &=&  \Delta \theta^{\top} \paren{ I_K \otimes V_{\rms} } \paren{ I_K \otimes \Lambda_{\rms} } \paren{ I_K \otimes V_{\rms}^{\top} } \Delta \theta \label{eq:use_ABCD_Kronecker_product}  \\
    &=& \normtwo{ \paren{ I_K \otimes \Lambda_{\rms}^{\frac{1}{2}} } \paren{ I_K \otimes V_{\rms}^{\top} } \Delta \theta  }^2  \label{eq:I_K_Sigma_s_half} \\
    &\geq& \lambda_{\min } ( \Lambda_{\rms} ) \norm{ \paren{ I_K \otimes V_{\rms}^{\top} } \Delta \theta }_2^2  \nonumber \\
    &=&  \lambda_{\min } ( \Lambda_{\rms} ) \Delta \theta^{\top}\left(I_K \otimes V_{\rms} V_{\rms}^{\top}\right) \Delta \theta   \label{eq:lower_bound_norm_Delta_Theta_phi}  .
\end{eqnarray}
Equation~\eqref{eq:use_ABCD_Kronecker_product} follows from the property of Kronecker product that $ (A \otimes B) ( C \otimes D) = (AC) \otimes (BD)$ valid for shape-compatible matrices $A,B,C,D$.
Equation~\eqref{eq:I_K_Sigma_s_half} is due to the simple fact that $ \paren{ I_K \otimes \Lambda_{\rms}^{1 / 2} }^{\top} = I_K \otimes \paren{\Lambda_{\rms}^{1 / 2}}^{\top}$.
For the inequality, we note that the eigenvalues of $  I_K \otimes \Lambda_{\rms}^{1 / 2}  $ are given by $ \lambda_j ( I_K ) \cdot \lambda_i ( \Lambda_{\rms}^{1 / 2} ) = \lambda_i ( \Lambda_{\rms}^{1 / 2} )$ for $i=1,2,\ldots,d_{\rms}$ and $j=1,2,\ldots,K$. Hence, the minimum eigenvalue of $ \Lambda_{\rms}^{1 / 2} \otimes I_K$ is equal to $\lambda_{\min } ( \Lambda_{\rms}^{1 / 2} ) = \sqrt{ \lambda_{\min } ( \Lambda_{\rms} ) }$.

It remains to relate the projected quadratic form in \eqref{eq:lower_bound_norm_Delta_Theta_phi} to $\normtwo{\Delta\theta}^2$. Since $\Sigma_{\rms}=\mathbb{E}[\phi_{\rms}(x)\phi_{\rms}(x)^{\top}]=V_{\rms}\Lambda_{\rms}V_{\rms}^{\top}$,
\begin{eqnarray*}
    \mathbb{E}\normtwo{\left(I-V_{\rms}V_{\rms}^{\top}\right)\phi_{\rms}(x)}^2
    &=& \tr{\left[\left(I-V_{\rms}V_{\rms}^{\top}\right)\mathbb{E}\left[\phi_{\rms}(x)\phi_{\rms}(x)^{\top}\right]\right]}  \\
    &=& \tr{\left[\left(I-V_{\rms}V_{\rms}^{\top}\right)V_{\rms}\Lambda_{\rms}V_{\rms}^{\top}\right]}=0,
\end{eqnarray*}
where $\left(I-V_{\rms}V_{\rms}^{\top}\right)V_{\rms}=0$. Since the left-hand side is nonnegative, $\phi_{\rms}(x)=V_{\rms}V_{\rms}^{\top}\phi_{\rms}(x)$ almost surely. Hence, for any $\Theta$, $\left(V_{\rms}V_{\rms}^{\top}\Theta\right)^{\top}\phi_{\rms}(x)=\Theta^{\top}V_{\rms}V_{\rms}^{\top}\phi_{\rms}(x)=\Theta^{\top}\phi_{\rms}(x)$ almost surely. Moreover, for any $\Theta\in\calC$, $(V_{\rms}V_{\rms}^{\top}\Theta)\bfone=V_{\rms}V_{\rms}^{\top}(\Theta\bfone)=0$.
Therefore, it is without loss of generality to assume that $\Theta_{\rms}^*=V_{\rms} V_{\rms}^{\top} \Theta_{\rms}^*$ and $ \widehat{\Theta}_{\rms}=V_{\rms} V_{\rms}^{\top} \widehat{\Theta}_{\rms}$.
Subtracting the two projected identities and using the algebraic identity $\operatorname{vec}(ABC)=(C^{\top}\otimes A)\operatorname{vec}(B)$, we have
\begin{eqnarray*}
    \Delta\Theta
    &=& \widehat{\Theta}_{\rms}-\Theta_{\rms}^{*}
     = V_{\rms}V_{\rms}^{\top}\Delta\Theta, \qquad
    \Delta\theta
     = \operatorname{vec}\left(V_{\rms}V_{\rms}^{\top}\Delta\Theta\right)
     = \left(I_K\otimes V_{\rms}V_{\rms}^{\top}\right)\Delta\theta .
\end{eqnarray*}
Hence,
$
    \Delta\theta^{\top}\left(I_K\otimes V_{\rms}V_{\rms}^{\top}\right)\Delta\theta
    =
    \Delta\theta^{\top}\Delta\theta
    =
    \normtwo{\Delta\theta}^{2}.
$
We hence have shown that $ \mathbb{E}\left[\left\|\Delta \Theta^{\top} \phi_{\rms}(x)\right\|_2^2 \mid \tilde{\mathcal{B}}\right]   \geq \lambda_{\min } ( \Lambda_{\rms} ) \normtwo{\Delta \theta}^2$.

\textbf{Step 2: lower bound $ \normtwo{\Delta \theta} $.}

Let
$$
F_n(\Theta) \defeq \frac{1}{n} \sum_{i=1}^{n} c\left(\tilde{w}\left(P_{\Theta}^{\mathrm{s}} (\tilde{x}_i)\right), \tilde{\xi}_i\right) .
$$
Let $\Pi = I_K - \frac{1}{K} \bfone \bfone^{\top} \in \mathbb{R}^{K \times K}$ be the projection matrix to the subspace $\{ v \in \mathbb{R}^K : v^{\top} \bfone = 0 \}$.
We denote $\Theta(t) \defeq \Theta_{\rms}^*+t \paren{\widehat{\Theta}_{\rms}-\Theta_{\rms}^*} \in \mathbb{R}^{d_{\phi} \times K} $ for $t\in[0,1]$.
Since $\widehat{\Theta}_{\rms} = \argmin_{\Theta \in \calC } ~ F_n (\Theta) $, the KKT conditions imply that there exists a multiplier $\nu \in \mathbb{R}^{d_{\phi}}$ such that
\begin{equation}
    \nabla_{\Theta} F_n \paren{ \widehat{\Theta}_{\rms} } + \nu \bfone^{\top} = 0 ,
    \quad \text{ and } \quad
    \widehat{\Theta}_{\rms} \bfone = 0 .
\end{equation}
Right-multiplying the above condition by $\Pi$ and using the fact that $ \bfone^{\top} \Pi = 0$, we have
\begin{equation} \label{eq:KKT_F_n_equiv}
    \nabla_{\Theta} F_n \paren{ \widehat{\Theta}_{\rms} } \Pi = 0 .
\end{equation}

Applying the fundamental theorem of calculus to $g \paren{ \Theta \paren{t} } \defeq \operatorname{vec}(\nabla_{\Theta} F_n(\Theta(t)) \Pi )$, we have
\begin{equation} \label{eq:fundamental_thm_calculas_g_theta_t}
    g \paren{ \widehat{\Theta}_{\rms} } - g \paren{ \Theta_{\rms}^* }
    = \int_0^1 \frac{d }{d t} g \paren{ \Theta \paren{t} } d t
    = \int_0^1 \paren{ \Pi \otimes I_{d_{\phi}}  } \nabla_{\Theta}^2 F_n\left( \Theta \paren{t} \right) \Delta \theta d t .
\end{equation}
We denote $ H_n \defeq \int_0^1 \paren{ \Pi \otimes I_{d_{\phi}}  }   \nabla_{\Theta}^2 F_n\left( \Theta \paren{t} \right)  d t  \in \mathbb{R}^{d_{\phi} K \times d_{\phi} K} $ and $ \varphi_n \defeq g \paren{ \Theta_{\rms}^* } = \operatorname{vec} \paren{ \nabla_{\Theta} F_n \paren{ \Theta_{\rms}^* } \Pi } $.
Equation~\eqref{eq:KKT_F_n_equiv} and \eqref{eq:fundamental_thm_calculas_g_theta_t} hence imply that
\begin{equation}
    0 - \varphi_n = H_n \Delta \theta .
\end{equation}

By Assumption~\eqref{eq:assumption_hessian_c}, we have
\begin{equation}
    \normtwo{H_n}
\leq \int_0^1 \normtwo{ \paren{ \Pi \otimes I_{d_{\phi}}  }   \nabla_{\Theta}^2 F_n\left( \Theta \paren{t} \right) } d t
\leq \int_0^1 \normtwo{ \paren{ \Pi \otimes I_{d_{\phi}}  } }
\normtwo{ \nabla_{\Theta}^2 F_n\left( \Theta \paren{t} \right) } d t
\leq L_H .
\end{equation}
The last inequality follows because $\normtwo{ \Pi \otimes I} = \normtwo{\Pi} = 1$.
Hence, we obtain
\begin{equation}
    \normtwo{ \varphi_n } = \normtwo{ H_n \Delta \theta }
\leq \normtwo{H_n} \normtwo{ \Delta \theta }
\leq L_H  \normtwo{ \Delta \theta }   .
\end{equation}
Continuing from \eqref{eq:lower_bound_norm_Delta_Theta_phi} and taking expectation with respect to $\tilde{\mathcal{B}}$, we have
\begin{equation}
    \mathbb{E}\left[\left\|\Delta \Theta^{\top} \phi_{\rms}(x)\right\|_2^2\right]
    = \EE{\tilde{\mathcal{B}}} { \mathbb{E}\left[\left\|\Delta \Theta^{\top} \phi_{\rms}(x)\right\|_2^2 \mid \tilde{\mathcal{B}}\right] }
    \geq  \frac{\lambda_{\min }\left(\Lambda_{\rms}\right)}{L_H^2} \EE{\tilde{\mathcal{B}}}{ \left\|\varphi_n\right\|_2^2 } .
\end{equation}

\textbf{Step 3: combine everything. }

First, we claim that
\begin{equation} \label{eq:per_sample_gradient_projected_same}
    \nabla_{\Theta} c\left(\tilde{w}\left(P_{\Theta_{\rms}^*}^{\mathrm{s}} (x)\right), \xi\right) \Pi
    = \nabla_{\Theta} c\left(\tilde{w}\left(P_{\Theta_{\rms}^*}^{\mathrm{s}} (x)\right), \xi\right) .
\end{equation}
To see this, we first note that
$
    \nabla_{\Theta} c\left(\tilde{w}\left(P_{\Theta_{\rms}^*}^{\mathrm{s}} (x)\right), \xi\right) \Pi
    = \nabla_{\Theta} c\left(\tilde{w}\left(P_{\Theta_{\rms}^*}^{\mathrm{s}} (x)\right), \xi\right) \paren{ I_K - \frac{1}{K} \bfone \bfone^{\top} } .
$
Hence, it suffices to show that
$
    \nabla_{\Theta} c\left(\tilde{w}\left(P_{\Theta_{\rms}^*}^{\mathrm{s}} (x)\right), \xi\right) \bfone = 0 .
$

Given any $u \in \mathbb{R}^{d_{\phi}}$, we let $U  = u \bfone^{\top} \in \mathbb{R}^{d_{\phi} \times K}$.
Define the scalar function of one variable $t$ by
$
    \psi(t)  \defeq c\left(\tilde{w}\left(P_{\Theta_s^* + t U}^{\mathrm{s}} (x)\right), \xi\right) .
$
Then for every $t$,
\begin{equation}
    \paren{ \Theta_{\rms}^{*} + t u \bfone^{\top} }^{\top} \phi(x)
    = \Theta_{\rms}^{* \top} \phi(x) + t \bfone u^{\top} \phi(x)  \nonumber
\end{equation}
so the logits are shifted by the same constant $t u^{\top} \phi(x)$ across all classes.
Since softmax is invariant to such shifts, we have
$
    P_{\Theta_s^* + t U}^{\mathrm{s}} (x) = P_{\Theta_s^*}^{\mathrm{s}} (x) .
$
Therefore, $\psi(t)$ is a constant function in $t$. So we have $\psi'(0) = 0$.
By the chain rule, one can verify that
\begin{equation}
    0 = \psi'(0) = \inner{ \nabla_{\Theta} c\left(\tilde{w}\left(P_{\Theta_s^*}^{\mathrm{s}} (x)\right), \xi\right) , U }_F
    = \inner{ \nabla_{\Theta} c\left(\tilde{w}\left(P_{\Theta_s^*}^{\mathrm{s}} (x)\right), \xi\right) \bfone , u } . \nonumber
\end{equation}
Since $u$ is arbitrary, it follows that
\begin{equation}
    \nabla_{\Theta} c\left(\tilde{w}\left(P_{\Theta_s^*}^{\mathrm{s}} (x)\right), \xi\right) \bfone = 0 . \nonumber
\end{equation}
Hence, we have proved \eqref{eq:per_sample_gradient_projected_same}.

Now, we proceed to lower bound $\EE{ \tilde{\mathcal{B}} }{ \left\|\varphi_n\right\|_2^2 } $.
Recalling the definition $\varphi_n \defeq \operatorname{vec} \paren{ \nabla_{\Theta}  \frac{1}{n} \sum_{i=1}^{n} c\left(\tilde{w}\left(P_{\Theta}^{\mathrm{s}} (\tilde{x}_i)\right), \tilde{\xi}_i\right) \Pi }$, we have
\begin{eqnarray}
    \EE{ \tilde{\mathcal{B}} }{ \left\|\varphi_n\right\|_2^2 }
&=& \operatorname{tr}\left(\operatorname{Var}_{\tilde{\mathcal{B}}} \left(\varphi_n\right)\right)
+ \left\|\mathbb{E}_{ \tilde{\mathcal{B}} } \left[\varphi_n\right]\right\|_2^2 \nonumber  \\
&\geq& \operatorname{tr}\left(\operatorname{Var}_{\tilde{\mathcal{B}}} \left(\varphi_n\right)\right) \nonumber \\
&=& \frac{1}{n} \operatorname{tr}\left( \operatorname{Var}_{\tilde{\mathcal{B}}} \paren{ \operatorname{vec} \left( \nabla_{\Theta} c\left(\tilde{w}\left(P_{\Theta_{\rms}^*}^{\mathrm{s}} (\tilde{x}_i)\right), \tilde{\xi}_i\right) \Pi \right) } \right) \nonumber \\
&=& \frac{1}{n} \operatorname{tr}\left( \operatorname{Var}_{\tilde{\mathcal{B}}} \paren{ \operatorname{vec} \left( \nabla_{\Theta} c\left(\tilde{w}\left(P_{\Theta_{\rms}^*}^{\mathrm{s}} (\tilde{x}_i)\right), \tilde{\xi}_i\right)  \right) } \right) \label{eq:applyeq:per_sample_gradient_projected_same}\\
&=& \frac{1}{n} \operatorname{tr}\left(\Sigma_\varphi\right) .
\end{eqnarray}
The first equality follows from the fact that for any psd matrix $M$ and random vector $Z$, we have
$
\mathbb{E}\left[Z^{\top} M Z\right]=\operatorname{tr}(M \operatorname{Var}(Z))+\mathbb{E}[Z]^{\top} M \mathbb{E}[Z] \geq \operatorname{tr}(M \operatorname{Var}(Z))
$.
Equation~\eqref{eq:applyeq:per_sample_gradient_projected_same} follows from \eqref{eq:per_sample_gradient_projected_same}.

\end{myproof}

\subsection{Proof of Theorem~\ref{thm:weak_model_upper_bound}}
\label{appendix:proof_thm_weak_model}
\begin{myproof}

We denote $P_{\Theta}^{\rmw} (  x ) = \operatorname{softmax} \paren{ \Theta^{\top} \phi_{\rmw} (x) } \in \mathbb{R}^{K}$  and $p_{\Theta}^{\rmw}$ to be the corresponding probability mass function.
We let
\begin{equation}
    \ell \paren{ \Theta; \tilde{x}_i, \tilde{\xi}_i }
    \defeq - \ln p_{\Theta}^{\rmw} \paren{ \tilde{\xi}_i \mid \tilde{x}_i }
    = - \eta_{\Theta, \tilde{\xi}_i }^{\rmw} (\tilde{x}_i)
    + \ln \paren{ \sum_{k=1}^{K} e^{ \eta_{\Theta, k}^{\rmw} (\tilde{x}_i) } } .
\end{equation}
With slight abuse of notation, we denote $\theta = \operatorname{vec} ( \Theta ) \in \mathbb{R}^{d_{\phi} K}$ to be the vectorized version of $\Theta$.
We also write
\[
    \widetilde{\calC}_B
    \defeq
    \braces{\operatorname{vec}\paren{\Theta}:\Theta\in\calC_B},
    \qquad
    \widetilde{\calC}
    \defeq
    \braces{\operatorname{vec}\paren{\Delta}:\Delta\in\mathbb{R}^{d_{\phi}\times K},\ \Delta\bfone=0}.
\]
In this proof, we denote $F(\Theta) = \E{ \ell \paren{\Theta; \tilde{x}_i, \tilde{\xi}_i} }$ and $\widehat{F}_n(\Theta) = \frac{1}{n} \sum_{i=1}^{n} \ell \paren{\Theta; \tilde{x}_i, \tilde{\xi}_i }$.
We recall that $\Theta^{*}_{\rmw}$ is the minimizer of \eqref{eq:approximation_error_w}.
Under Assumption~\ref{assumption:identifiability_weak}, $\Theta^{*}_{\rmw}$ is also the minimizer of $\argmin_{\Theta \in \calC_B} F(\Theta)$, by noting that
\begin{equation}
    \argmin_{\Theta \in \calC_B} ~ F(\Theta)
    = \argmin_{\Theta \in \calC_B} ~ \E{ - \ln p_{\Theta}^{\rmw} \paren{ \tilde{\xi}_i \mid \tilde{x}_i } }
    = \argmin_{\Theta \in \calC_B} ~ \E{ \operatorname{KL} \paren{ P^*( \tilde{\xi}_i \mid \tilde{x}_i ) ~||~ P^{\rmw}_{\Theta} (\tilde{\xi}_i \mid \tilde{x}_i)  } } .
\end{equation}

Assumption~\ref{assumption:identifiability_weak} also implies $\Sigma_{\rmw}$ is nonsingular. Suppose instead that $v\neq0$ and $v^{\top}\Sigma_{\rmw}v=0$; since $\Sigma_{\rmw}=\E{\phi_{\rmw}(x)\phi_{\rmw}(x)^{\top}}$, this gives $v^{\top}\phi_{\rmw}(x)=0$ almost surely. Choose $a\neq0$ with $a^{\top}\bfone=0$, possible since $K\geq2$. Then $va^{\top}\bfone=0$, so the centering constraint in $\calC_B$ is preserved, and $\paren{va^{\top}}^{\top}\phi_{\rmw}(x)=a\,v^{\top}\phi_{\rmw}(x)=0$ almost surely. Hence $\Theta_{\rmw}^{*}+\epsilon va^{\top}$ leaves $\Theta^{\top}\phi_{\rmw}(x)$ unchanged and induces the same conditional model as $\Theta_{\rmw}^{*}$. Since $\Theta_{\rmw}^{*}$ is interior, $\Theta_{\rmw}^{*}+\epsilon va^{\top}\in\calC_B$ for all sufficiently small $\epsilon\neq0$, giving a distinct point with the same value of $F$ and contradicting uniqueness; hence $\lambda_{\min}\paren{\Sigma_{\rmw}}>0$ and $d_{\rmw}=d_{\phi}$.

Under the assumption that $\norm{\Theta_{\rmw}^{*}}_F < B_{\Theta}$, there exists $\nu \in \mathbb{R}^{d_{\phi}}$ such that the KKT conditions hold:
\begin{equation} \label{eq:mle_kkt_raw}
    \nabla_{\Theta} F(\Theta_{\rmw}^{*})  + \nu \bfone^{\top} = 0 , \quad \text{ and } \quad \Theta_{\rmw}^{*} \bfone = 0 .
\end{equation}
Hence, right multiplying $\bfone$ on both sides yields $ \nabla_{\Theta} F(\Theta_{\rmw}^{*}) \bfone + \nu \bfone^{\top} \bfone = \nabla_{\Theta} F(\Theta_{\rmw}^{*}) \bfone + \nu K = 0$.
One can show that
$\nabla_{\Theta} \ell \paren{\Theta; \tilde{x}_i, \tilde{\xi}_i} = \phi_{\rmw} (\tilde{x}_i)  \paren{\operatorname{softmax} \paren{ \Theta^{\top} \phi_{\rmw} (\tilde{x}_i) } - e_{\tilde{\xi}_i}}^{\top}  $.
Hence, $ \nabla_{\Theta} F(\Theta_{\rmw}^{*}) \bfone = \E{ \nabla_{\Theta} \ell \paren{\Theta_{\rmw}^{*}; \tilde{x}_i, \tilde{\xi}_i } \bfone  } = 0 $. Therefore, it has to be that $\nu = 0$ in the KKT condition \eqref{eq:mle_kkt_raw}.
Hence, the KKT condition \eqref{eq:mle_kkt_raw} reduces to
\begin{equation} \label{eq:KKT_mle_weak}
    \nabla_{\Theta} F(\Theta_{\rmw}^{*})  = 0 , \quad \text{ and } \quad \Theta_{\rmw}^{*} \bfone = 0 .
\end{equation}

Recalling the formula
$\operatorname{vec}\left(u v^{\top}\right)=v \otimes u$, we get
$\nabla_{\theta} \ell \paren{\theta; \tilde{x}_i, \tilde{\xi}_i } \defeq \operatorname{vec} \paren{ \nabla_{\Theta} \ell \paren{\Theta; \tilde{x}_i, \tilde{\xi}_i} }
= \paren{ \operatorname{softmax} \paren{ \Theta^{\top} \phi_{\rmw} (\tilde{x}_i) } - e_{\tilde{\xi}_i} } \otimes \phi_{\rmw} (\tilde{x}_i)
$.
Denote $J(p) = \operatorname{Diag}(p) - p p^{\top} \in \mathbb{R}^{K \times K}$ for probability vector $p \in \Delta_K$.
One can verify that the Hessian in vector $\theta$ is given by
\begin{equation} \label{eq:mle_ell_hessian}
    \nabla_{\theta}^2 \ell \paren{\theta; \tilde{x}_i, \tilde{\xi}_i }
    =  J \paren{ p_{\Theta}^{\rmw} ( \tilde{x}_i ) }
    \otimes \paren{ \phi_{\rmw} (\tilde{x}_i) \phi_{\rmw} (\tilde{x}_i)^{\top} } .
\end{equation}

We denote $\theta^* = \operatorname{vec} ( \Theta_{\rmw}^{*} )$ and $\widehat{\theta} = \operatorname{vec} ( \widehat{\Theta}_{\rmw} )$.
We let $\delta = \widehat{\theta} - \theta^*$ and $\Delta = \widehat{\Theta}_{\rmw} - \Theta_{\rmw}^*$.
By convexity of the objective $\widehat{F}_n$ and the feasible region $\widetilde{\calC}_B$, the optimality of $\widehat{\Theta}_{\rmw}$ implies that
\begin{equation}
    \paren{ \theta - \widehat{\theta}}^{\top} \nabla_{\theta} \widehat{F}_n ( \widehat{\theta} ) \geq 0 , \quad \forall ~ \theta \in \widetilde{\calC}_B . \nonumber
\end{equation}
In particular, taking $\theta = \theta^* \in \widetilde{\calC}_B$, we have
\begin{equation} \label{eq:mle_foc}
    \delta^{\top} \nabla_{\theta} \widehat{F}_n ( \widehat{\theta} ) \leq 0 .
\end{equation}

By the fundamental theorem of calculus, we have
\begin{equation} \label{eq:mle_theorem_of_calculus}
    \nabla_{\theta} \widehat{F}_n ( \widehat{\theta} ) - \nabla_{\theta} \widehat{F}_n ( \theta^* )
    = \int_0^1 \nabla_{\theta}^2 \widehat{F}_n ( \theta^* + t \delta ) dt \cdot \delta .
\end{equation}
For brevity, we denote $H_n = \int_0^1 \nabla_{\theta}^2 \widehat{F}_n ( \theta^* + t \delta ) dt$.

Combining \eqref{eq:mle_foc} and \eqref{eq:mle_theorem_of_calculus}, we have
\begin{equation}
    \delta^{\top} H_n \delta + \delta^{\top} \nabla_{\theta} \widehat{F}_n ( \theta^* )
    = \delta^{\top} \nabla_{\theta} \widehat{F}_n ( \widehat{\theta} ) \leq 0 .
\end{equation}

Let the good event $\calG$ be defined as
\begin{equation}
    \calG = \braces{  \inf_{ \theta \in \widetilde{\calC}_B } \inf_{ u \in \widetilde{\calC}, \normtwo{u} = 1 }
        u^{\top} \nabla_{\theta}^2 \widehat{F}_n ( \theta ) u \geq \frac{1}{2} \mu_{\rmw}
    } .
\end{equation}
On the good event $\calG$, we have
$
    \delta^{\top} H_n \delta
    \geq \frac{1}{2} \mu_{\rmw} \normtwo{\delta}^2
$, which in turn implies that $ \frac{1}{2} \mu_{\rmw} \normtwo{\delta}^2  \leq - \delta^{\top} \nabla_{\theta} \widehat{F}_n ( \theta^* ) \leq \normtwo{\delta} \normtwo{  \nabla_{\theta} \widehat{F}_n ( \theta^* ) } $ and hence
\begin{equation}  \label{eq:mle_good_event_implies}
    \normtwo{\delta}^2 \leq \frac{4}{\mu_{\rmw}^2} \normtwo{ \nabla_{\theta} \widehat{F}_n ( \theta^* ) }^2  .
\end{equation}
Therefore, by conditioning on the good event $\calG$, we conclude that
\begin{eqnarray}
    \E{ \normtwo{\delta}^2 } &=& \E{ \normtwo{\delta}^2 \one{\calG} + \normtwo{\delta}^2 \one{\calG^{\complement}} } \nonumber \\
    &\leq& \frac{4}{\mu_{\rmw}^2} \E{ \normtwo{ \nabla_{\theta} \widehat{F}_n ( \theta^* ) }^2 \one{\calG} } + \E{ \normtwo{\delta}^2 \one{\calG^{\complement}} } \label{eq:mle_use_good_event} \\
    &\leq& \frac{4}{\mu_{\rmw}^2} \E{ \normtwo{ \nabla_{\theta} \widehat{F}_n ( \theta^* ) }^2  }
    + 4 B_{\Theta}^2 \pr{ \calG^{\complement} } \label{eq:mle_delta_squared_decomposition} .
\end{eqnarray}
Inequality~\eqref{eq:mle_use_good_event} follows from \eqref{eq:mle_good_event_implies}.
Inequality~\eqref{eq:mle_delta_squared_decomposition} follows from the fact that both $\widehat{\Theta}_{\rmw}$ and $\Theta_{\rmw}^{*}$ lie in $\calC_B$.
To control $\E{ \normtwo{ \nabla_{\theta} \widehat{F}_n ( \theta^* ) }^2  } $, we observe that
\begin{eqnarray}
    \E{ \normtwo{ \nabla_{\theta} \widehat{F}_n ( \theta^* ) }^2 } &=& \E{  \normtwo{ \frac{1}{n} \sum_{i=1}^{n} \nabla_{\theta} \ell \paren{\theta^*; \tilde{x}_i, \tilde{\xi}_i } }^2 } \nonumber \\
    &=& \frac{1}{n^2} \sum_{i=1}^{n} \E{ \normtwo{ \nabla_{\theta} \ell \paren{\theta^*; \tilde{x}_i, \tilde{\xi}_i } }^2 }
    + \frac{2}{n^2} \E{ \sum_{i < j} \inner{ \nabla_{\theta} \ell \paren{\theta^*; \tilde{x}_i, \tilde{\xi}_i } , \nabla_{\theta} \ell \paren{\theta^*; \tilde{x}_j, \tilde{\xi}_j } } } \nonumber \\
    &=& \frac{1}{n^2} \sum_{i=1}^{n} \E{ \normtwo{ \nabla_{\theta} \ell \paren{\theta^*; \tilde{x}_i, \tilde{\xi}_i } }^2 } ,
\end{eqnarray}
where the last equality follows from the independence of the samples and the fact that $\E{ \nabla_{\theta} \ell \paren{\theta^*; \tilde{x}_i, \tilde{\xi}_i } } = \nabla_{\theta} F(\theta^*) = 0$ by \eqref{eq:KKT_mle_weak}.
Now, using the algebraic identity $\normtwo{a \otimes b} = \normtwo{a} \normtwo{b}$, we have
\begin{eqnarray}
    \E{ \normtwo{ \nabla_{\theta} \ell \paren{\theta^*; \tilde{x}_i, \tilde{\xi}_i } }^2 }
    &=& \E{ \normtwo{ \paren{ \operatorname{softmax} \paren{ \paren{\Theta_{\rmw}^{*}}^{\top} \phi_{\rmw} (\tilde{x}_i) } - e_{\tilde{\xi}_i} } \otimes \phi_{\rmw} (\tilde{x}_i) }^2 } \nonumber \\
    &=& \E{ \normtwo{ \operatorname{softmax} \paren{ \paren{\Theta_{\rmw}^{*}}^{\top} \phi_{\rmw} (\tilde{x}_i) } - e_{\tilde{\xi}_i}}^2 \normtwo{ \phi_{\rmw} (\tilde{x}_i)}^2 }  .
\end{eqnarray}
To proceed, for brevity, we denote $p \defeq \operatorname{softmax} \paren{ \paren{\Theta_{\rmw}^{*}}^{\top} \phi_{\rmw} (\tilde{x}_i) } \in \mathbb{R}^K$ and $p_k$ to be its $k$th coordinate.
We denote $e_{\tilde{\xi}}$ to be the one-hot encoding in $\mathbb{R}^K$ with $1$ in the $\tilde{\xi}$-th coordinate and $0$ elsewhere.
We note that conditional on $\tilde{x}_i$,
\begin{eqnarray}
    \E{ \normtwo{p-e_{\tilde{\xi}_i}}^2 \mid \tilde{x}_i  } &=& \sum_{k=1}^{K} p_k \normtwo{p - e_k}^2
    = \sum_{k=1}^K p_k\left(\|p\|_2^2+1-2 p_k\right)
    = 1 - \|p\|_2^2 \leq 1 ,
\end{eqnarray}
and hence
\begin{equation}
    \E{  \normtwo{ \nabla_{\theta} \ell \paren{\theta^*; \tilde{x}_i, \tilde{\xi}_i } }^2 }
 =  \E{ \normtwo{ \phi_{\rmw} (\tilde{x}_i)}^2 \E{ \normtwo{p-e_{\tilde{\xi}_i}}^2 \mid \tilde{x}_i  } }
 \leq \E{ \normtwo{ \phi_{\rmw} (\tilde{x}_i)}^2 } = \tr{\Sigma_{\rmw}}.
\end{equation}

It remains to bound $\pr{ \calG^{\complement} }$ in \eqref{eq:mle_delta_squared_decomposition}. To this end, we introduce the notation $\widehat{\Sigma}_n \defeq \frac{1}{n} \sum_{i=1}^{n} \phi_{\rmw} (\tilde{x}_i) \phi_{\rmw} (\tilde{x}_i)^{\top}$.

We proceed to show that whenever $\lambda_{\min} \paren{ \widehat{\Sigma}_n } \geq \frac{1}{2} \lambda_{\min} \paren{ \Sigma_{\rmw} }$, then for any $\theta \in \widetilde{\calC}_B$ and every $\delta \in \widetilde{\calC}$,
\begin{equation} \label{eq:covgood_implies_good}
    \delta^{\top} \nabla_{\theta}^2 \widehat{F}_n ( \theta ) \delta
    \overset{(\text{*})}{\geq} \frac{e^{- 2 B_{\Theta}B_{\phi}}}{K}  \lambda_{\min} \paren{ \widehat{\Sigma}_n } \normtwo{ \delta }^2
    \geq \frac{1}{2} \frac{e^{- 2 B_{\Theta}B_{\phi}}}{K}  \lambda_{\min} \paren{ \Sigma_{\rmw} } \normtwo{ \delta }^2 ,
\end{equation}
which implies that the good event $\calG$ holds.
A standard matrix Chernoff lower tail bound ensures that for $\varepsilon \in (0,1)$,
\begin{equation}
    \pr{ \lambda_{\min} \paren{ \widehat{\Sigma}_{n} } \leq (1-\varepsilon) \lambda_{\min} \paren{ \Sigma_{\rmw} }  }
    \leq d_{\rmw} \exp \paren{ - \frac{ n \lambda_{\min} \paren{ \Sigma_{\rmw} } \varepsilon^2 }{ 2 B_{\phi}^2 } } .
\end{equation}
Taking $\varepsilon = \frac{1}{2}$ yields that
\begin{equation}
    \pr{ \calG^{\complement} } \leq d_{\rmw} \exp \paren{ - \frac{ n \lambda_{\min} \paren{ \Sigma_{\rmw} } }{ 8 B_{\phi}^2 } } .
\end{equation}
Now, we turn to prove the inequality marked as $(\text{*})$ above in \eqref{eq:covgood_implies_good}.
For any $\delta \in \widetilde{\calC}$,
\begin{eqnarray}
    \delta^{\top} \nabla_{\theta}^2 \widehat{F}_n ( \theta ) \delta
    &=& \frac{1}{n} \sum_{i=1}^{n} \delta^{\top} \nabla_{\theta}^2 \ell \paren{\theta; \tilde{x}_i, \tilde{\xi}_i } \delta \nonumber \\
    &\overset{(\text{a})}{=}& \frac{1}{n} \sum_{i=1}^{n} \delta^{\top} \paren{ J \paren{ p_{\Theta}^{\rmw} ( \tilde{x}_i ) } \otimes \paren{ \phi_{\rmw} (\tilde{x}_i) \phi_{\rmw} (\tilde{x}_i)^{\top} } } \delta \nonumber \\
    &\overset{(\text{b})}{=}& \frac{1}{n} \sum_{i=1}^{n} \tr{ \phi_{\rmw} (\tilde{x}_i) \phi_{\rmw} (\tilde{x}_i)^{\top} \Delta J \paren{ p_{\Theta}^{\rmw} ( \tilde{x}_i ) }  \Delta^{\top} } \nonumber \\
    &\overset{(\text{c})}{=}& \frac{1}{n} \sum_{i=1}^{n} \paren{ \Delta^{\top} \phi_{\rmw} (\tilde{x}_i) }^{\top} J \paren{ p_{\Theta}^{\rmw} ( \tilde{x}_i ) } \paren{ \Delta^{\top} \phi_{\rmw} (\tilde{x}_i) } \nonumber \\
    &\overset{(\text{d})}{\geq}& \frac{e^{- 2 B_{\Theta} B_{\phi}}}{K} \frac{1}{n} \sum_{i=1}^{n} \normtwo{ \Delta^{\top} \phi_{\rmw} (\tilde{x}_i) }^2  . \label{eq:mle_lower_bound_delta_hat_F_delta_first_step}
\end{eqnarray}
In Equation~(a), we plug in the expression of the Hessian \eqref{eq:mle_ell_hessian}.
Equation~(b) follows from the algebraic identity $\operatorname{vec}(X)^{\top}(A \otimes B) \operatorname{vec}(X)=\operatorname{tr}\left(X^{\top} B X A\right)$ for compatible matrices $A,B,X$.
Equation~(c) is due to the cyclic property of trace.
To conclude Inequality~(d), we need the following Lemma.

\begin{lemma} \label{lemma:softmax_J_dominates_projection_under_cushion}
    Let $p \in \Delta_K = \braces{ u \in \mathbb{R}^K : u_k \geq 0, \sum_{k=1}^{K} u_k = 1}$. Let $m = \min_{k \in [K]} p_k$.
    We define $J(p) = \operatorname{diag}(p) - p p^{\top} \in \mathbb{R}^{K \times K}$.
    Then for every $u \in \mathbb{R}^K$ satisfying $\bfone^{\top} u = 0$, we have
    \begin{equation}
        u^{\top} J(p) u \geq m \norm{u}_2^2 .
    \end{equation}
\end{lemma}

\begin{myproof}
    Denote $q = p - m \bfone \in \mathbb{R}^K$, so $q_k \geq 0$ and $\sum_{k=1}^{K} q_k = 1 - K m$.
    For any $u$ with $\mathbf{1}^{\top} u=0$, we have $p^{\top} u=(m \mathbf{1}+q)^{\top} u=q^{\top} u$.
    Then
    \begin{eqnarray}
        u^{\top} \paren{ \operatorname{diag} \paren{p} - p p^{\top} } u
        &=& \sum_{k=1}^{K} p_k u_k^2 - (p^{\top} u)^2 \nonumber \\
        &=& m \sum_{k=1}^{K} u_k^2 + \sum_{k=1}^{K} q_k u_k^2 - (q^{\top} u)^2 \nonumber \\
        &\geq& m \norm{u}_2^2 + \sum_{k=1}^{K} q_k u_k^2 - \paren{1-Km} \sum_{k=1}^{K} q_k u_k^2 \label{eq:apply_cauchy_J_eta_u} \\
        &=& m \norm{u}_2^2 + K m \sum_{k=1}^{K} q_k u_k^2 \nonumber \\
        &\geq& m \norm{u}_2^2 . \label{eq:diag_p_p_p_top_geq}
    \end{eqnarray}
    Inequality~\eqref{eq:apply_cauchy_J_eta_u} follows from the Cauchy-Schwarz inequality.

\end{myproof}

We note that given $ \norm{\Theta}_F \leq B_{\Theta}$ and $ \normtwo{\phi_{\rmw}(x)} \leq B_{\phi}$, it is straightforward to verify that for every $x \in \mathcal{X}$,
\begin{equation}
    \min_{k \in [K]} P_{\Theta}^{\rmw} ( \xi = z_k \mid x)
    \geq \frac{ e^{-2 B_{\Theta} B_{\phi}} }{K} ,
\end{equation}
as $P_{\Theta}^{\rmw} ( \xi = z_k \mid x) = \frac{e^{e_k^{\top} \Theta^{\top} \phi_{\rmw}(x)}}{\sum_{j=1}^{K} e^{e_j^{\top} \Theta^{\top} \phi_{\rmw}(x)}} \geq \frac{e^{- B_{\Theta} B_{\phi} }}{K e^{ B_{\Theta} B_{\phi} }}$.
Moreover, by noting that $ \paren{\Delta^{\top} \phi_{\rmw} (\tilde{x}_i) }^{\top} \bfone = \phi_{\rmw}(\tilde{x}_i)^{\top} \Delta \bfone = 0$, Lemma~\ref{lemma:softmax_J_dominates_projection_under_cushion} implies that Inequality~(d) holds.

Then, continuing from \eqref{eq:mle_lower_bound_delta_hat_F_delta_first_step}, by cyclic and linear properties of trace, we have
\begin{eqnarray}
    \frac{1}{n} \sum_{i=1}^{n} \normtwo{ \Delta^{\top} \phi_{\rmw} (\tilde{x}_i) }^2
    &=& \tr{ \Delta^{\top} \paren{ \frac{1}{n} \sum_{i=1}^{n} \phi_{\rmw} (\tilde{x}_i) \phi_{\rmw} (\tilde{x}_i)^{\top} } \Delta }
    \geq \lambda_{\min} \paren{ \widehat{\Sigma}_n } \norm{ \Delta }_{F}^2
    = \lambda_{\min} \paren{ \widehat{\Sigma}_n } \normtwo{ \delta }^2 , \nonumber
\end{eqnarray}
where the inequality follows from the fact that for any real symmetric matrix $M$, $\tr{ X^{\top} M X } \geq \lambda_{\min} (M) \norm{X}_F^2$ for any compatible matrix $X$.

The proof is now complete.

\end{myproof}

\section{Additional Experiments for Section~\ref{subsec:synthetic_experiments}}
\label{appendix:two_product_newsvendor}

In this section, we consider a multi-item newsvendor problem with a capacity constraint, and we show that the empirical results remain consistent with the theory, in the constrained case as well.

\paragraph{Setup.}
We consider two products. The demand vector has finite support $\Xi=\left\{\left(z_1, z_2\right) \in \mathbb{Z}_{+}^2: z_1+z_2 \leq C_z \right\}$.
The feasible set of our action is $\mathcal{A}=\left\{\left(w_1, w_2\right) \in \mathbb{R}_{+}^2: w_1+w_2 \leq C_w \right\}$ where $C_w < C_z$, which ensures the constraint binding.
Given demand $z$, the cost function is $ c(z,w) = \sum_{i=1}^{2} c_{u,i} (z_i - w_i)_+ + c_{o,i} (w_i-z_i)_+ $.
The strong and weak features are constructed in the same spirit as in Section~\ref{subsec:synthetic_experiments}, so that we can precisely control the overlap dimension.

In this constrained setting, the optimal order vector does not admit a closed-form expression. During training, we therefore use a differentiable soft-route surrogate analogous to Equation~\eqref{eq:surrogate_oracle_synthetic}. For evaluation, we compute the exact optimal action by enumerating the finite set of feasible actions.

We sample the context vector $x$ according to $x \sim \mathcal{N}\left(0, I_{d_x}\right)$. Let $q_{\mathrm{M}}$ and $q_{\mathrm{P}}$ be two orthonormal directions contained in both the strong and weak feature subspaces. We define the conditional demand rates by
$
\log \lambda_1(x)=\log \lambda_1^0+\alpha_{\mathrm{M}} q_{\mathrm{M}}^{\top} x+\alpha_{\mathrm{P}} q_{\mathrm{P}}^{\top} x
$
and
$
\log \lambda_2(x)=\log \lambda_2^0+\alpha_{\mathrm{M}} q_{\mathrm{M}}^{\top} x-\alpha_{\mathrm{P}} q_{\mathrm{P}}^{\top} x .
$
Conditional on $x$, we independently sample two Poisson random variables
$$
\tilde{\xi}_1 \sim \operatorname{Poisson}\left(\lambda_1(x)\right), \quad \tilde{\xi}_2 \sim \operatorname{Poisson}\left(\lambda_2(x)\right) .
$$
If $\tilde{\xi}_1+\tilde{\xi}_2>C_z$, we reject the pair and resample; otherwise, we set the demand realization to be $\xi=\left(\tilde{\xi}_1, \tilde{\xi}_2\right)$. This rejection procedure ensures that the resulting demand vector has support $\Xi$.

For the numerical study, we set $d_x=80$, $d_{\rms}=10$, $d_{\rmw}=48$, $C_z=12$, and $C_w=9$. We take $(\lambda_1^0,\lambda_2^0)=(4,3)$, $\alpha_{\mathrm{M}}=\alpha_{\mathrm{P}}=0.18$, $(c_{u,1},c_{u,2})=(3,4)$, and $(c_{o,1},c_{o,2})=(1,1.2)$. We write $r=d_{\rms\wedge\rmw}$ and vary $n\in\{100,300,1000\}$, $N\in\{100,300,1000,3000,10000\}$, and $r\in\{2,3,4,6,8,10\}$. Each configuration is evaluated over 60 independent trials.

\begin{figure}[htbp]
    \centering
    \includegraphics[width=\linewidth]{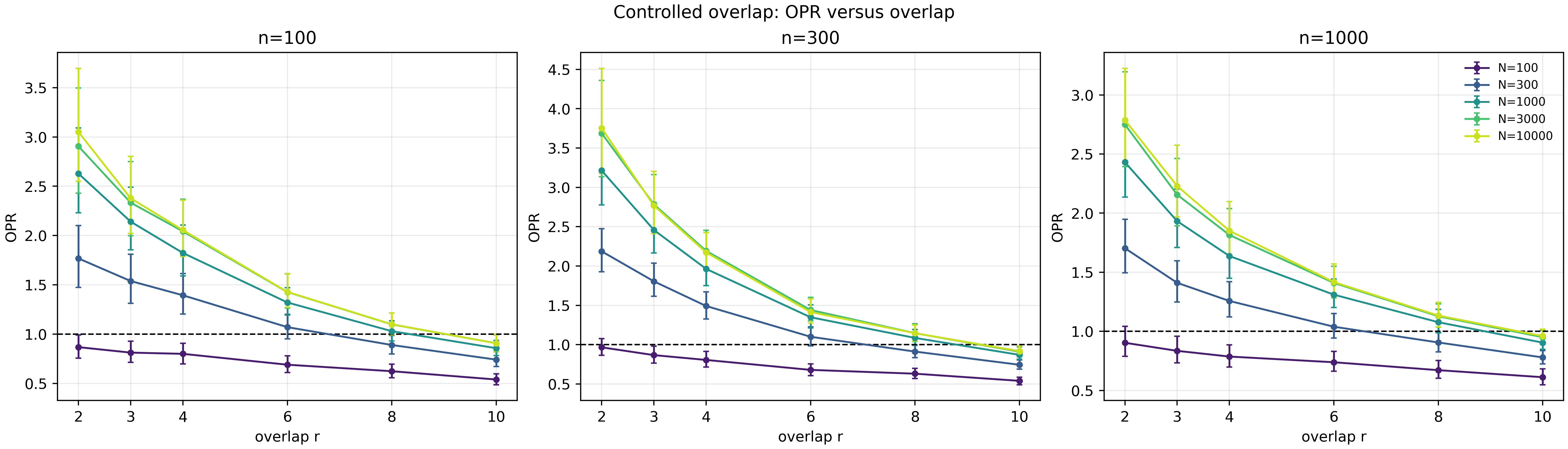}
    \caption{
    OPR versus overlap dimension in the capacity-constrained two-product newsvendor problem. Each panel fixes $n$, and each curve varies $N$. The dashed line marks $\mathrm{OPR}=1$, and the error bars report 95\% confidence intervals using bootstrap over 60 independent trials.
    }
    \label{fig:two_product_opr_vs_overlap}
\end{figure}

\begin{figure}[htbp]
    \centering
    \includegraphics[width=\linewidth]{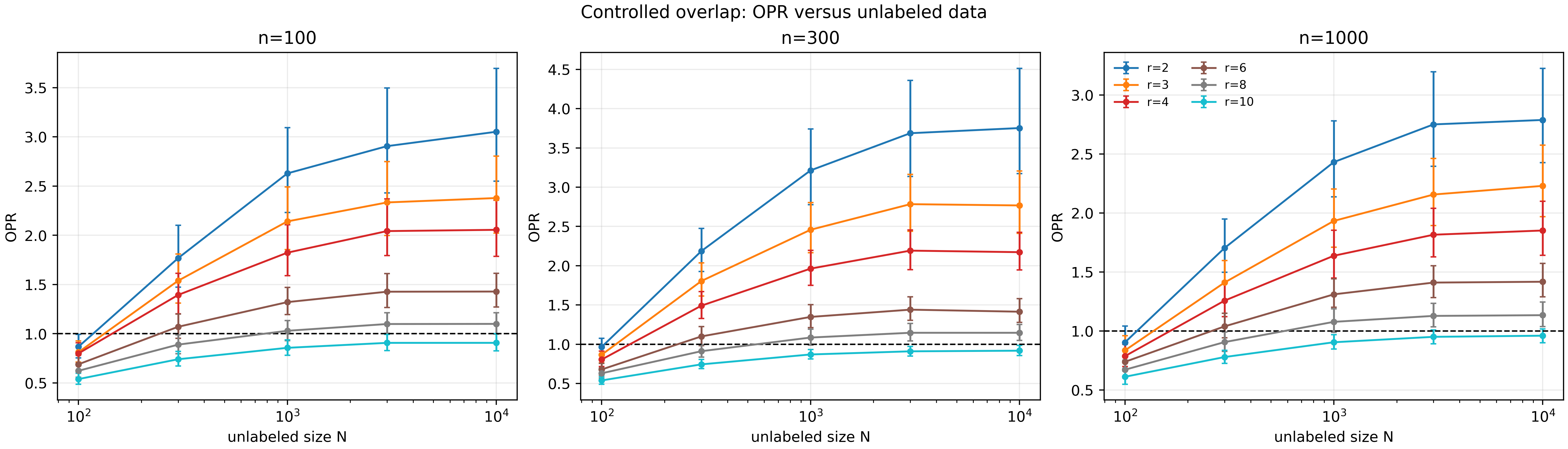}
    \caption{
    Effect of the unlabeled sample size in the capacity-constrained two-product newsvendor problem. Each panel fixes the labeled sample size $n$, and each curve corresponds to an overlap dimension $r=d_{\rms\wedge\rmw}$. The dashed horizontal line marks $\mathrm{OPR}=1$. The error bars report 95\% confidence intervals using bootstrap over 60 independent trials
    }
    \label{fig:two_product_opr_vs_N}
\end{figure}

\paragraph{Results.}
Figures~\ref{fig:two_product_opr_vs_overlap} and~\ref{fig:two_product_opr_vs_N} show that the two main comparative-statics patterns from the single-item experiment continue to hold under the shared capacity constraint. First, for every displayed pair $(n,N)$, the OPR decreases as the overlap dimension $r$ increases. Second, for fixed $(n,r)$, the OPR generally rises with $N$ and then levels off. In particular, when $N=100$, the OPR is below one throughout the displayed grid, whereas increasing $N$ produces substantial gains when the overlap is small. This behavior is consistent with the decomposition following Theorem~\ref{thm:upper_bound_w2s}: increasing $N$ reduces the unlabeled-sample statistical error and the $N^{-1}$ component of the weak-to-strong term, while a larger $d_{\rms\wedge\rmw}$ deteriorates the OPR. Thus, the same empirical patterns persist when the two order quantities are coupled through a shared capacity constraint, providing further evidence that our framework extends beyond the single-item setting.

\section{Additional Details for Section~\ref{subsec:comment_moderation_experiments}}
\label{appendix:moderation_model_pretraining}
\label{appendix:moderation_cost_matrix}

\paragraph{Pre-training} The task pre-training stage uses $2000$ labeled comments sampled from the pre-training split of the augmented comment moderation data.
The weak model is trained by cross-entropy for $500$ optimizer steps with batch size $32$, learning rate $5\cdot 10^{-5}$, and weight decay $0.01$.
The strong model is trained by cross-entropy for $100$ optimizer steps with batch size $32$, learning rate $2\cdot 10^{-5}$, and weight decay $0.01$.
In both cases, all encoder parameters and classifier-head parameters are trainable.
The resulting weak model initializes the downstream weak-teacher runs, and the resulting strong model initializes the downstream strong-only and W2S runs.

\paragraph{Cost matrix} The route-severity cost matrix used in the comment moderation experiment is
\[
\begin{array}{c|ccc}
    & \xi=0 & \xi=1 & \xi=2 \\
    \hline
    \text{allow} & 0.000 & 1.000 & 3.000 \\
    \text{remove} & 2.050 & 1.050 & 3.050 \\
    \text{standard review} & 0.300 & 2.700 & 11.573 \\
    \text{hold review} & 2.400 & 0.800 & 3.218 \\
    \text{priority review} & 2.850 & 0.967 & 1.706
\end{array}
\]
where $\xi=0,1,2$ denote non-toxic, toxic, and severely toxic comments, respectively.
The entries combine exposure cost, delay cost, human-review cost, user-friction cost from suppressing benign comments, and penalties for automatic toxic removals.
For example, when $\xi=2$, the matrix assigns cost $3.000$ to automatic approval, $3.050$ to automatic removal, and $1.706$ to priority review.
This reflects the following modeling choice: automatic approval leaves harmful content visible, while automatic removal avoids exposure but is penalized for making an enforcement decision without human review.
Priority review is cheapest in this column because it handles the severe comment quickly while still using human review.
The W2S framework does not rely on this particular numerical specification; other applications or moderation policies can be represented by replacing this matrix with the appropriate route-severity cost matrix.

\begin{figure}[H]
    \centering
    \includegraphics[width=0.75\linewidth]{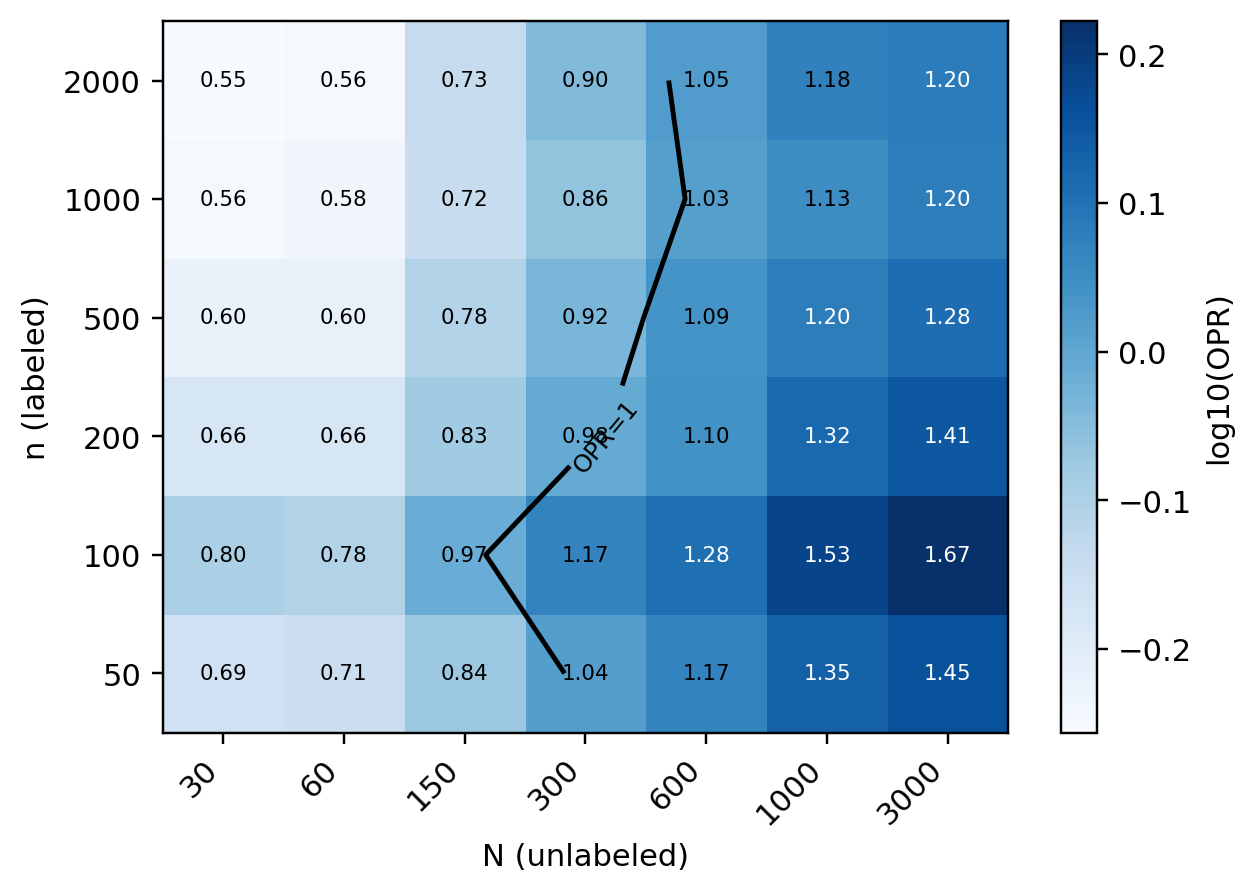}
    \caption{
    The heatmap reports the log-scaled empirical hard-route OPR over the labeled and unlabeled sample-size grid; the contour marks empirical hard-route OPR equal to one.
    }
    \label{fig:moderation_hard_route_opr_heatmap}
\end{figure}

\clearpage
\section{Standard Facts}

\begin{lemma} [Corollary 4 in {\citep{maurer2016vector}}] \label{lemma:vector_contraction_inequality}
    Let $\mathcal{X}$ be any set, and fix samples $x_1, \cdots, x_n \in \mathcal{X}$.
    Denote $\ell_2$ the real Hilbert space of square-summable sequences equipped with the norm $\|u\|_2:=\left(\sum_{k \geq 1} u_k^2\right)^{1 / 2}$.
    For any function class $\mathcal{F}$ mapping $\mathcal{X}$ to $\ell_2$, we write $f(x)=\left(f_k(x)\right)_{k \geq 1}$.
    Let $\mathcal{F}$ be any class of functions $f: \mathcal{X} \rightarrow \ell_2$. Let $h_i: \ell_2 \rightarrow \mathbb{R}$ for $i=1, \ldots, n$ be $L$-Lipschitz with respect to $\|\cdot\|_2$, i.e. $\left|h_i(u)-h_i(v)\right| \leq L\|u-v\|_2 $ for all $u, v \in \ell_2$, and all $i$.
    Let $\varepsilon=\left(\varepsilon_i\right)_{i=1}^n$ be independent Rademacher random variables (each takes values in $\{-1,+1\}$ with probability $1 / 2)$, and let $\varepsilon'=\left(\varepsilon_{i k}\right)_{i=1, \ldots, n ; k \geq 1}$ be an independent doubly indexed family of Rademacher variables. Then
    \begin{equation}
        \mathbb{E}_{\varepsilon}\left[\sup _{f \in \mathcal{F}} \sum_{i=1}^n \varepsilon_i h_i\left(f\left(x_i\right)\right)\right] \leq \sqrt{2} L \mathbb{E}_{\varepsilon^{\prime}}\left[\sup _{f \in \mathcal{F}} \sum_{i=1}^n \sum_{k \geq 1} \varepsilon_{i k} f_k\left(x_i\right)\right] .
    \end{equation}
\end{lemma}

\begin{lemma}[Theorem~5.39 in \cite{vershynin2010introduction}] \label{lemma:matrix_concentration_2_norm}
    Assume that $A$ is an $N \times n$ real matrix whose rows $A_i$ are independent sub-gaussian random vectors in $\mathbb{R}^n$ with second moment matrix $\Sigma = \E{ A_i A_i^{\top} }$. Then for every $t \geq 0$, the following inequality holds with probability at least $1-2 \exp(-c t^2)$:
    \begin{equation}
        \normtwo{ \frac{1}{N} A^{\top} A - \Sigma } \leq \max \left( \delta, \delta^2 \right), \quad \text{where } \delta = C \sqrt{\frac{n}{N}} + \frac{t}{\sqrt{N}} .
    \end{equation}
    Here $C$ and $c$ are constants that depend only on the sub-gaussian norm of the rows $A_i$.
\end{lemma}

\end{APPENDICES}
\bibliographystyle{plainnat}
\bibliography{ref.bib}

\end{document}

%% file: macros.tex
\usepackage[dvipsnames]{xcolor}

\newcommand{\RN}[1]{%
  \textup{\uppercase\expandafter{\romannumeral#1}}%
}

\usepackage[hidelinks]{hyperref}

\usepackage[ruled, vlined, linesnumbered, commentsnumbered, longend]{algorithm2e}
\SetKwInput{KwInput}{Input}
\SetKwInput{KwOutput}{Output}

\usepackage{endnotes}
\usepackage{bbm}
\usepackage{amssymb}
\usepackage[normalem]{ulem}
\usepackage{graphicx}
\usepackage{float}
\usepackage{epstopdf}
\usepackage{subcaption}
\usepackage{caption,enumitem,thm-restate}
\usepackage{enumitem}
\setlist{leftmargin=*}
\let\footnote=\endnote

\usepackage{natbib}
 \bibpunct[, ]{(}{)}{,}{a}{}{,}%
 \def\bibfont{\small}%
\usepackage{bm}
\usepackage{comment}

\newenvironment{myproof}{ \paragraph{Proof: } } {\hfill$\square$}

\newcommand{\E}[1]{\mathbb{E} \left[ {#1} \right]}
\newcommand{\EE}[2]{\mathbb{E}_{ {#1} } \left[ {#2} \right]}
\newcommand{\one}[1]{\mathbbm{1} \left[ {#1} \right]}
\newcommand{\inner}[1]{\left \langle {#1} \right \rangle }
\newcommand{\pr}[1]{\mathbf{Pr} \left[ {#1} \right]}
\newcommand{\prr}[2]{\mathbf{Pr}_{ {#1} } \left[  \allowbreak {#2} \allowbreak \right]}
\newcommand{\tr}[1]{\operatorname{tr} \left( {#1} \right)}

\definecolor{darkolivegreen}{rgb}{0.33, 0.42, 0.18}
\DeclareSymbolFont{extraup}{U}{zavm}{m}{n}
\DeclareMathSymbol{\varheart}{\mathalpha}{extraup}{86}
\DeclareMathSymbol{\vardiamond}{\mathalpha}{extraup}{87}

\newcommand{\bfsigma}{\bm{\sigma}}

\newcommand{\bfone}{\mathbf{1}}

\newcommand{\defeq}{\stackrel{\rm def}{=}}

\newcommand{\calA}{\mathcal{A}}
\newcommand{\calB}{\mathcal{B}}
\newcommand{\calC}{\mathcal{C}}
\newcommand{\calD}{\mathcal{D}}

\newcommand{\calG}{\mathcal{G}}
\newcommand{\calH}{\mathcal{H}}

\newcommand{\rmw}{\mathrm{w}}
\newcommand{\rms}{\mathrm{s}}

\newcommand{\order}[1]{O \left({#1}\right)}

\newcommand{\paren}[1]{\left({#1}\right)}
\newcommand{\brackets}[1]{\left[{#1}\right]}
\newcommand{\braces}[1]{\left\{{#1}\right\}}
\newcommand{\abs}[1]{\left | {#1} \right |}
\newcommand{\normtwo}[1]{\left \| {#1} \right \|_{2}}

\newcommand{\norm}[1]{\left \| {#1} \right \|}